\documentclass[twoside]{article}

\usepackage[accepted]{aistats2025}

\usepackage[hidelinks]{hyperref}
\usepackage{url}

\usepackage{caption}
\usepackage{subcaption}

\usepackage{mathtools}
\usepackage{amssymb}
\usepackage{algorithm}
\usepackage{algpseudocode}
\usepackage{enumitem}
\usepackage[dvipsnames]{xcolor}
\usepackage{cleveref}

\usepackage{listings}
\newcommand{\nocontentsline}[3]{}
\newcommand{\tocless}[2]{\bgroup\let\addcontentsline=\nocontentsline#1{#2}\egroup}

\makeatletter
\renewcommand*\env@matrix[1][\arraystretch]{%
  \edef\arraystretch{#1}%
  \hskip -\arraycolsep
  \let\@ifnextchar\new@ifnextchar
  \array{*\c@MaxMatrixCols c}}
\makeatother

\usepackage{chngcntr}

\newtheorem{definition}{Definition}[section]

\newcommand{\x}{{\boldsymbol{x}}}
\newcommand{\X}{{\boldsymbol{X}}}
\newcommand{\z}{{\boldsymbol{z}}}
\newcommand{\Z}{{\boldsymbol{Z}}}
\newcommand{\f}{{\boldsymbol{f}}}
\newcommand{\g}{{\boldsymbol{g}}}
\newcommand{\J}{{\boldsymbol{J}}} %

\renewcommand{\r}{{\boldsymbol{r}}}
\newcommand{\ph}{{\boldsymbol{\varphi}}}
\newcommand{\sig}{\boldsymbol{\sigma}}
\newcommand{\Jphi}{{\boldsymbol{J}^{\Phi}}}
\newcommand{\Jphiu}[1]{{\boldsymbol{J}^{\Phi}_{#1}}}

\newcommand{\A}{\boldsymbol{A}}

\renewcommand{\b}{\boldsymbol{b}}

\newcommand{\mm}{{\boldsymbol{\mu}}}

\renewcommand{\S}{\boldsymbol{\Sigma}}

\newcommand{\U}{\boldsymbol{U}}

\newcommand{\V}{\boldsymbol{V}}

\renewcommand{\L}{\boldsymbol{\Lambda}}
\newcommand{\I}{\boldsymbol{I}}

\newcommand{\set}[1]{{\mathbb{#1}}}
\newcommand{\nset}[1]{{\overline{\mathbb{#1}}}}

\newcommand{\Man}{{\mathcal{M}}}

\DeclareMathOperator{\Expectation}{\mathbb{E}}
\newcommand{\Expt}[2]{\Expectation\displaylimits_{#1}\left[#2\right]}

\usepackage[round]{natbib}

\newcommand{\uk}[1]{{#1}} %

\newcommand{\obso}[1]{} %

\begin{document}

\twocolumn[

\aistatstitle{Analyzing Generative Models by Manifold Entropic Metrics}
\vspace{-10pt}
\aistatsauthor{ Daniel Galperin \And Ullrich Köthe }

\aistatsaddress{ Heidelberg University \And Heidelberg University}
\vspace{-20pt}
\aistatsaddress{\texttt{daniel.galperin@iwr.uni-heidelberg.de} \And \texttt{ullrich.koethe@iwr.uni-heidelberg.de}}

]

\begin{abstract}

Good generative models should not only synthesize high quality data, but also utilize interpretable representations that aid human understanding of their behavior.
However, it is difficult to measure objectively if and to what degree desirable properties of disentangled representations have been achieved.
Inspired by the principle of independent mechanisms, we address this difficulty by introducing a novel set of tractable information-theoretic evaluation metrics.
We demonstrate the usefulness of our metrics on illustrative toy examples and conduct an in-depth comparison of various normalizing flow architectures and $\beta$-VAEs on the EMNIST dataset.
Our method allows to sort latent features by importance and assess the amount of residual correlations of the resulting concepts.
The most interesting finding of our experiments is a ranking of model architectures and training procedures in terms of their inductive bias to converge to aligned and disentangled representations during training.

\end{abstract}

\tocless\section{INTRODUCTION}
\uk{Deep generative models -- for example VAEs, GANs, normalizing flows, diffusion models, and flow matching -- have recently made great progress in Density Estimation (DE), with flow matching achieving highest accuracy at the moment.
Besides generative accuracy, human interpretability of the learned representation is highly desirable, and Disentangled Representation Learning (DRL) is a key tool for this, see \cite{bengio2014representationlearningreviewnew} and \cite{wang2024disentangledrepresentationlearning}.
Intuitively, DRL means that each latent variable should effect only a single, distinct semantic property of the generated data instances.

We consider the problem of {\em measuring} if and to what degree a given model has actually learned a disentangled representation.
Most prior work addresses this question in a supervised setting, where the true generative factors are known (see Related Work).
Since this assumption is often violated in the real world, we instead focus on the unsupervised case.
That is, we do not ask if the model has learned the (unknown) true factors, but only if it has learned any disentangled representation at all.
The learned representation might be close to the true one, if certain identifiability conditions are fulfilled, but this is beyond the scope of our paper.

Our work rests on the {\em manifold hypothesis} which states that data in a $D$-dimensional space often reside near a manifold $\mathcal{M}$ of much lower dimension $d\ll D$.
Variations along the manifold correspond to semantically important differences between data instances, whereas off-manifold variations are considered as unimportant or noise.
This is familiar from PCA, where one interprets directions of high data variability as important, whereas directions of low variability are irrelevant.
PCA achieves this under the assumption that the manifold $\mathcal{M}$ is a linear subspace, and DRL seeks to generalize this to non-linear models.

If the important dimensions indeed span the manifold $\mathcal{M}$, the representation is called {\em aligned}.
Our method clearly highlights \textit{Alignment} by assigning a high manifold entropy to the important features and low to unimportant ones, see \cref{fig:Sketch of derivation}.
Moreover, it allows sorting of latent variables by importance so that the cut-off between important and irrelevant can be adjusted later according to the needs of an application, analogous to PCA's ordering by variance.

\begin{figure*}
    \centering
    \includegraphics[width=\linewidth]{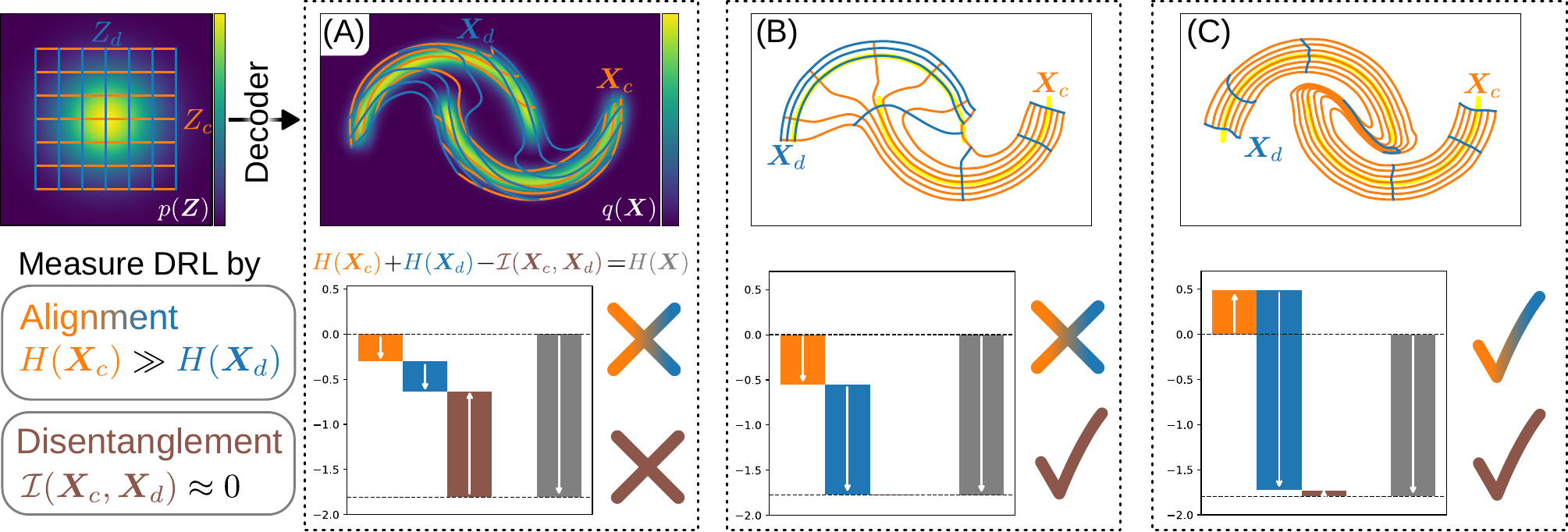}
    \caption{The two moons distribution illustrates how manifold entropic metrics quantify DRL in terms of alignment and disentanglement.
    (Top left) The latent prior and a Cartesian grid spanned by the latent variables $Z_c$ (orange) and $Z_d$ (blue).
    The latent distribution is mapped to data space by three generative models with equal accuracy, but vastly different representations.
    This can be seen by the differences in the transformed grid spanned by the manifold random variables $\X_c$ (orange) and $\X_d$ (blue) in the top row, and the corresponding values of our metrics {\em manifold entropy} $H(\X_c)$, $H(\X_d)$ and {\em manifold mutual information} $\mathcal{I}(\X_c, \X_d)$ in the bottom row.
    The total entropy of the distribution (gray) is the signed sum of the three terms.
    (A) The latent manifolds are entangled (and thus not interpretable), and our metric indicates this by high mutual information (brown).
    (B) The latent manifolds are locally orthogonal everywhere and have low mutual information.
    However, alignment is inconsistent ($\X_d$ aligns with the upper moon, $\X_c$ with the lower), resulting in comparable manifold entropy of both variables.
    (C) The representation is disentangled and aligned.
    The manifold entropy is high for the important variable $\X_c$ (orange) and low for the noise variable $\X_d$ (blue), and their mutual information is small.
    }
    \label{fig:Sketch of derivation}
\end{figure*}

{\em Disentanglement}, i.e. the statistical independence between latent factors, has been addressed by Independent Component Analysis \citep[ICA,][]{COMON1994287} under the assumption of a linear data-generating process (DGP), $\x = \A \boldsymbol{s}$.
When linearity holds, the true generative factors $\boldsymbol{s}$ are identifiable if they are independent and follow a non-Gaussian distribution.
However, identifiability is generally lost for non-linear DGPs $\x = \Phi(\boldsymbol{s})$, \citep{HYVARINEN1999429}.
Finding conditions on $\boldsymbol{s}$ to restore identifiability is a major focus of current research, see \citep{hyvarinen2023nonlinearindependentcomponentanalysis} for a recent survey.

Alternatively, one can restrict the class of permitted mixing functions $\Phi$, and this approach primarily inspired the present work.
Independent Mechanism Analysis \citep[IMA,][]{gresele2022independentmechanismanalysisnew} postulates, that the contributions ${\partial \Phi}/{\partial s_i}$ of each source $s_i$ to the mixing function $\Phi$ should not be fine-tuned to each other and thus independent. This ultimately leads to the condition that the Jacobian ${\partial \Phi}/{\partial s}$ should be orthogonal in every point.
Advantageously this approach circumvents a number of non-identifiability issues and doesn't impose non-Gaussianity on the components.
Principal Component Flows \citep{cunningham2022principal, pmlr-v162-cunningham22a} realize this idea by adding a loss term to normalizing flow training that encourages orthogonality.
It can be shown that orthogonality of the Jacobian is equivalent to minimizing the mutual information between the {\em image} of the corresponding features in data space {\em after} mapping them through the decoder (see \cref{fig:Sketch of derivation} and Appendix).
This is crucial: In contrast to supervised disentanglement metrics, which are usually defined with respect to the encoder, meaningful unsupervised metrics must be defined in terms of the {\em decoder} mapping from latent to data space.

Specifically, we make the following contributions:
\begin{itemize}
    \item We propose Manifold Entropic Metrics to evaluate Disentangled Representation Learning: \\A theoretical framework for information-theoretic measures of latent manifolds defined on the decoder of a generative model
    \item We introduce Alignment as an important complementary condition to Disentanglement in the Independent Mechanism Analysis framework  %
    \item We show the usefulness of our metrics at dissecting generative models in order to reveal and quantitatively measure their behaviours. %
\end{itemize}
}

\tocless\section{RELATED WORK}

Disentanglement has been mostly tackled using $\beta$-VAE \cite{higgins2017betavae} and its many successors.
Evaluating the learned representations in a supervised way is possible through a plethora of metrics such as
\cite{higgins2017betavae,
kim2019disentanglingfactorising,
kim2019relevancefactorvaelearning,
chen2019isolatingsourcesdisentanglementvariational,
kumar2018variationalinferencedisentangledlatent,
eastwood2018a,
ridgeway2018learningdeepdisentangledembeddings,
tokui2022disentanglementanalysispartialinformation,
reddy2021causallydisentangledrepresentations}.

\cite{HYVARINEN1999429} and later \cite{locatello2019challengingcommonassumptionsunsupervised} proved that pure unsupervised DRL is theoretically impossible without inductive bias on methods and datasets.
Comparing two latent representations in a linear fashion is possible using methods such as canonical-correlation analysis \cite{CCA} or more recently in \cite{duan2020unsupervisedmodelselectionvariational}.

Complementary to our work, \cite{pmlr-v235-buchholz24a} investigate theoretical conditions for unsupervised representation learning to succeed, whereas we propose metrics to quantify the actual success experimentally and to compare the quality of competing learned representations.

Most similar to ours, \cite{do2021theoryevaluationmetricslearning} introduce two information-based metrics on $\beta$-VAEs, \textit{Informativeness} and \textit{Separability}, which measure how much information from the data is captured in the latent representations and how well the latter are disentangled from one another.
But as their formulation solely relies on the encoder, they can only capture how data samples cause the latent representations to form but not necessarily the effect of the representations themselves.
We argue that this approach is somewhat ill-posed, for which we will show empirical evidence, but more importantly it lacks interpretation as these metrics are defined solely in the abstract representations space and not in the data space, which is of our interest.
Therefore it is more sensible to define metrics which quantify the effect a latent representation induces on generating new data, similar to the contributions of each source in IMA, thus relying on the decoder instead.

While DRL has been mostly tackled using VAEs, GANs and Diffusion-based approaches, Normalizing Flows (NF) incorporate a deterministic, tractable and bijective mapping from latent to data space, best suited for our approach.
It will allow us to formulate conditions on DRL via the decoder, which is trained implicitly during training through the encoder.
See \cite{Kobyzev_2021} for a summary on NFs.

\vspace{5pt}
\tocless\section{METHOD}
\vspace{-5pt}

Let us first formalize how achieving Density Estimation and Disentangled Representation Learning can be formulated in terms of the IMA principle:\\
A semantic feature vector $\boldsymbol{s}$ is sampled from a (simple) prior distribution $p(\boldsymbol{s})$ and pushed through an unknown \textbf{true} mixing function $\g^*$ to produce the data samples $\x$.
We assume that varying a single feature $\Delta s_i$ should only vary a single semantically meaningful and isolated variation in the data $\Delta \x_i$.
A successful DRL method recovers the semantic features by learning the distribution of data $p(\x)$ and restricting the decoder function $\g$ such that all latent representations $z_i=f_i(\x)$, computed by the encoder, match the semantic features $s_i$ up to a permutation.
We can split this task into two necessary conditions:
\begin{enumerate}
    \item[-] One latent variable $z_i$ should model the same semantic feature $s_m$ globally. 
    \item[-] Different latent variables $\{z_i,..., z_j\}$ should not model the same semantic feature $s_m$ locally.
\end{enumerate}

Even though we don't have access to the true semantic features, they can be locally estimated through the notion of principal components as the orthogonal directions of maximum variance around a data point, see e.g. \cite{pmlr-v162-cunningham22a}. %
One drawback of this local definition is that the meaning of a latent variable (i.e. the associated semantic feature) might change across the latent space, and the first condition will be violated.
This motivates us to derive global measures quantifying the extent to which both conditions are fulfilled in the unsupervised setting.

\tocless\subsection{Normalizing Flows}
Our entropic metrics require a trained generative model with a fixed tractable latent prior distribution and a corresponding decoder, which maps deterministically from latent to data space.
The following is based on Normalizing Flows (NF), 
but applies similarly to other generative models.
NFs transform data samples to the latent space via the encoder
\begin{equation} \label{eq: NF encoder}
    \z = \f(\x)
\end{equation}
and latent samples back to data space via the decoder
\begin{equation} \label{eq: NF decoder}
    \x = \f^{-1}(\z) \eqqcolon \g(\z)
\end{equation}
where $\g$ is defined as the exact inverse of $\f$.
The latent prior is usually assumed to be a standard normal distribution with a pdf
\begin{equation} %
    p(\Z=\z) = \mathcal{N}(\z\,|\, 0, \I_D)
\end{equation}
To generate synthetic data points, one samples from the latent prior and pushes the instances to data space through the decoder.

To specify subsets of the latent variables, we use index sets $\set{S}\subseteq \{1,...,D\}$ with complement $\nset{S}$. 
A corresponding split of the latent vector $\z$ is expressed as $\z=[\z_\set{S}, \z_\nset{S}]$.
The Jacobian of the decoder at point $\z$ is $\J(\z) \coloneqq \frac{\partial \g(\z')}{\partial \z'}\big|_{\z'=\z}$ and the submatrix $\J_\set{S}(\z)$ contains only the columns of $\J(\z)$ with indices in $\set{S}$.

\tocless\subsection{Derivation recipe of our metrics}
The derivation of our manifold entropic metrics can be summarized by the following 5 steps:
\begin{enumerate}
    \item We define a \textbf{latent manifold} $\Man_\set{S}$ as the set of all points generated by the decoder when only the latent variables with index set $\set{S}\subseteq \{1,...,D\}$ is varied, while the remaining ones are kept fixed.
    $\Man_\set{S}$ is the image of $\Z_\set{S}$ through the decoder while $\z_\nset{S}$ remains fixed:
    \begin{equation}
        \Man_\set{S}(\z_\nset{S}) =\big\{ \x=\g([\z_\set{S}, \z_\nset{S}]): \z_\set{S}\in \mathbb{R}^{|\set{S}|}\big\}
    \end{equation}
    
    \item We define the \textbf{manifold random variable} $\X_\set{S}$ as the transformation of the latent random variable $\Z_\set{S}$ onto its latent manifold $\Man_\set{S}$ as:
    \begin{equation} %
    \X_\set{S} \coloneqq \g(\left[\Z_\set{S}, \z_\nset{S}\right]) 
    \end{equation}
    
    \item We define the \textbf{manifold pdf} $q_\set{S}(\X_\set{S})$ as the pdf of the manifold random variable induced via the change-of-variables formula on a latent manifold. For this we use the injective change-of-variables formula \citep{köthe2023review} at $\x_\set{S}=\g([\z_\set{S}, \z_\nset{S}])$ (with $\z_\nset{S}$ fixed)
    \begin{equation} \label{eq:manifold-pdf}
        q_\set{S}\big(\X_\set{S}\big) 
        = p_\set{S}\big(\Z_\set{S} = \z_\set{S}\big) \big| \J_\set{S}(\z) \big|^{-1}
    \end{equation}
    where $|.|$ denotes the volume of a squared or rectangular matrix $\boldsymbol{A} \in \mathbb{R}^{n \times m}$ with $n\ge m$ according to $|\boldsymbol{A}| \coloneqq \det(\boldsymbol{A}^T\boldsymbol{A})^\frac{1}{2}$. 
    It is easy to see that this formula reduces to the usual change-of-variables formula for NFs by setting $\set{S}=\{1,...,D\}$.
    
    \item We define the \textbf{manifold entropy} $H(q_\set{S})$ as the differential entropy of a manifold random variable via its manifold pdf.
    \item We define the \textbf{manifold mutual information} $\mathcal{I}(q_\set{S}, q_\set{T})$ via the manifold entropy of disjoint index sets $\set{S}$ and $\set{T}$.
\end{enumerate}

These concepts are depicted in \cref{fig:Sketch of derivation} and the formal definitions with more rigorous explanations are provided in the Appendix \ref{sec:METHOD}.

\obso{
Ultimately this novel formulation quantifies the amount of, total or shared, information a set of latent variables induce in the data.
As latent manifolds can be defined for any index set of latent dimensions, we can address both \textit{dimension-wise} and \textit{vector-wise} DRL methods, and as we can additionally define complementary metrics, like a conditional entropy, we are in principle able to address both \textit{flat} and \textit{hierarchical} DRL methods (see \citep{wang2024disentangledrepresentationlearning}).
}
\tocless\subsection{Alignment and Disentanglement} %
To formulate conditions on DRL in an unsupervised setting we use our entropic metrics and make two assumptions about the (unknown) DGP:
\begin{enumerate}
    \item[a)] The importance of different semantic features varies greatly $\Leftrightarrow$ The manifold entropy of corresponding latent manifolds shall vary accordingly\footnote{We can always resort the latent space, thus the ordering is generally not important.} %
    $\Rightarrow$ \textit{Alignment}
    \item[b)] Semantic features mostly model independent variations in the data $\Leftrightarrow$ The manifold mutual information between latent manifolds of disjoint index sets shall be small 
    $\Rightarrow$ \textit{Disentanglement}
\end{enumerate}
Using this formulation we can quantify the degree to which a generative model achieves DRL.

\obso{
We can deduce two additional conditions if the DGP respects the manifold hypothesis:
\begin{enumerate}
    \item[3.] There is an intrinsic dimensionality $d$ of the data $\Leftrightarrow$ The manifold entropy is non-vanishing for exactly $d$ dimensions
    \item[4.] The contribution of noise is small and independent of the data $\Leftrightarrow$ The manifold entropy of $D-d$ dimensions is vanishingly small and the manifold mutual information between any pair of index sets vanishes
\end{enumerate}

We note that although DE and DRL are complementary, in practice they are not necessarily independent as models often exhibit a reconstruction-disentanglement trade-off (see \cite{burgess2018understandingdisentanglingbetavae}).
}

\tocless\subsection{Interpretation of PCA via manifold entropy and mutual information} \label{main sec:Interpretation via PCA}

In order to gain some intuition of the \textbf{manifold entropy} and the \textbf{manifold mutual information}, we look at a toy example and assume that the data is a zero-mean Gaussian.
Then, decoder and encoder of a NF are just linear transformations defined by the eigendecomposition of the covariance matrix, i.e. identical to  Principal Component Analysis (PCA).
In this interpretation, PCA is a generative model achieving both Density Estimation (since the volume change in (\ref{eq:manifold-pdf}) equals the square root of the determinant of the data covariance matrix) and Disentangled Representation Learning (since the latent dimensions after whitening are statistically independent).
Specifically, the PCA eigenvectors a) represent the directions of maximal variation in the data and b) are orthogonal to each other. 
The eigenvalues represent the squared width of the data distribution along the corresponding eigendirection.

These properties can be equivalently quantified via the \textbf{manifold entropy} and the \textbf{manifold mutual information} using a) \textit{Alignment} and b) \textit{Disentanglement}:
The latent manifolds of a Gaussian data distribution simply become linear subspaces of $\mathbb{R}^D$, and the manifold pdfs are also Gaussian (of lower dimensions).
Furthermore, the manifold entropies are proportional to the log-determinant of the covariance matrix of the respective manifold pdf. 
Most notably, in the case of $\set{S}=i$ this simplifies further to the log-variance of a one-dimensional manifold pdf. 
Thus, we can interpret the 1D manifold entropy as the logarithm of the average length of the manifold pdf (for example, the total length of both moons in \cref{fig:Sketch of derivation} (C) is absorbed into $\X_c$).
Furthermore, since all higher moments of a Gaussian are zero, the manifold mutual information between two index sets becomes a measure of the orthogonality of the two latent manifolds, analogously to a cosine similarity (see \ref{sec:Geometric perspective of the Manifold Pairwise Mutual Information}). 
The manifold mutual information is zero only if the respective subspaces are orthogonal to each other, which is precisely the solution of PCA.

This procedure ultimately allows us to link \textit{local} geometrical attributes like lengths and angles to \textit{global} information-theoretic metrics via the differential entropy and mutual information.
Whereas geometric measures are only defined \textit{locally}, i.e. at a single data instance, our manifold entropic metrics are a natural generalization to an interpretable \textit{globally}-defined measure, i.e. on the entire distribution.

A full derivation is provided in the Appendix \ref{sec:INTERPRETATION VIA PCA}.

\obso{
\section{Related work on Regularizing NFs to achieve DE and DRL}

Very close to our work, \cite{pmlr-v162-cunningham22a} tackle Disentanglement but disregard the issue of Alignment.
\cite{flouris2023canonicalnormalizingflowsmanifold} follow a somewhat similar approach at Disentanglement by instead regularizing the off-diagonal terms of the volume-change matrix through an ad-hoc loss.
\cite{ross2021tractable} introduce a NF-architecture via Conformal Mappings which achieves Disentanglement natively but strongly lacks expressivity.
Similarly \cite{cramer2023nonlinear} show that Isometric Autoencoders from \cite{gropp2020isometricautoencoders} can be used to compose an injective NF-architecture with a tractable training objective, though they don't use the disentangled properties of the latter.
Inspired by the training objective of Autoencoders, Alignment can be achieved by an additional reconstruction loss which regularizes the L2-distance from a data-point to its projection onto the core latent manifold. This allows one to absorb the most important variations in the data into the core part. \cite{rippel2014learningorderedrepresentationsnested} and \cite{bekasov2020orderingdimensionsnesteddropout} achieve a hierarchical sorting of latent variables through Nested Dropout, an extension of the reconstruction loss.
}

\newpage

\tocless\section{MANIFOLD ENTROPIC METRICS}
We present our manifold entropic metrics which are used to evaluate how well a generative model performs at Disentangled Representation Learning.
Our framework allows the derivation of a wide range of entropic metrics tailored to specific tasks and for any index set $\set{S} \in \mathcal{P}$, where $\mathcal{P}$ is a partition over $D$.
Though in an attempt to simplify notation and because we often evaluate the effect of a single latent dimension $z_i$, we introduce the metrics using a partition size of {\em one} for all latent manifolds: $\mathcal{P} \equiv \{1,...,D\}$ with $\set{S} \equiv i$.
For notation $\J_{i}$ denotes the $i$-th column vector of $\J$ and $\J_{\{ij\}}$ the concatenation of $\J_i$ and $\J_j$.

\begin{enumerate}[wide]%
\item The \textbf{Total Entropy} $H(q) \in \mathbb{R}$ is the differential entropy\footnote{For clarity we omit writing \textit{differential} as we only define metrics on continuous distributions.} of the generated data $\X=\g(\Z)$: \vspace{-5pt}
    \begin{equation} \label{eq:Total Entropy}
        H(q) = \frac{D}{2}\left(1+ \log(2\pi)\right) + \Expt{\z}{\log\left|\J(\z)\right| } %
    \end{equation}
    \begin{enumerate}[wide, label={}]%
    \item %
    This is the information induced by the (full) generative pdf $q(\X)$.
    By itself it is not a useful metric.
    \end{enumerate}
    \item The \textbf{Manifold Entropy} (ME) $H(q_i) \in \mathbb{R}^{D}$ is the marginal entropy of the \textit{manifold random variable} $\X_i$: \vspace{-5pt}
    \begin{equation} \label{eq:Manifold Entropy}
        H(q_i) = \frac{1}{2}\left(1+\log(2\pi)\right) + \Expt{\z}{\log\left|\J_i(\z)\right| } %
    \end{equation}
    \begin{enumerate}[wide, label={}] %
    \item %
    This metric denotes the amount of information induced by the \textit{manifold pdf} $q_i$.
    It gives a notion of importance to each latent dimension as it is generalized from the standard deviation or equivalently the mean L2-distance of the reconstruction error.
    It is important to note that the expectation is performed over the full latent space $\z \sim p(\z)$ and not just over $z_i$.
    \item %
    With it we can sort the latent manifolds by their magnitude and visualize the resulting spectrum.
    Ultimately it can be used to asses the \textit{Alignment} of a generative model.
    \end{enumerate}

\item The \textbf{Manifold Total Correlation} (MTC) $\mathcal{I} \in \mathbb{R}$ is the total correlation between all \textit{manifold random variables} $\{\X_i\}_{i=1}^D$: \vspace{-5pt}
    \begin{equation} \label{eq: Manifold Total Correlation}
        \mathcal{I} = \sum_{i=1}^{D} H(q_i) - H(q)
    \end{equation}
    \begin{enumerate}[wide, label={}] %
    \item %
    This metric denotes the amount of shared information between all \textit{manifold pdfs} $\{q_i\}_{i=1}^D$ and indicates how entangled the latent manifolds are.
    \item %
    It is identical to $\mathcal{I}_{\mathcal{P}}$ in \cite{pmlr-v162-cunningham22a} and the global IMA contrast from \cite{gresele2022independentmechanismanalysisnew}.
    It is Non-negative (see e.g. \cite{pmlr-v162-cunningham22a} claim 5) and becomes zero only if the full pdf can be factorized into $D$ manifold pdfs. 
    \item %
    Ultimately it can be used to assess the \textit{Disentanglement} of a generative model.
    \end{enumerate}

\item The \textbf{Manifold Pairwise Mutual Information} (MPMI) $\mathcal{I}_{ij}(q_i,q_j) \in \mathbb{R}^{D\times D}$ is the mutual information between a pair of \textit{manifold random variables} $\X_i$ and $\X_j$: \vspace{-5pt}
    \begin{equation}
    \begin{split}
    &\mathcal{I}_{ij}(q_i,q_j) =\\
    &\Expt{\z}{\log\left|\J_i(\z)\right| + \log\left|\J_j(\z)\right| - \log\left|\J_{\{i,j\}}(\z)\right|}
    \end{split}
    \end{equation}
    \begin{enumerate}[wide, label={}] %
    \item %
    This metric denotes the amount of shared information between two \textit{manifold pdfs} $q_i$ and $q_j$. %
    It indicates if two latent manifolds are globally entangled or equivalently correlated with each other as it is a non-linear analogue of the classical Pearson correlation coefficient.
    \item %
    It is identical to $\mathcal{I}_{\set{S},\set{T}}$ from \cite{pmlr-v162-cunningham22a}, with $\set{S}=i$ and $\set{T}=j$.
    The resulting matrix is symmetric where the diagonal elements are undefined and an entry becomes zero only if two latent manifolds cross orthogonally everywhere.
    \item %
    With it we can asses the dimension-wise disentanglement of a generative model.
    It is useful to sort the latent dimensions by magnitude of the ME first to discard unimportant ones.
    Then one can identify latent dimensions which might model similar features, i.e. they are entangled, or ones which are isolated from all other, i.e. disentangled.
    \end{enumerate}
    
\item The \textbf{Manifold Cross-Pairwise Mutual Information} (MCPMI) $\mathcal{I}^{ab}_{ij}(q^a_i,q^b_j) \in \mathbb{R}^{D\times D}$ is the mutual information between a pair of \textit{manifold random variables} where each stems from a different generative model with decoders $\g^a$ and $\g^b$ respectively: \vspace{-5pt}
    \begin{equation} \label{eq:MCPMI}
    \begin{split}
    &\mathcal{I}^{ab}_{ij}(q^a_i,q^b_j) =\\
    &\Expt{\z}{\log\left|\J^a_i(\z)\right| + \log\left|\J^b_j(\z)\right| - \log\left|\J^{ab}_{\{ij\}}(\z)\right|}
    \end{split}
    \end{equation}
    \begin{enumerate}[wide, label={}] %
    \item %
    This metric denotes the amount of shared information between two \textit{manifold pdfs} $q^a_i$ and $q^b_j$, indicating how much two latent manifolds from two different models correlate with each other.
    It is computed similarly as the MPMI, while the Jacobian column vectors $\J^a_i$ and $\J^b_j$ stem from the respective models.
    \item %
    All entries in the matrix are unique.
    If two latent manifolds model the same feature, the corresponding entry will tend to infinity.
    If all latent manifolds from both models coincide with each other, and the dimensions are pre-sorted, only the diagonal will be non-zero.
    \item %
    With it we can asses the dimension-wise correlations between two decoders and infer to which degree they are similar.
    \end{enumerate}

\end{enumerate}

\newpage

\tocless\section{NUMERICS}
The core computation of our method is the calculation of Jacobian matrices $\J(\z)$ of the decoder.
This used to be very expensive, but progress in automatic differentiation has made these computations much cheaper.
Modern libraries such as pytorch and tensorflow provide a vector-Jacobian product (vjp) primitive based on efficient reverse-mode auto-diffenentiation that allows us to compute Jacobians one column at a time:

\noindent For $k \in \{1,..., D\}$ do:\\
\phantom{MM} $\g(\z), \J_{k, \cdot} = \texttt{vjp}(\g, \z, \boldsymbol{e}_k)$

where $\boldsymbol{e}_k$ is the standard basis vector for index $k$.
Our actual pytorch implementation is shown in \ref{sec:Python Implementation}.
This computation has to be done only once per instance.
All required submatrices $\J_\set{S}(\z)$ are then easily obtained by slicing, followed by standard matrix computations to get the determinant.
The case of a single column is especially simple, because $|\J_i(\z)| = ||\J_i(\z)||_2$.
The expectation over $\z\sim p(\z)$ in our metrics is approximated by an empirical average, and 100 to 1000 instances turned out to give sufficient accuracy.

It is noteworthy to add that exact determinant computations are not necessarily required during inference. Specifically:
For the Manifold Entropy over a single latent dimension (see eq. (7)), which is arguably the most useful metric as it quantifies the importance of individual latent dimensions, the determinant computation simplifies significantly. Here, the Jacobian submatrix $\J_i$ reduces to a column vector, and its determinant degenerates to the L2-norm. Thus, computing the Manifold Entropy reduces to taking expectations over vector norms, which is computationally efficient. In practice, we found this step to require negligible time compared to the Jacobian computation itself.
For the Total Entropy (see \cref{eq:Total Entropy}), most NF architectures are designed to make full Jacobian determinant computations tractable and fast, effectively providing this metric "for free". Using \cref{eq: Manifold Total Correlation}, we can then directly obtain the MTC to assess model disentanglement.

In summary, computing the Jacobians $\J(\z)$, the total entropy $H$ and the manifold entropies $H_i$ of a trained NF using 1000 MNIST-sized images ($D=784 = 1 \times 28 \times 28$) takes about $10$s on a NVIDIA RTX 4090 GPU. 
Therefore, exact test-time evaluation of $\J(\z)$ and subsequently of our metrics is feasible for high-dimensional problems.%

The code for computing the metrics and the proof-of-concept examples are available on GitHub.\footnote{\url{https://github.com/DanielGalperin/ManifoldEntropicMetrics}}

\vfill

\tocless\section{PROOF OF CONCEPT}

\paragraph{Two Moons:} For \cref{fig:Sketch of derivation}, we created 2-dimensional data with the function \texttt{sklearn.make\_moons} 
and Gaussian noise variance $0.01$.
All models use the same RQS-NF architecture \citep{durkan2019neural}.
Panel (A) is the result of Maximum Likelihood (ML) training alone. In (B), we train with ML and additionally minimize the manifold total correlation $\mathcal{I}$ as in \cite{pmlr-v162-cunningham22a}. In (C) we train with ML and additionally minimize the reconstruction loss $\mathcal{L}_\text{rec}$ (see Experiments) as in \cite{NEURIPS2021_4c07fe24}. See Appendix for more details.
The latent manifolds of $\X_c$ and $\X_d$ are visualized by equidistant sampling from $z_c$ and $z_d$ in latent space and transforming the resulting grid to data space.

\begin{figure*}[!htbp]
    \centering
    \begin{subfigure}[t]{.47\linewidth}
        \centering
        \includegraphics[width=\linewidth]{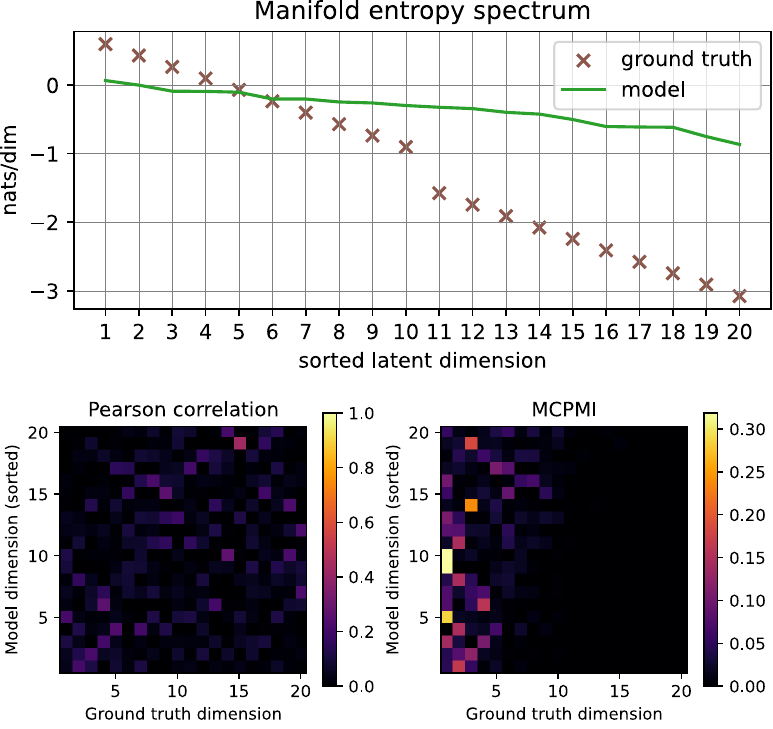}
        \caption{Vanilla Maximum Likelihood training}
        \label{fig:1}
    \end{subfigure}
    \begin{subfigure}[t]{.47\linewidth}
        \centering
        \includegraphics[width=\linewidth]{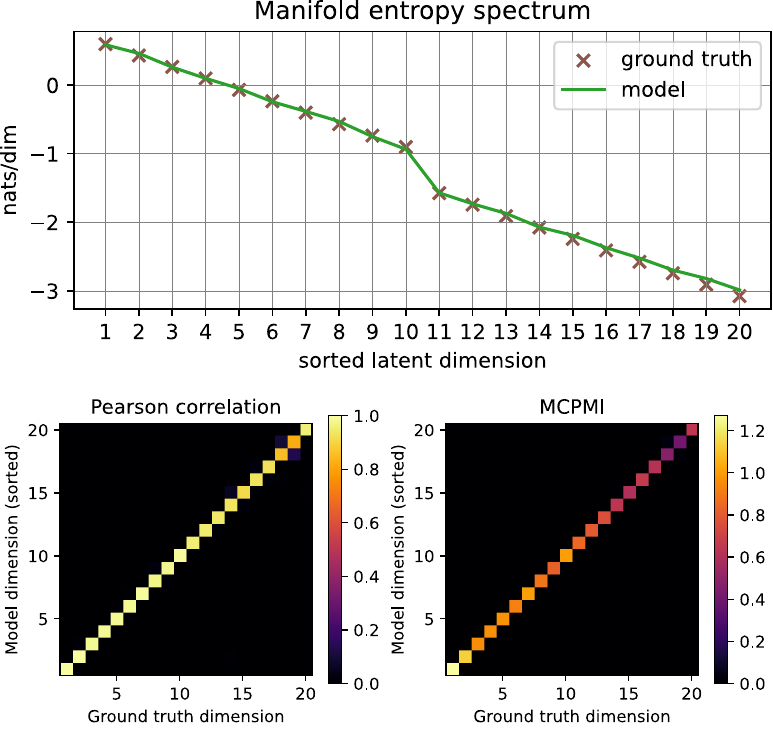}
        \caption{Additional disentanglement regularization}
        \label{fig:2}
    \end{subfigure}
    \caption{Application of our manifold entropic metrics in order to infer if the ground-truth DGP has been learned (b) or not (a). (Top) Manifold entropy spectrum and (Bottom) Pearson Correlation matrix and MCPMI matrix (left to right) comparing a trained model with the ground truth. As expected, statistical independence between latent variables (diagonal MCPMI) implies uncorrelated features (diagonal Pearson).}
    \label{fig:toy-data entropic metrics}
\end{figure*}

\paragraph{10-D Torus:} The manifold $\Man$ is a 10-dimensional torus embedded in a 20-dimensional Euclidean space. 
Following the manifold hypothesis, we sample data on the surface of the torus plus Gaussian noise.
Successful representation learning should discover the polar representation in terms of azimuthal angles and radii from a training dataset in Cartesian coordinates.
We design the data generating process such that the manifold entropy in the azimuthal variables $\ph$ is higher than in the radial variables $\r$, i.e. the noise is small relative to the size of the torus.
Thus, the azimuthal variables should be associated with the important (\textit{core}) subset $\z_\set{C}$ of the latent space (with $\set{C} =\{0, ..., 9\}$), and the radial variables are the unimportant details to be associated with $\z_\nset{C}$.
Precisely, we define for $i=0,...,19$ 
\begin{equation}
    x_i = \begin{cases}
    r_{\lfloor i/2 \rfloor}\cos(\varphi_{\lfloor i/2 \rfloor}),& \text{if } i \text{ is even}\\
    r_{\lfloor i/2 \rfloor}\sin(\varphi_{\lfloor i/2 \rfloor}),& \text{if } i \text{ is odd}
    \end{cases}
\end{equation}
and sample $\varphi$ and $r$ by reparameterization in terms of $\z\sim \mathcal{N}(0, \I_{20})$:
\begin{align}
    \varphi_{\lfloor i/2 \rfloor} &= \sig^{(\varphi)}_{\lfloor i/2 \rfloor} \cdot  \z_{\lfloor i/2 \rfloor} \\ 
    r_{\lfloor i/2 \rfloor} &= 1 + \sig^{(r)}_{\lfloor i/2 \rfloor} \cdot \z_{10+\lfloor i/2 \rfloor}
\end{align}
The standard deviations are monotonically decreasing, so that the true latent space has a unique importance ordering which DRL should recover:
\begin{align}
    \sig^{(\varphi)} &= 0.07 \cdot 2 \pi \cdot \exp(-[0,...,1.5]) \\ \sig^{(r)} &= 0.05 \cdot \exp(-[0,...,1.5])
\end{align}
To discourage the NF from aligning with the data manifold accidentally, we additionally apply a random fixed rotation and normalization to the Cartesian dataset prior to training.

The crosses in \cref{fig:toy-data entropic metrics} show the true manifold entropy (which can be calculated analytically and is proportional to $\log(\sigma_i)$) of the latent variables according to this ordering.
Alignment is achieved if the model correctly identifies this ordering, and in particular the subsets $\z_\set{C}$ and $\z_\nset{C}$.
Disentanglement is achieved when the matrix of pairwise manifold mutual information is diagonal, and the manifold total correlation vanishes.
We train two identical NFs on this dataset, one solely with the Maximum Likelihood objective and one with an additional regularization term to encourage Disentanglement by minimizing the estimated manifold total correlation following the PCF objective in \citep{pmlr-v162-cunningham22a}.
\obso{
To inspect visually if the data density has been learned correctly, one could generate data samples and plot the projection onto two data dimensions which is infeasible as there are $\frac{D\cdot(D-1)}{2}$ pairs of data-dimensions.
To inspect visually if the decoder inverts the DGP exactly, on would have to visualize the contour grids for all $\frac{D\cdot(D-1)}{2}$ pairs of latent-dimensions, thus a staggering $\frac{D^2\cdot(D-1)^2}{4}$ plots in total.
}

We check for both models if the DGP has been recovered by evaluating our manifold entropic metrics, \cref{fig:toy-data entropic metrics}.
Sorting latent dimensions by their Manifold Entropy (ME) shows that pure ML training does not produce meaningful alignment as the spectrum is almost flat and doesn't coincide with the ground-truth.
With additional disentanglement regularization the spectrum exactly follows the ground-truth.
To see if the learned representations match the ground-truth, we compute the manifold cross-pairwise mutual information (MCPMI) matrix between the trained decoder and the ground truth decoder, plus the classical Pearson correlation matrix for comparison (see \ref{sec:Pearson Cross-Correlation matrix}).
The matrices of the vanilla model have no visible structure, whereas the matrices of the regularized model are diagonal (bottom row).
Thus, only the regularized model has successfully learned the true DGP.

\tocless\section{EXPERIMENTS}
Here we will conduct experiments on the digits subclass of the EMNIST dataset \cite{cohen2017emnistextensionmnisthandwritten}. It consists of 240.000 training examples of hand-written digits in black and white with a resolution of $1 \times 28 \times 28$, hence $D=784$.

We will train both NFs and $\beta$-VAEs to evaluate their DRL capabilities using our manifold entropic metrics.
We train NFs with the Maximum Likelihood-objective\footnote{By minimizing the negative log-likelihood loss} and reveal that the right architecture, which is biased for the image domain, can achieve surprisingly well disentangled and meaningful representations.
Note that currently there is no tractable method to regularize disentanglement on high-dimensional data directly by minimizing the manifold total correlation.
\cite{pmlr-v162-cunningham22a} mostly provide approximate objectives with a main application to injective NFs, where the latent space is smaller than the data space. %

\begin{figure}[!htbp]
    \centering
    \includegraphics[width=.97\linewidth]{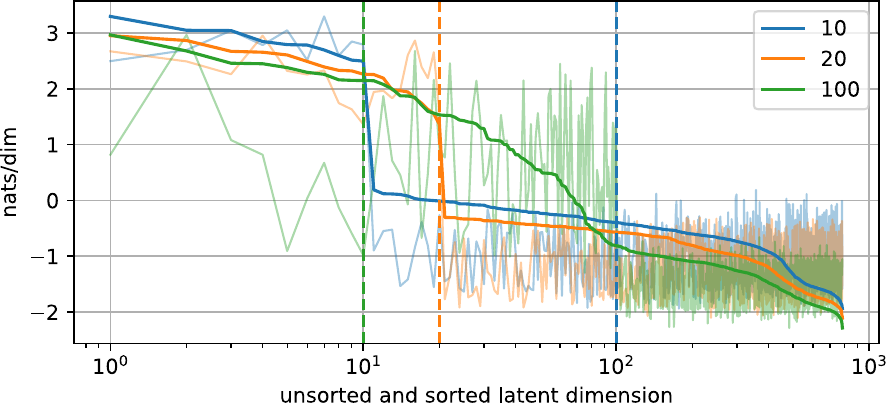}
    \caption{Manifold Entropy spectrum of NFs with unsorted (faded color) and sorted (full color) latent dimensions. Trained with additional reconstruction loss for $10$, $20$ and $100$ core dimensions $\set{C}$.}
    \label{fig:H_rec}
\end{figure}

\paragraph{Normalizing Flows:}
To showcase the effects of additional regularization, we train NFs with an additional reconstruction loss $\mathcal{L}_\text{rec}$ with $10$, $20$ and $100$ reconstructed dimensions.
For this we divide the latent space into a \textit{core} index set $\set{C}$, and its complementary index set $\nset{C}$.
The loss minimizes the squared distance between a data sample and its reconstruction after projection onto $\z_\set{C}$.
This approach is a naive application of an Autoencoder and similar to how many works enforce Injective NFs \citep{kothari2021trumpets, brehmer2020flows, caterini2021rectangular} to absorb the most important variations in the data into the core part.
In the resulting manifold entropy spectrum \cref{fig:H_rec} we can see a (sharp) knee after the first $10$, $20$ and $100$ latent dimensions respectively, i.e. the additional regularization induces alignment between the core and detail part.
Additionally the MPMI matrix reveals that entries in the region relating $\set{C}$ and $\nset{C}$ are less active, i.e. the additional regularization induces disentanglement between the two latent manifolds.
\obso{
Unfortunately the training is unstable and quickly diverges, as the reconstruction loss will generally conflict with Maximum Likelihood training as noted in \cite{cramer2023nonlinear}. We could only train for 10 epochs with small weighting, thus this is not a promising way to achieve Alignment.
}

\begin{figure}[!htbp]
    \centering
    \includegraphics[width=.97\linewidth]{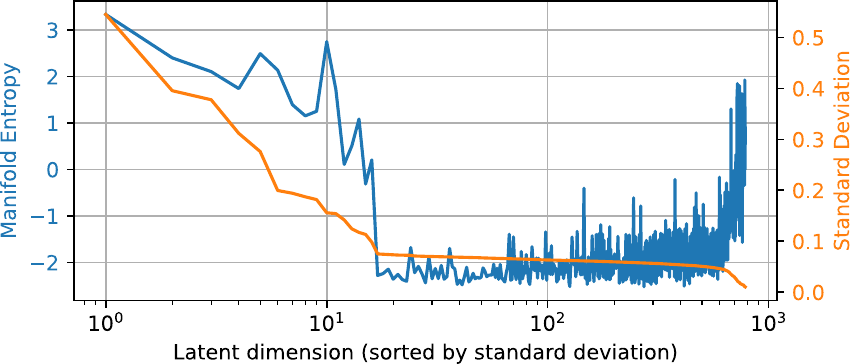}
    \caption{Sorting latent dimensions by the standard deviation of each latent variable as in GIN can reveal a spurious notion of importance.}
    \label{fig:GIN sorting spurious}
\end{figure}

We train a GIN-model as in \cite{Sorrenson2020Disentanglement}, which employed an NF-architecture based on GLOW \citep{kingma2018glow} (adapted to ensure that the NF is volume-preserving) and the Maximum Likelihood objective.
Here the latent space can not be treated as a fixed prior but is learned as a Gaussian Mixture Model.
Samples from each digit (0-9) correspond to a single mixture component where the mean and the diagonal covariance matrices are estimated jointly while training.
By estimating the variance of each latent dimension averaged over all components and sorting by the magnitude, the authors reveal a sudden decrease in the spectrum after 22 latent dimensions.
We replicate their findings and can observe an apparent flattening in the resulting spectrum after $\sim 20$ dimensions as well.
Surprisingly by instead sorting using the manifold entropy we find dimensions with a negligible latent variance but they nevertheless exhibit meaningful variations in the data space.
Furthermore we observe a gradual decrease in the manifold entropy spectrum, a more reasonable result also revealed by classical methods like PCA.
Observing the spectra in \cref{fig:GIN sorting spurious} we can conclude that measuring the variance of a latent variable only weakly correlates with the manifold entropy and is thus not a direct measure of importance in the data space.
This exemplifies that DRL metrics defined in terms of the encoder can lead to spurious results, contrary to decoder-based metrics.

\obso{
To help with interpretability we additionally trained cINNs \citep{ardizzone2019guided}, NFs which are conditioned on the digit label, allowing the latent space to decouple the digit label from its learned attributes.
Training on a GLOW-like architecture, similar to that of GIN, we find interesting and interpretable manifold directions in the data. In \cref{fig:20 EMNIST Eigenvectors} we visualize the mean gradient norm of the 20 most important latent manifolds, which is similar to eigenvectors revealed by PCA.
There, one can observe a peculiar latent manifold at the $18$-th place, which which was previously not found.
We suspect that this latent manifold, which we found consistently in many trained models, shows an artefact of data preparation in the creation of EMNIST.

\begin{figure*}
    \centering
    \includegraphics[width=\textwidth]{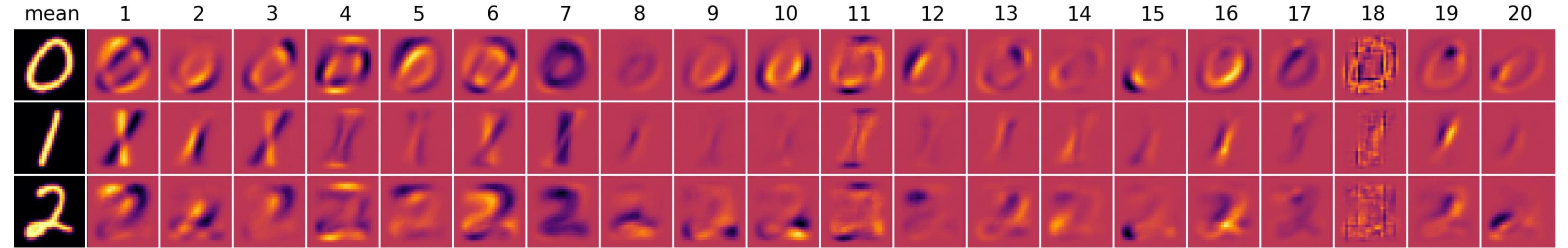}
    \caption{Depiction of the 20 most important latent manifolds of EMNIST visualized via the mean gradient norms of each latent dimensions. The NF is conditioned on each digit label, where only three digits are shown.}
    \label{fig:20 EMNIST Eigenvectors}
\end{figure*}
}
We can also use the entropic metrics to infer biases in particular architectures.
This might be useful in order to check if a given model is a good fit for a certain problem.
\obso{
As we analyze images it is useful to use NF-architectures which employ convolutional layers.
In an effort to increase expressivity of NFs, some works introduced orthogonal layers which act as a learnable or fixed rotation matrix between the individual layers.
We show that using them can, in some cases, be detrimental to DRL resulting in a strongly entangled and not aligned latent space (see Appendix).
}
As an example we examine a Wavelet-Flow \citep{yu2020wavelet}.
In contrast to GLOW, it uses a hierarchical architecture with fixed invertible Haar-wavelet downsamplings to encourage faster training.
The Haar transform splits the input into low- and high-frequency components.
The latter are directly transformed to a standard normal, conditioned on the former.
The low-frequency part is forwarded to the next layer, and the process is repeated recursively until the downsampled data can be handled by a low-dimensional unconditional NF.
For EMNIST we use a Wavelet-flow with two downsamplings resulting in three ascending index sets $\set{C}=\{1,...,49\}$, $\set{M}=\{50,...,196\}$ and $\set{D}=\{197,...,784\}$, which denote the post-processed latent space part from coarsest to finest.

\begin{figure*}[!htbp]
    \centering
    \includegraphics[width=.42\linewidth]{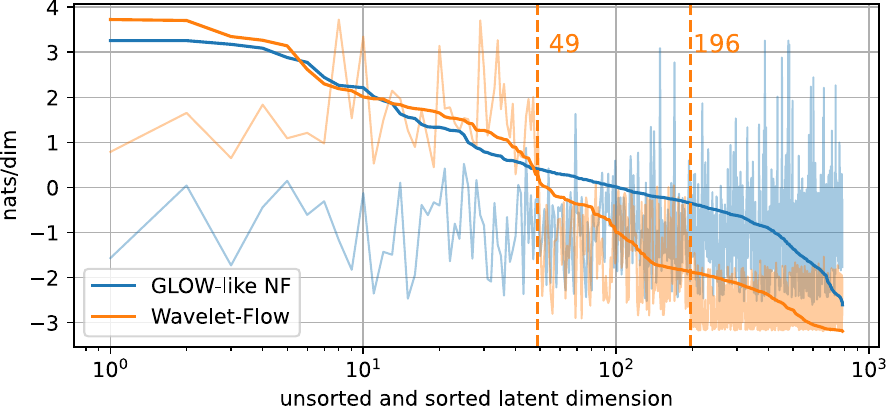}
    \hspace{5pt}
    \includegraphics[width=.25\linewidth]{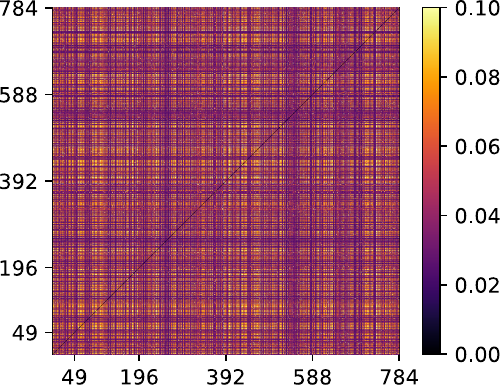}
    \hspace{5pt}
    \includegraphics[width=.25\linewidth]{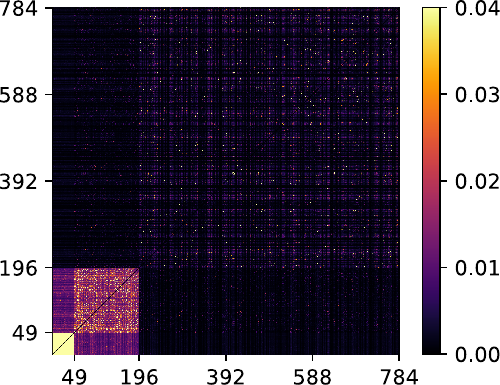}
    \caption{Manifold Entropy Spectra (left) with unsorted (faded color) and sorted (full color) latent dimensions, MPMI matrix of a trained GLOW-like NF (middle) and MPMI matrix of a Wavelet-flow (right).}
    \label{fig:MPMI conv-NF wavelet comparison}
\end{figure*}

In order to show the behaviour of a Wavelet-flow architecture compared to a GLOW-like one (from above), we plot the ME Spectrum and the MPMI matrix \cref{fig:MPMI conv-NF wavelet comparison} of both trained models.
There we can observe that information is mostly concentrated in the coarser set $\set{C}$ with significantly lower contributions from $\set{M}$ and $\set{D}$.
This makes sense, as the coarse set can model low-frequency variations and the finer ones only higher-frequency variations, which are negligible for EMNIST-images.
In the MPMI the hierarchical structure of the Wavelet-Flow can be directly observed, as correlations are more pronounced between dimensions in the coarse set $\set{C}$ than in the finer parts $\set{M}$ and $\set{D}$.
Additionally the manifold total correlation of the wavelet-flow is much lower, $\mathcal{I} \approx 0.5$, than that of the GLOW-like NF, $\mathcal{I} \approx 1.9$.
Ultimately we see can say a wavelet-Flow doesn't solve alignment or disentanglement entirely, but shows that choosing the right architecture can be very beneficial in achieving DRL.

{\bf $\beta$-VAEs:}
Lastly, to show that our approach can be applied to generative models other than Normalizing Flows, we train and analyze $\beta$-VAEs, introduced in \cite{higgins2017betavae}.
For that we need to treat the probabilistic decoder of an VAE $q(\x|\z)$ in a deterministic fashion by omitting the variance of the generated data point.
This is common practice (e.g. \cite{reizinger2023embracegapvaesperform}) and doesn't affect the interpretation as the variance part of the decoder only adds Gaussian noise onto the generated data samples.

\begin{figure}[!htbp]
    \centering
    \includegraphics[width=.97\linewidth]{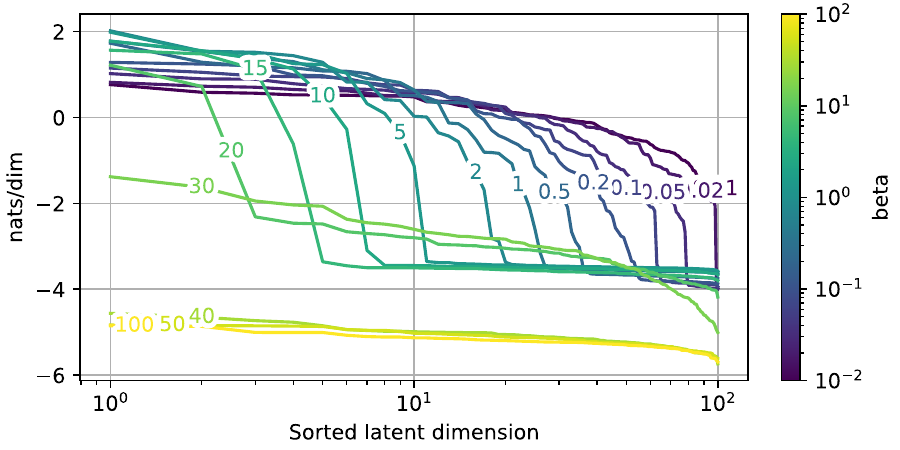}\ %
    \caption{Manifold entropy spectra for different trained $\beta$-VAEs with a latent space size of 100. The latent space is being compressed more with increasing $\beta$-value and becomes uninformative for $\beta>20$.}
    \label{fig:beta-VAE manifold-entropy}
\end{figure}

We trained $\beta$-VAEs on EMNIST with $\beta$ ranging from $0.01$ to $100$ for models with $100$ (and $10$) latent dimensions and evaluated the deterministic decoder by our manifold entropic metrics.
Most strikingly we observe a very sharp knee in the manifold entropy spectrum at a specific cutoff-dimension \cref{fig:beta-VAE manifold-entropy}.
This cutoff divides the latent space into a core and a detail part, where the former only models noise.
For increasing $\beta$-values the cutoff decreases and the manifold entropy of the core part increases, indicating that the latent space is being compressed more.
This coincides with the findings observed in \cite{do2021theoryevaluationmetricslearning} on the Informativeness of latent variables.
For $\beta$-values exceeding a certain threshold, here $\sim 20$, the training doesn't converge to a useful generative model anymore as can be seen by a vanishingly small manifold entropy over all latent dimensions.
We can also observe that the manifold total correlation doesn't decrease meaningfully with an increasing $\beta$, as disentanglement is generally not regularized with the $\beta$-VAE loss (see \ref{sec:beta-VAE Additional results}).

\tocless\section{IMPACT AND LIMITATIONS}
\vspace{-10pt}

Normalizing flows can also be run in a bottleneck mode by setting some of the latent dimensions to zero \citep{brehmer2020flows}. Our method allows to sort latent dimensions according to their importance, so that the appropriate bottleneck size (i.e. the manifold dimension) can be determined after training, instead of being pre-determined as a hyperparameter.
This enables specifying the amount of retained information, measured in terms of nats, at inference time and offers precise knowledge and control over the amount of information loss via the bottleneck.

The biggest limitation of our framework and a general problem of DRL is the issue of identifiability. That is, given a generative model which was trained using a particular Disentanglement method, how do we know if the learned representations are in fact the true ones (i.e. the generative factors of the DGP) and not just some other “good” representation instead?
Guaranteeing the correspondence between true and learned representations is an important research topic \citep{hyvarinen2024identifiability, reizinger2024identifiable, wiedemer2023compositional}, but beyond the scope of this paper.

If the model is not properly aligned with the underlying DGP or fails to accurately learn the true distribution, our metrics may not reflect the true properties of the data and could yield spurious results. We demonstrated that training with an additional reconstruction loss can force the NF to concentrate entropy in an arbitrarily chosen (core) set of dimensions (see \cref{fig:H_rec}) and not necessarily revealing the underlying data manifold.

\tocless\section{CONCLUSION}
\vspace{-10pt}
We introduced a novel set of information-based metrics for the unsupervised evaluation of generative models.
These metrics have been practically enabled by recent progress in automatic differentiation.
Our metrics demonstrate on toy datasets and EMNIST that generative models will learn meaningful manifold alignment and latent disentanglement if network architectures and/or training procedures provide suitable inductive biases.

\section*{Acknowledgement}
This work is supported by Deutsche Forschungsgemeinschaft (DFG, German Research Foundation) under Germany’s Excellence Strategy EXC-2181/1 - 390900948 (the Heidelberg STRUCTURES Cluster of Excellence).
It is also supported by Informatics for Life funded by the Klaus Tschira Foundation and by the Carl-Zeiss-Stiftung (Projekt P2021-02-001 "Model-based AI").
DG acknowledges support by the German Federal Ministery of Education and Research (BMBF) (project EMUNE/031L0293A).
The authors acknowledge support by the state of Baden-Württemberg through bwHPC and the German Research Foundation (DFG) through grant INST 35/1597-1 FUGG.
UK thanks the Klaus Tschira Stiftung for their support via the SIMPLAIX project.

\bibliography{references}
\bibliographystyle{apalike}

\section*{Checklist}

\begin{enumerate}

\item For all models and algorithms presented, check if you include:
\begin{enumerate}
\item A clear description of the mathematical setting, assumptions, algorithm, and/or model. [Yes, see THEORY section in the Appendix \ref{sec:METHOD}]
\item An analysis of the properties and complexity (time, space, sample size) of any algorithm. [Yes, see NUMERICS section in the Appendix \ref{sec:NUMERICS}]
\item (Optional) Anonymized source code, with specification of all dependencies, including external libraries. [Yes, see \url{https://github.com/DanielGalperin/ManifoldEntropicMetrics}]
\end{enumerate}

\item For any theoretical claim, check if you include:
\begin{enumerate}
    \item Statements of the full set of assumptions of all theoretical results. [Not Applicable]
    \item Complete proofs of all theoretical results. [Not Applicable]
    \item Clear explanations of any assumptions. [Yes]     
\end{enumerate}

\item For all figures and tables that present empirical results, check if you include:
\begin{enumerate}
    \item The code, data, and instructions needed to reproduce the main experimental results (either in the supplemental material or as a URL). [Yes, see NUMERICS section in the Appendix \ref{sec:NUMERICS}]
    \item All the training details (e.g., data splits, hyperparameters, how they were chosen). [Yes, see in the Appendix \ref{sec:two moons training}, \ref{sec:emnist training} and \ref{sec:beta-VAE training}]
    \item A clear definition of the specific measure or statistics and error bars (e.g., with respect to the random seed after running experiments multiple times). [Not Applicable]
    \item A description of the computing infrastructure used. (e.g., type of GPUs, internal cluster, or cloud provider). [Yes, see NUMERICS section]
\end{enumerate}

\item If you are using existing assets (e.g., code, data, models) or curating/releasing new assets, check if you include:
\begin{enumerate}
    \item Citations of the creator If your work uses existing assets. [Yes]
    \item The license information of the assets, if applicable. [Not Applicable]
    \item New assets either in the supplemental material or as a URL, if applicable. [Not Applicable]
    \item Information about consent from data providers/curators. [Not Applicable]
    \item Discussion of sensible content if applicable, e.g., personally identifiable information or offensive content. [Not Applicable]
\end{enumerate}

\item If you used crowdsourcing or conducted research with human subjects, check if you include:
\begin{enumerate}
    \item The full text of instructions given to participants and screenshots. [Not Applicable]
    \item Descriptions of potential participant risks, with links to Institutional Review Board (IRB) approvals if applicable. [Not Applicable]
    \item The estimated hourly wage paid to participants and the total amount spent on participant compensation. [Not Applicable]
\end{enumerate}

\end{enumerate}

\newpage

\onecolumn

\appendix
\counterwithin{figure}{section}
\counterwithin{table}{section}
\counterwithin{equation}{section}
\counterwithin{lstlisting}{section}

\aistatstitle{Supplementary Materials}

\tableofcontents

\newpage

\section{METHOD} \label{sec:METHOD}

We outline how the manifold entropic metrics will be introduced through the following 5 definitions:
\begin{enumerate}
    \item We define a \textbf{latent manifold} $\Man_\set{S}$ as the set of all points generated by the decoder when only the latent variables with index set $\set{S}\subseteq \{1,...,D\}$ is varied, while the remaining ones are kept fixed. See \ref{sec:latent manifold}
    \item We define the \textbf{manifold random variable} $\X_\set{S}$ as the transformation of the latent random variable $\Z_\set{S}$ onto its latent manifold. See \ref{sec:manifold random variable}.
    \item We define the \textbf{manifold pdf} $q_\set{S}(\X_\set{S})$ as the pdf of the manifold random variable induced via the change-of-variables formula on a latent manifold. See \ref{sec:manifold pdf}.
    \item We define the \textbf{manifold entropy} $H(q_\set{S})$ as the differential entropy of a manifold random variable via its manifold pdf. See \ref{sec:manifold entropy}.
    \item We define the \textbf{manifold mutual information} $\mathcal{I}(q_\set{S}, q_\set{T})$ via the manifold entropy of disjoint index sets $\set{S}$ and $\set{T}$. See \ref{sec:manifold mutual information} and \ref{sec:manifold total correlation},\ref{sec:manifold cross-mutual information}.
\end{enumerate}

\subsection{Normalizing Flows}

A Normalizing Flow (NF) relates each data sample $\x \in \mathbb{R}^D$ and each latent vector/code $\z \in \mathbb{R}^D$ through a bijective mapping realized via the encoder $\z = \f(\x)$ and the decoder $\x = \g(\z)$.

For simplicity we assume that the data/latent space are unbounded and the encoder/decoder are defined everywhere.

The pdf of the latent space is assumed to be a standard normal distribution:
\begin{equation} \label{latent space full}
    p(\Z=\z) = \mathcal{N}(\z| 0, \I_D) = \frac{1}{\sqrt{(2\pi)^D}} \exp{\left(-\frac{1}{2} \left|\z\right|^2 \right)}
\end{equation}
where $\I_D \in \mathbb{R}^{D \times D}$ is the identity matrix and $|\z|$ denotes the L2-norm of $\z$.

\subsection{Latent space decomposition}
As we seek to analyze the effect that one single latent variable or a subset of latent variables can induce, we split the latent space into a partition of disjoint index sets $\mathcal{P}=\left\{ \set{S}_1, \dots, \set{S}_M \right\}$ such that every index in $\{1,...,D\}$ appears exactly once.
We can choose the partition arbitrary, though we assume that each set $\set{S} \in \mathcal{P}$ only contains ascending indices such that $|\mathbb{S}| = \max(\mathbb{S})-\min(\mathbb{S})$.
This is not a restriction, as the latent space can be renumbered i.e. re-sorted afterwards\obso{, which we often do in practice}.
This formulation allows us to decompose a latent code variable $\z$ into 
\begin{equation}
    \z = \begin{pmatrix} \z_{\set{S}_1} \\ \vdots \\ \z_{\set{S}_M} \end{pmatrix} \ \text{with } \z \in \mathbb{R}^D \text{ and } \z_{\set{S}_i} \in \mathbb{R}^{|\set{S}_i|}
\end{equation}

In practice we will refer to $\set{S}$ as the place-holder for an arbitrary index set of latent dimensions $\set{S}\subseteq \{1,...,D\}$.
For that we additionally define the complement $\nset{S} \coloneqq \{1,...,D\} \setminus \set{S}$, which contains all latent dimensions not contained in $\set{S}$.
This allows us to express a specific split of the latent vector, and as such the latent space, as:
\begin{equation}
     \z = [\z_{\set{S}}, \z_{\nset{S}}]
\end{equation}
If we refer to $\set{S}$ and $\set{T}$, we assume that they are disjoint index sets, i.e. $\set{S} \cap \set{T} = \varnothing$, if not noted otherwise.

Moreover we will often denote quantities which are defined over the entire latent space $\mathbb{R}^D$ without a subscript, e.g. $\J$, and those defined over a index set $\set{S}$ with the corresponding subscript, e.g. $\J_\set{S}$.

\subsection{Latent Manifold} \label{sec:latent manifold}

\begin{definition}%
    We define the latent manifold over $\mathbb{S}$ at a particular $\z_\nset{S}$ as
    \begin{equation}
        \Man_\set{S}(\z_\nset{S}) \coloneqq \left\{ \x = \g\left( \left[ \z_\set{S}, \z_\nset{S} \right] \right) : \ \z_\set{S} \in \mathbb{R}^{|\set{S}|} \right\}
    \end{equation}
    which is the set of all points in the latent subspace $\z_\set{S} \in \mathbb{R}^{|\set{S}|}$ transformed through the decoder $\g$ at a fixed $\z_\nset{S}$.
        For clarity we will often omit the dependence on $\z_\nset{S}$ and assume it implicitly such that
    \begin{equation}
        \Man_\set{S} = \Man_\set{S}(\z_\nset{S})
    \end{equation}
\end{definition}
The latent manifold over a single index, $\set{S}=i$, becomes simply a curve. Thus we can depict latent manifolds in two dimension as a \textit{contour grid} (see fig.1 in main paper) which is realized by transforming points sampled from a equidistant grid from latent to data space.

\subsection{Manifold random variable} \label{sec:manifold random variable}
\begin{definition}%
    We define the \textit{manifold random variable} over $\set{S}$ as
    \begin{equation} \label{def: manifold random variable}
    \X_\set{S} \coloneqq \g(\left[\Z_\set{S}, \z_\nset{S}\right]) 
    \end{equation}%
    which is the latent random variable $\Z_\set{S}$ transformed through the decoder $\g$ at a fixed $\z_\nset{S}$.
    It is the induced random variable from latent to data space and can only take values on the corresponding latent manifold $\Man_\set{S}(\z_\nset{S})$. \\
\end{definition}

Note that a single instance of the manifold random variable $\x_\set{S}=\X_\set{S}$ can take any value $\x \in \mathbb{R}^D$ if the latent code $\z_\nset{S}$ is set appropriately.
More formally: %
\begin{equation}
    \exists \z_\nset{S} \ \forall \x \ \text{s.t.} \ \x_\set{S} \equiv \x
\end{equation}
Equivalently the set of all latent manifolds over any $\set{S}$ spans the entire data space:
\begin{equation}
    \{\Man_\set{S}(\z_\nset{S}) \ \Big| \ \z_\nset{S} \in \mathbb{R}^{|\nset{S}|} \} \equiv \mathbb{R}^D
\end{equation}

Although we aim to formalize how to measure the effect of a subset of latent space variables $\z_{|\set{S}|}$, we still have to probe the entire latent space.%
This is because the decoder is a non-linear transformation and will not treat each latent variable independently.

Thus if we mention ''the latent manifold of $\set{S}$'', we usually don't refer to one particular manifold realization $\Man_\set{S}$ at a constant $\z_\nset{S}$ but rather the infinite collection of manifolds, generated by varying $\z_\set{S}$ at every $\z_\nset{S}$.

\subsection{Jacobian decomposition} \label{sec:jacobian decomposition}
In order to formulate a probability density over $\X_\set{S}$, we need to know how a latent manifold locally behaves with respect to $\Z_\set{S}$, i.e. we need to formalize ${\Delta \x_\set{S}}/{\Delta \z_\set{S}}$.
For this we note that the tangent space of a latent manifold $T_{\g(\z)}\Man_\set{S}$ is characterized by the matrix
\begin{equation}
     \frac{\partial \g(\z')}{\partial \z_\set{S}'} \Bigg|_{\z'=\z} \in \mathbb{R}^{D \times |\set{S}|}
\end{equation}
This can be checked as varying $\z_\set{S} \in \mathbb{R}^{|\set{S}|}$ will induce a change in $\g([\z_\set{S}, \z_\nset{S}]) \in \mathbb{R}^D$ along the latent manifold.

Remember that the Jacobian of the decoder at a point $\z$ is defined as $\J(\z) \coloneqq \frac{\partial \g(\z')}{\partial \z'}\big|_{\z'=\z}$.
We then define the Jacobian over $\set{S} = \left\{\set{S}_1, \dots \set{S}_{|\set{S}|} \right\}$ as the submatrix of $\J(\z)$ containing only the columns with indices in $\set{S}$ as
\begin{equation}
    \J_\set{S} \coloneq \frac{\partial \g(\z')}{\partial \z_\set{S}'} \Bigg|_{\z'=\z} =
    \begin{pmatrix}%
    \frac{\partial \g_1}{\partial \z_{\set{S}_1}} & \dots & \frac{\partial \g_1}{\partial \z_{\set{S}_{|\set{S}|}}} \\
    \vdots & \ddots & \vdots \\
    \frac{\partial \g_D}{\partial \z_{\set{S}_1}} & \dots & \frac{\partial \g_D}{\partial \z_{\set{S}_{|\set{S}|}}}
    \end{pmatrix} 
\end{equation}

The Jacobian over a joint index set $\set{ST} \coloneq \set{S} \cup \set{T}$ is denoted as the concatenation of the Jacobian submatrices over the individual index sets as
\begin{equation} \label{eq:Jacobian concatenation}
    \J_{\set{ST}} = [\J_\set{S}, \J_\set{T}]
\end{equation}

\subsection{Manifold pdf} \label{sec:manifold pdf}
Let us revisit the general cov-formula for a NF:
\begin{equation} \label{eq:general cov formula}
    q(\X = \g(\Z)) = p(\Z) \cdot \left| \J(\z) \right|^{-1}
\end{equation}
where $|.|$ denotes the volume of a squared or rectangular matrix $\boldsymbol{A} \in \mathbb{R}^{n \times m}$ with $n\ge m$ according to $|\boldsymbol{A}| \coloneqq \det(\boldsymbol{A}^T\boldsymbol{A})^\frac{1}{2}$.

Remember that the latent pdf factorizes:
\begin{equation} \label{eq:latent space factorization}
    p(\Z) = p_\set{S}(\Z_\set{S}) \cdot p_\nset{S}(\Z_\nset{S})
\end{equation}

We can now formulate the generative pdf of a manifold random variable defined via the cov-formula on a latent manifold.

\begin{definition}%
    We define the manifold pdf (mpdf) over $\set{S}$ as
    \begin{equation} \label{eq:manifold cov formula}
        q_\set{S}\left(\X_\set{S} = \g([\Z_\set{S}, \z_\nset{S}]) \right) = p_\set{S}(\Z_\set{S}) \cdot \left| \J_\set{S}(\z) \right|^{-1}
    \end{equation}
    in terms of the latent pdf $p_\set{S}(\Z_\set{S})$ and a volume-change term via the change-of-variables $\Z_\set{S} \rightarrow \X_\set{S}$.
    $q_\set{S}(\X_\set{S})$ is the (generative) pdf of the manifold random variable $\X_\set{S}$, which is defined on the latent manifold $\Man_\set{S}$, but can be evaluated everywhere $\forall \z$.
\end{definition}

The volume of $\J_\set{S}(\z)$ characterizes the volume-change between the latent code subspace of $\Z_\set{S}$ and the corresponding tangent space on the latent manifold $T_{\g(\z)}\Man_\set{S}$.
\paragraph{Note:}
It is easy to see that this formula reduces to the usual cov-formula for NFs by setting $\set{S}=\{1,...,D\}$ and a generalization of the cov-formula for injective flows (\cite{köthe2023review, kothari2021trumpets, caterini2021rectangular}).
Note however that for an injective flow, the dimensionality of the latent space is strictly smaller than that of the data space and thus the generative pdf is only defined for points lying on a single manifold while the mpdf can be evaluated everywhere.
In a similar vein, \cite{pmlr-v162-cunningham22a} define the logarithm of a mpdf $\log(q_\set{S})$ as the \textit{''log likelihood of a contour''}.

\paragraph{Motivation:}\mbox{}

Quantifying how much information a manifold random variable holds, will be achieved by defining the \textit{manifold entropy} in \ref{sec:manifold entropy}.

Let us investigate under which condition the full generative pdf $q(\X)$ factorizes into its manifold pdfs.
For this we write the full generative pdf in terms of the cov-formula and using \cref{eq:latent space factorization}:
\begin{equation}
    q(\X) = p(\Z) \cdot |\J(\z)|^{-1} = p_\set{S}(\Z_\set{S}) p_\nset{S}(\Z_\nset{S}) \cdot |\J_\set{S}(\z)|^{-1} |\J_\nset{S}(\z)|^{-1} \frac{|\J(\z)|^{-1}}{|\J_\set{S}(\z)|^{-1} |\J_\nset{S}(\z)|^{-1}}
\end{equation}
Using the definition of a mpdf \cref{eq:manifold cov formula} and rearranging the terms allows us to rewrite $q$ in terms of $q_\set{S}$ and $q_\nset{S}$:
\begin{equation}
    q(\X=[\X_\set{S}, \X_\nset{S}]) = q_\set{S}(\X_\set{S}) \cdot q_\nset{S}(\X_\nset{S}) \cdot \frac{|\J_\set{S}| |\J_\nset{S}|}{|\J|}
\end{equation}
We see that factorization is achieved for $|\J(\z)| \equiv |\J_\set{S}(\z)| \cdot |\J_\nset{S}(\z)| \ \forall \z$.
This holds true if all column vectors in $\J_\set{S}$ are orthogonal to that in $\J_\nset{S}$ everywhere, or in other words the latent manifolds intersect orthogonally everywhere $\Man_\set{S} \perp \Man_\nset{S}$.
Full proofs can be found in \cite{pmlr-v162-cunningham22a} Claims 6 and 8.
Thus the following equality holds
\begin{equation}
    \Man_\set{S} \perp \Man_\nset{S} \quad \Leftrightarrow \quad q(\X) = q_\set{S}(\X_\set{S}) \cdot q_\nset{S}(\X_\nset{S})
\end{equation}
Quantifying to which extent the full generative pdf factorizes into the manifold pdfs will be achieved by defining the \textit{manifold mutual information} in \ref{sec:manifold mutual information}.

\subsection{Total Entropy}
It is straight forward to define the differential entropy of the random variable $\X$ induced by the full generative pdf $q(\X)$ as:
\begin{equation}
    H(q) = -\int_{\mathbb{R}^{D}}{q(\X=\x) \log(q(\X = \x))} d\x
\end{equation}
Using the cov-formula \cref{eq:general cov formula} and noting that the latent space is normal \cref{latent space full} it can be reformulated in the latent space as
\begin{equation} \label{eq:total entropy latent}
\begin{split}
    \boxed{ H(q) } &= -\int_{\mathbb{R}^{D}}{p(\Z=\z) \log(p(\Z = \z) \left| \J(\z) \right|^{-1} )} d\z \\
    &= \int_{\mathbb{R}^{D}} p(\z) \left[ - \log\left( p(\z)\right) + \log \left| \J(\z) \right| \right] d\z \\
    &= \int_{\mathbb{R}^{D}} p(\z) \left[\frac{D}{2}\log(2\pi) + \frac{1}{2}\left|\z\right|^2 + \log \left| \J (\z) \right| \right] d\z \\
    &= \boxed{ \frac{D}{2} \left(\log(2\pi) + 1\right) + \Expt{\z}{\log \left| \J(\z) \right|} }
\end{split}
\end{equation}
\obso{where the constants denote the differential entropy of the latent prior $p(\Z)$.}

This quantity denotes how much information is induced in the data space by the entire latent space.
By itself is not a useful metric as it doesn't tell us anything about the influence of single latent dimensions.

\paragraph{Note:}
If we would sample from the true data pdf, with $\x^* \sim p^*(\x^*)$, transform the samples to the latent space by the encoder $\f$, with $\z^* = \f(\x^*)$, and finally estimate the differential entropy via $\Expt{\z*}{\cdot}$ instead of $\Expt{\z}{\cdot}$, this quantity would become the cross-entropy between the true pdf and the generative pdf and is widely known as the negative-log-likelihood (NLL), used to evaluate and compare the performance of generative models on a dataset.
\obso{
\paragraph{Reminder:}
The differential entropy can be negative, in contrast to the ''normal'' entropy.
}
\subsection{Manifold Entropy} \label{sec:manifold entropy}
Here we will introduce the manifold entropy, which is the central manifold entropic metric and a building block for complementary metrics.

Naively we can first define the differential entropy of a manifold random variable $\X_\set{S}$, with its corresponding mpdf $q_\set{S}$, at a fixed $\z_\nset{S}$, by application of the general definition:
\begin{equation} \label{eq:naive entropy of q_S}
    H_\set{S}(q_\set{S}| \z_\nset{S}) = -\int_{\Man_\set{S}(\z_\nset{S})}{q_\set{S}(\X_\set{S}=\x_\set{S}) \log(q_\set{S}(\X_\set{S} = \x_\set{S}))} d\x_\set{S}
\end{equation}
We denote the entropy in terms of the mpdf $q_\set{S}$ and not $\X_\set{S}$ to avoid confusion regarding the functional dependency to a specific $\z_\nset{S}$.
We denote the entropy with the subscript ${\cdot}_\set{S}$ as it is a measure of the subspace $\Man_\set{S}$.%

Inserting the definition of the mpdf eq.(\ref{eq:manifold cov formula}), of the latent pdf eq.(\ref{eq:latent space factorization}, \ref{latent space full}) and changing the integration variables $\X_\set{S} \rightarrow \Z_\set{S}$ allows us to evaluate it:
\begin{equation} \label{eq:naive entropy of q_S latent}
\begin{split}
    H_{\set{S}}(q_{\set{S}}| \z_\nset{S}) &= -\int_{\mathbb{R}^{|\set{S}|}} p_{\set{S}}(\z_\set{S}) \log\left( p_{\set{S}}(\z_{\set{S}}) \left| \J_{\set{S}} (\z=[\z_\set{S}, \z_\nset{S}]) \right|^{-1} \right) d\z_\set{S} \\
    &= \int_{\mathbb{R}^{|\set{S}|}} p_{\set{S}}(\z_\set{S}) \left[ - \log\left( p_{\set{S}}(\z_{\set{S}})\right) + \log \left| \J_{\set{S}}(\z=[\z_\set{S}, \z_\nset{S}]) \right| \right] d\z_\set{S} \\
    &= \int_{\mathbb{R}^{|\set{S}|}} p_{\set{S}}(\z_\set{S}) \left[\frac{|\set{S}|}{2}\log(2\pi) + \frac{1}{2}\left|\z_\set{S}\right|^2 + \log \left| \J_{\set{S}}(\z=[\z_\set{S}, \z_\nset{S}]) \right| \right] d\z_\set{S} \\
    &= \frac{|\set{S}|}{2} \left(\log(2\pi) + 1\right) + \Expt{\z_\set{S}}{\log \left| \J_{\set{S}}(\z=[\z_\set{S}, \z_\nset{S}]) \right|}
\end{split}
\end{equation}

In order to obtain a global metric, similar to the total entropy from above, we average $H_{\set{S}}(q_{\set{S}}| \z_\nset{S})$ over the subspace of $\Z_\nset{S}$ as $\Expt{\z_\nset{S}}{ H_{\set{S}}(q_{\set{S}}| \z_\nset{S}) }$ which gets rid of the dependency on $\z_\nset{S}$ and leads to the definition:
\begin{definition}%
    We define the manifold entropy over $\set{S}$ as
    \begin{equation}
        H(q_\set{S}) \coloneq -\int_{\mathbb{R}^D}{q(\X=\x) \log(q_\set{S}(\X_\set{S} = \x))} d\x
    \end{equation}
    which is the differential entropy of a manifold random variable $\X_\set{S}$, with its corresponding manifold pdf $q_{\set{S}}$, marginalized over the complement space $\Man_\nset{S}$.
\end{definition}

We drop the subscript on $H$ as this quantity is a measure evaluated in the entire space.

Writing out the manifold entropy, analogously to \cref{eq:naive entropy of q_S latent}, leads to:
\begin{equation} \label{eq:manifold entropy latent}
\begin{split}
    \boxed{ H(q_{\set{S}}) } &= -\int_{\mathbb{R}^{D}} p(\z) \log\left( p_{\set{S}}(\z_{\set{S}}) \left| \J_{\set{S}} (\z) \right|^{-1} \right) d\z \\
    &= \int_{\mathbb{R}^{D}} p(\z) \left[ - \log\left( p_{\set{S}}(\z_{\set{S}})\right) + \log \left| \J_{\set{S}}(\z) \right| \right] d\z_\set{S} \\
    &= \int_{\mathbb{R}^{D}} p(\z) \left[\frac{|\set{S}|}{2}\log(2\pi) + \frac{1}{2}\left|\z_\set{S}\right|^2 + \log \left| \J_{\set{S}} (\z) \right| \right] d\z \\
    &=\boxed{ \frac{|\set{S}|}{2} \left(\log(2\pi) + 1\right) + \Expt{\z}{\log \left| \J_{\set{S}}(\z) \right|} }
\end{split}
\end{equation}

Finally, this metric can be readily evaluated by sampling from the prior and computing the Jacobian submatrix over $\set{S}$ at each point (see \ref{sec:jacobian decomposition}).

\obso{Finally, we can interpret the final form of the total entropy \cref{eq:total entropy latent} or the manifold entropy \cref{eq:manifold entropy latent} as a analogous change-of-variables formula for the differential entropy. More precisely:
\begin{equation}
\begin{split}
    H(q) &= H(p) + \Expt{\z}{\log \left|\J(\z)\right|} \\
    H(q_\set{S}) &= H(p_\set{S}) + \Expt{\z}{\log \left|\J_\set{S}(\z)\right|}
\end{split}
\end{equation}
and more importantly it factorizes similarly to the mpdfs, which we will see in ???.
As a final sanity check, we can infer easily that the manifold entropy over $\set{S}=\{1,...,D\}$ is exactly the total entropy.
}

\subsection{Manifold Mutual Information} \label{sec:manifold mutual information}
In analogy to the manifold entropy, we will first define the mutual information between two manifold random variables $\X_\set{S}$ and $\X_\set{T}$, with their corresponding mpdfs $q_\set{S}$ and $q_\nset{S}$.
Additionally we denote the joint manifold random variable $\X_\set{ST}$ with its mpdf $q_\set{ST}$.
The mutual information is then defined for a fixed $\z_\nset{ST}$ as
\begin{equation} \label{eq:naive manifold mutual information of q_S q_T}
    \mathcal{I}_\set{S,T}(q_\set{S}, q_\set{T}| \z_\nset{ST}) =\int_{\Man_\set{ST}(\z_\nset{ST})}
    q_\set{ST}(\X_\set{ST}=\x_\set{ST}) \log\left(\frac{q_{\set{ST}}(\X_{\set{ST}}=\x_\set{ST})}{q_{\set{S}}(\X_{\set{S}}=\x_\set{S}) \cdot q_{\set{T}}(\X_{\set{T}}=\x_\set{T})}\right) d\x_\set{ST}
\end{equation}

Inserting the definition of the mpdf eq.(\ref{eq:manifold cov formula}), of the latent pdf eq.(\ref{eq:latent space factorization}) and changing the integration variables $\X_\set{ST} \rightarrow \Z_\set{ST}$ allows us to evaluate it:
\begin{equation} \label{eq:naive manifold mutual information of q_S q_T latent}
\begin{split}
    &\mathcal{I}_\set{S,T}(q_\set{S}, q_\set{T}| \z_\nset{ST}) =\int_{\mathbb{R}^{|\set{S}|\times |\set{T}|}} p_{\set{ST}}(\z_\set{ST}) \log\left( \frac{p_{\set{ST}}(\z_\set{ST}) \left| \J_{\set{ST}} (\z=[\z_\set{ST}, \z_\nset{ST}]) \right|^{-1}}{p_{\set{S}}(\z_\set{S}) \left| \J_{\set{S}} (\z=[\z_\set{ST}, \z_\nset{ST}]) \right|^{-1} \cdot p_{\set{T}}(\z_\set{T}) \left| \J_{\set{T}} (\z=[\z_\set{ST}, \z_\nset{ST}]) \right|^{-1}} \right) d\z_\set{ST} \\
    &= \int_{\mathbb{R}^{|\set{S}|\times |\set{T}|}} p_{\set{ST}}(\z_\set{ST}) \left[ \log \left| \J_{\set{S}}(\z=[\z_\set{ST}, \z_\nset{ST}]) \right| + \log \left| \J_{\set{T}}(\z=[\z_\set{ST}, \z_\nset{ST}]) \right| - \log \left| \J_{\set{ST}}(\z=[\z_\set{ST}, \z_\nset{ST}]) \right| \right] d\z_\set{S} \\
    &= \Expt{\z_\set{ST}}{\log \left| \J_{\set{S}}(\z=[\z_\set{ST}, \z_\nset{ST}]) \right| + \log \left| \J_{\set{T}}(\z=[\z_\set{ST}, \z_\nset{ST}]) \right| - \log \left| \J_{\set{ST}}(\z=[\z_\set{ST}, \z_\nset{ST}]) \right|}
\end{split}
\end{equation}

Again, in order to obtain a global metric, we can simply average $\mathcal{I}_\set{S,T}(q_\set{S}, q_\set{T}| \z_\nset{ST})$ over the subspace of $\Z_\nset{ST}$ as $\Expt{\z_\nset{ST}}{ \mathcal{I}_\set{S,T}(q_\set{S}, q_\set{T}| \z_\nset{ST}) }$ which gets rid of the dependency on $\z_\nset{ST}$ and leads to the definition:
\begin{definition}%
    We define the manifold mutual information between $\set{S}$ and $\set{T}$ as
    \begin{equation}
        \mathcal{I}(q_\set{S}, q_\set{T}) \coloneq \int_{\mathbb{R}^D}{q(\X=\x) \log\left(\frac{q_{\set{S} \set{T}}(\X_{\set{S} \set{T}}=\x)}{q_{\set{S}}(\X_{\set{S}}=\x) \cdot q_{\set{T}}(\X_{\set{T}}=\x)}\right) d\x}
    \end{equation}
    which is the differential mutual information between two manifold random variables $\X_\set{S}$ and $\X_\set{T}$, with their corresponding manifold pdfs $q_{\set{S}}$, $q_{\set{T}}$, marginalized over the complement $\X_\set{ST}$.
\end{definition}

Writing out the manifold mutual information, analogously to \cref{eq:naive manifold mutual information of q_S q_T latent}, leads to
\begin{equation} \label{eq:manifold mutual information latent}
\begin{split}
    \boxed{ \mathcal{I}(q_\set{S}, q_\set{T}) } &=\int_{\mathbb{R}^{D}} p(\z) \log\left( \frac{p_{\set{ST}}(\z_\set{ST}) \left| \J_{\set{ST}} (\z) \right|^{-1}}{p_{\set{S}}(\z_\set{S}) \left| \J_{\set{S}} (\z) \right|^{-1} \cdot p_{\set{T}}(\z_\set{T}) \left| \J_{\set{T}} (\z) \right|^{-1}} \right) d\z \\
    &= \int_{\mathbb{R}^{D}} p(\z) \left[ \log \left| \J_{\set{S}}(\z) \right| + \log \left| \J_{\set{T}}(\z) \right| - \log \left| \J_{\set{ST}}(\z) \right| \right] d\z \\
    &= \boxed{ \Expt{\z}{\log \left| \J_{\set{S}}(\z) \right| + \log \left| \J_{\set{T}}(\z) \right| - \log \left| \J_{\set{ST}}(\z) \right|} }
\end{split}
\end{equation}

Finally, this metric can be readily evaluated by sampling from the prior and computing the Jacobian submatrices over $\set{ST}$, $\set{S}$ and $\set{T}$ at each point (see \ref{sec:jacobian decomposition}).

\paragraph{Note:}
This metric is strictly non-negative, see Claim 4 \citep{pmlr-v162-cunningham22a}, and zero only if the mpdf $q_\set{ST}$ factorizes into $q_\set{S}$ and $q_\set{T}$, which can be directly inferred from the definition.

\subsection{Entropic Decomposition}
Comparing the manifold entropy and manifold mutual information, we see that the manifold entropy of the joint index set $\set{S} \set{T} \coloneq \set{S} \cup \set{T}$ decomposes into the respective manifold entropic metrics as
\begin{equation} \label{manifold cross-entropy deconstruction}
    H(q_{\set{S} \set{T}}) = H(q_{\set{S}}) + H(q_{\set{T}}) - \mathcal{I}(q_{\set{S}}, q_{\set{T}})
\end{equation}
\obso{See e.g. fig.1 in the main paper.}
This means that the manifold mutual information can be directly obtained by computing the respective manifold entropies. In fact, all complementary metrics can be obtained similarly, which makes their definition easier.

\subsection{Manifold Total Correlation} \label{sec:manifold total correlation}
It is possible to decompose the total entropy into a sum of manifold entropies over $\set{S} \in \mathcal{P}$ minus a term which we call the \textit{manifold total correlation} $\mathcal{I}_{\mathcal{P}}$.
The decomposition is:
\begin{equation} \label{eq:total entropy decomposition}
    H(q) = \sum_{\set{S} \in \mathcal{P}} H(q_{\set{S}}) - \mathcal{I}_{\mathcal{P}}
\end{equation}
More formally we define:
\begin{definition}%
    We define the manifold total correlation over $\set{S} \in \mathcal{P}$ as
    \begin{equation}
        \mathcal{I}_{\mathcal{P}} \coloneq \int_{\mathbb{R}^D}{q(\X=\x) \log\left(\frac{q(\X=\x)}{\prod_{\set{S} \in \mathcal{P}} q_{\set{S}}(\X_{\set{S}}=\x)}\right) d\x}
    \end{equation}
    which is the total correlation between all manifold random variables $\X_\set{S}$, with their corresponding manifold pdfs $q_{\set{S}}$.
\end{definition}

Note that the total correlation is not simply a sum over manifold mutual information terms between all $\set{S}, \set{T} \in \mathcal{P} \text{ with } \set{S} \neq \set{T}$, but between all possible combinations of disjoint index sets in $\mathcal{P}$.
Therefore it quantifies the redundancy or dependency among all manifold random variables simultaneously.

Reformulating \cref{eq:total entropy decomposition} we can rewrite
\begin{equation} \label{eq:manifold total correlation decomposition}
    \boxed{ \mathcal{I}_{\mathcal{P}} = \sum_{\set{S} \in \mathcal{P}} H(q_{\set{S}}) - H(q) = \Expt{\z}{ \sum_{\set{S} \in \mathcal{P}} \log \left| \J_{\set{S}}(\z) \right| - \log \left| \J(\z) \right|} }
\end{equation}
which is similar to the ''pointwise mutual information of a partition'' in \cite{pmlr-v162-cunningham22a}.

\paragraph{Note:}
This metric is strictly non-negative, see Claim 6 \citep{pmlr-v162-cunningham22a}, and zero only if the full generative pdf factorizes as
\begin{equation}
    q(\X) = \prod_{\set{S} \in \mathcal{P}} q_{\set{S}}(\X_\set{S})
\end{equation}
which can be directly inferred from its definition.

\subsection{Manifold Cross-Mutual Information} \label{sec:manifold cross-mutual information}
Similarly to the manifold mutual information, it might be useful to measure a mutual information between manifold random variables a{\em cross} models.
For that let us assume that two models $a$ and $b$ were trained on the same dataset (and have the same latent prior!).
Convergence of the training implies that the generative pdf of each model approximates the true pdf sufficiently well and thus they are similar to each other as well: $q^a(\X) \approx q^b(\X)$.
However this does not imply that both models are identical, which is connected to the identifiability-problem of non-linear ICA.
As the decoders $\g^a$ and $\g^b$ are not fully determined by the data in the non-linear case, the latent manifolds of each model can be drastically different (see e.g. fig.1 in main paper).

We denote a manifold random variable of each model as $\X_\set{S}^a$ and $\X_\set{T}^b$, where $\set{S}$ and $\set{T}$ are arbitrary index sets of $\{1,...,D\}$ but not strictly disjoint.
With this we define the \textbf{manifold cross-mutual information} as the manifold mutual information between $\X_\set{S}^a$ and $\X_\set{T}^b$, with the respective mpdfs $q_\set{S}^a$ and $q_\set{T}^b$, as:
\begin{equation}
    \boxed{ \mathcal{I}(q^a_\set{S}, q^b_\set{T}) = \Expt{\z}{\log\left|\J_\set{S}^a(\z)\right| + \log\left|\J_\set{T}^b(\z)\right| - \log\left|\J_\set{ST}^{ab}(\z)\right|} }
\end{equation}
where $\J_\set{S}^a$ and $\J_\set{T}^b$ are the submatrices of the Jacobians evaluated for each model $a$ and $b$ separately and $\J_\set{ST}^{ab}$ is the concatenation of both submatrices (see \cref{eq:Jacobian concatenation}).

\subsection{Relation to Metrics in Main Paper}
In the main paper we introduced the manifold entropic metrics in a simplified way over index sets of single latent dimensions such that $\mathcal{P}=\{1,...,D\}$.
To point out differences in the notation we mention each:
\begin{enumerate}[wide] %
    \item ''Total Entropy'' $H(q) \in \mathbb{R}$.
    \item ''Manifold Entropy'' (ME) $H(q_i) \in \mathbb{R}^D$.
    \item ''Total Manifold Correlation'' (TMC) $\mathcal{I} \in \mathbb{R}$, where we drop the subscript ${\cdot}_{\mathcal{P}}$.
    \item ''Manifold Pairwise Mutual Information''-matrix (MPMI) $\mathcal{I}_{ij}(q_i, q_j) \in \mathbb{R}^{D \times D}$ in analogy to the manifold mutual information.
    With the additional ''pairwise'' we wish to emphasize that a pair of latent dimensions $i$ and $j$ are compared at a time, and the additional subscript ${\cdot}_{ij}$ emphasizes that this pair of dimensions appears as the $ij$-entry in the matrix.
    \item ''Manifold Cross-Pairwise Mutual Information''-matrix (MCPMI) $\mathcal{I}^{ab}_{ij}(q^a_i, q^b_j) \in \mathbb{R}^{D \times D}$ in analogy to the cross-manifold mutual information.
    With the additional ''cross-pairwise'' we wish to emphasize that a pair of latent dimensions $i$ and $j$ across two models $a$ and $b$ are compared at a time, and the additional subscript ${\cdot}_{ij}$ emphasizes that this pair of dimensions appears as the $ij$-entry in the matrix.
\end{enumerate}

\subsection{Additional DRL conditions}
In addition to Alignment and Disentanglement we can deduce two additional conditions if the DGP respects the manifold hypothesis:
\begin{enumerate}
    \item[c)] There is an intrinsic dimensionality $d$ of the data $\Leftrightarrow$ The manifold entropy is non-vanishing for exactly $d$ dimensions
    \item[d)] The contribution of noise is small and independent of the data $\Leftrightarrow$ The manifold entropy of $D-d$ dimensions is small and the manifold mutual information between the respective index sets vanishes
\end{enumerate}

\newpage

\section{INTERPRETATION VIA PCA} \label{sec:INTERPRETATION VIA PCA}

\subsection{PCA on Normalizing Flows}

Let us investigate how PCA is motivated and how it can be implemented using normalizing flows.
In the most restrictive case, we assume that data samples stem from a multivariate normal distribution
\begin{equation} \label{eq:PCA data normal dist}
    p^*(\X=\x) = \mathcal{N}(\x| \mm, \S)
\end{equation}
where $\mm \in \mathbb{R}^D$ is the mean of the data and $\S \in \mathbb{R}^{D \times D}$ its covariance matrix. \\
As the latent space of a normalizing flow is a standard normal $p(\Z=\z) = \mathcal{N}(\z| 0, \I_D)$ is is sufficient to linearly transform samples between data and latent space through an affine transformation.
Thus we define the decoder as:
\begin{equation} \label{eq:PCA general decoder}
    \g(\z) = \A \z + \b
\end{equation}
and the encoder conversely as
\begin{equation}
    \f(\x) = \A^{-1} \left(\x - \b\right)
\end{equation}
with $\A \in \mathbb{R}^{D \times D}$ and $\b \in \mathbb{R}^D$ being the learnable parameters of the flow.
\\ %
The generative pdf of the flow can be readily obtained as:
\begin{equation}
    q(\X=\x) = \mathcal{N}(\x | \b, \A \A^T)
\end{equation}
By comparing this with \cref{eq:PCA data normal dist} we can formalize the conditions of achieving Density Estimation as
\begin{equation} \label{eq:PCA DL b}
    \b \equiv \mm \coloneq \Expt{\x}{\x}
\end{equation}
\begin{equation} \label{eq:PCA DL A}
    \A\A^T \equiv \S \coloneq \Expt{\x}{\left(\x - \mm\right) \left(\x - \mm\right)^T}
\end{equation}

Thus, learning the data distribution is equivalent with estimating the mean and covariance matrix of the data.

However this solution is not unique as we can right-multiply $\A$ with any orthogonal matrix $\V$ such that $\A \coloneq \boldsymbol{\tilde{A}} \V$ and \cref{eq:PCA DL A} will still hold:
\begin{equation}
    \A \A^T = (\boldsymbol{\tilde{A}} \V) (\boldsymbol{\tilde{A}} \V)^T = \boldsymbol{\tilde{A}} \V \V^T \boldsymbol{\tilde{A}}^T = \boldsymbol{\tilde{A}} \boldsymbol{\tilde{A}}^T
\end{equation}
This redundancy is eliminated in PCA by performing a spectral decomposition on $\S$:
\begin{equation} \label{eq:PCA spectral decomposition}
    \S = \U \L \U^T
\end{equation}
where $\U$ is an orthogonal matrix containing the eigenvectors and $\L$ a diagonal matrix containing the eigenvalues. 

Comparing \cref{eq:PCA DL A} and \cref{eq:PCA spectral decomposition} the unique PCA-solution of $\A$ can be identified to
\begin{equation} \label{eq:PCA solution A}
    \A = \U\L^{\frac{1}{2}}
\end{equation}

\subsection{Alignment and Disentanglement of PCA}
Here we will investigate which desired properties the PCA-solution has.

The final decoder of the PCA-solution (using \cref{eq:PCA general decoder,eq:PCA solution A}) reads
\begin{equation} \label{eq:PCA final decoder}
    \g(\z) = \U\L^{\frac{1}{2}} \z + \b
\end{equation}
where the Jacobian can be easily evaluated as
\begin{equation} \label{eq:PCA final Jacobian}
    \J(\z) \coloneq \frac{\partial \g(\z)}{\partial \z} = \U\L^{\frac{1}{2}}
\end{equation}
As $\J$ is independent of $\z$, it describes the transformation both $\textbf{locally}$ and \textbf{globally}.
This will allow us to formulate conditions on DRL, in terms of the PCA solution, which are at first locally defined but will be generalized in a global manner using our manifold entropic metrics.

Remember that the change a single latent dimensions $i$ induces in the data is denoted by the $i$-th column vector of the Jacobian
\begin{equation} \label{eq:PCA jacobian i}
    \J_i = \U_i \Lambda_{ii}^{\frac{1}{2}}
\end{equation}
where $\U_i$, the $i$-th column vector of $\U$, is the $i$-th eigenvector and $\Lambda_{ii}^{\frac{1}{2}}$, the $i$-th diagonal entry of $\L^{\frac{1}{2}}$, is the $i$-th eigenvalue of the PCA solution.
In PCA we usually sort the latent dimensions $i$ in a descending order such that:
\begin{equation}
    \Lambda_{ii} \geq \Lambda_{jj} \ \forall i < j
\end{equation}

\begin{figure}[!htbp]
    \centering
    \begin{subfigure}[t]{.4\linewidth}
        \centering
        \includegraphics[width=\textwidth]{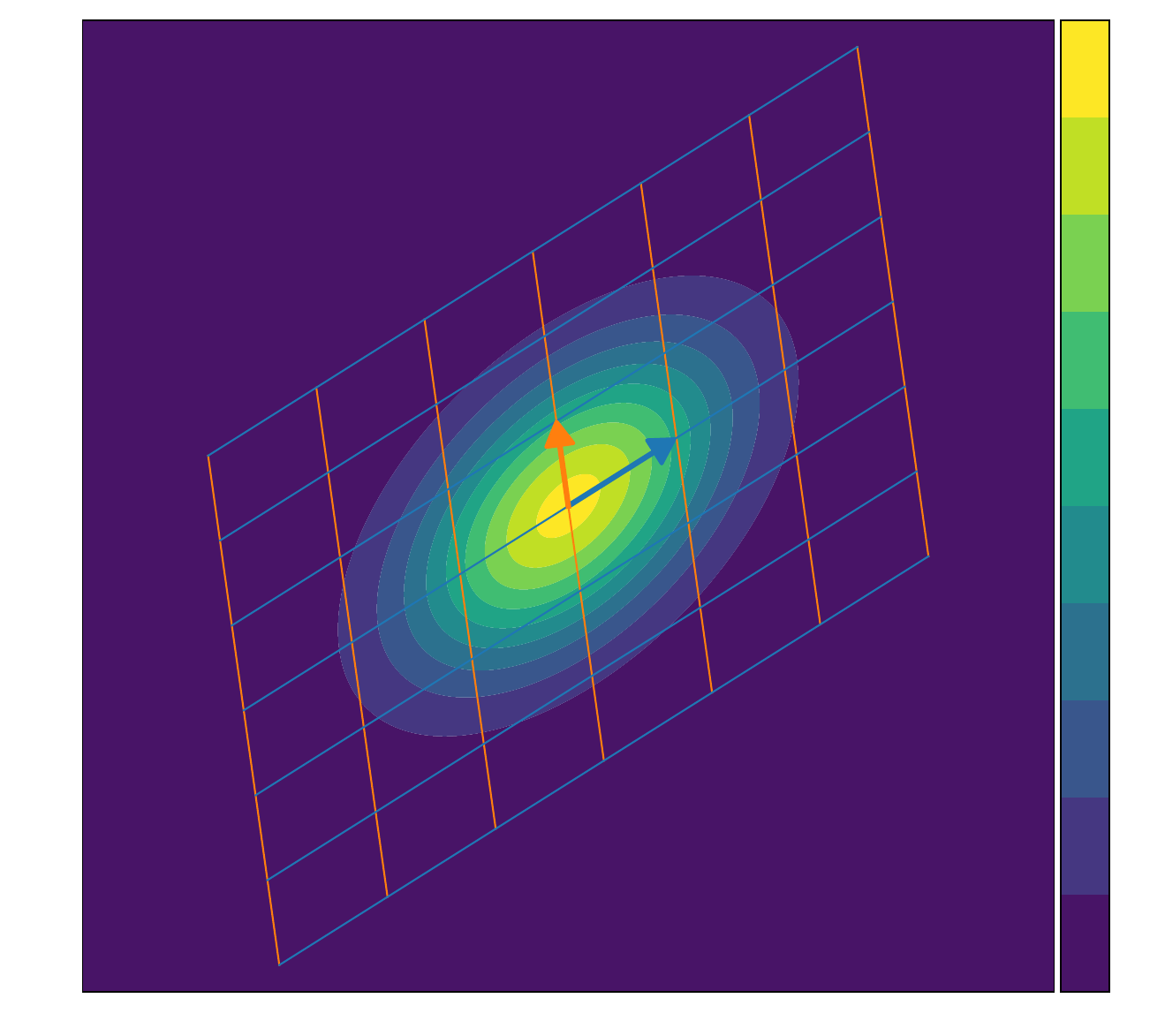}%
        \caption{A general NF solution. The Gaussian distribution is learned perfectly.}
        \label{fig:PCA solution 1}
    \end{subfigure}
    \hspace{10pt}
    \begin{subfigure}[t]{.4\linewidth}
        \centering
        \includegraphics[width=\textwidth]{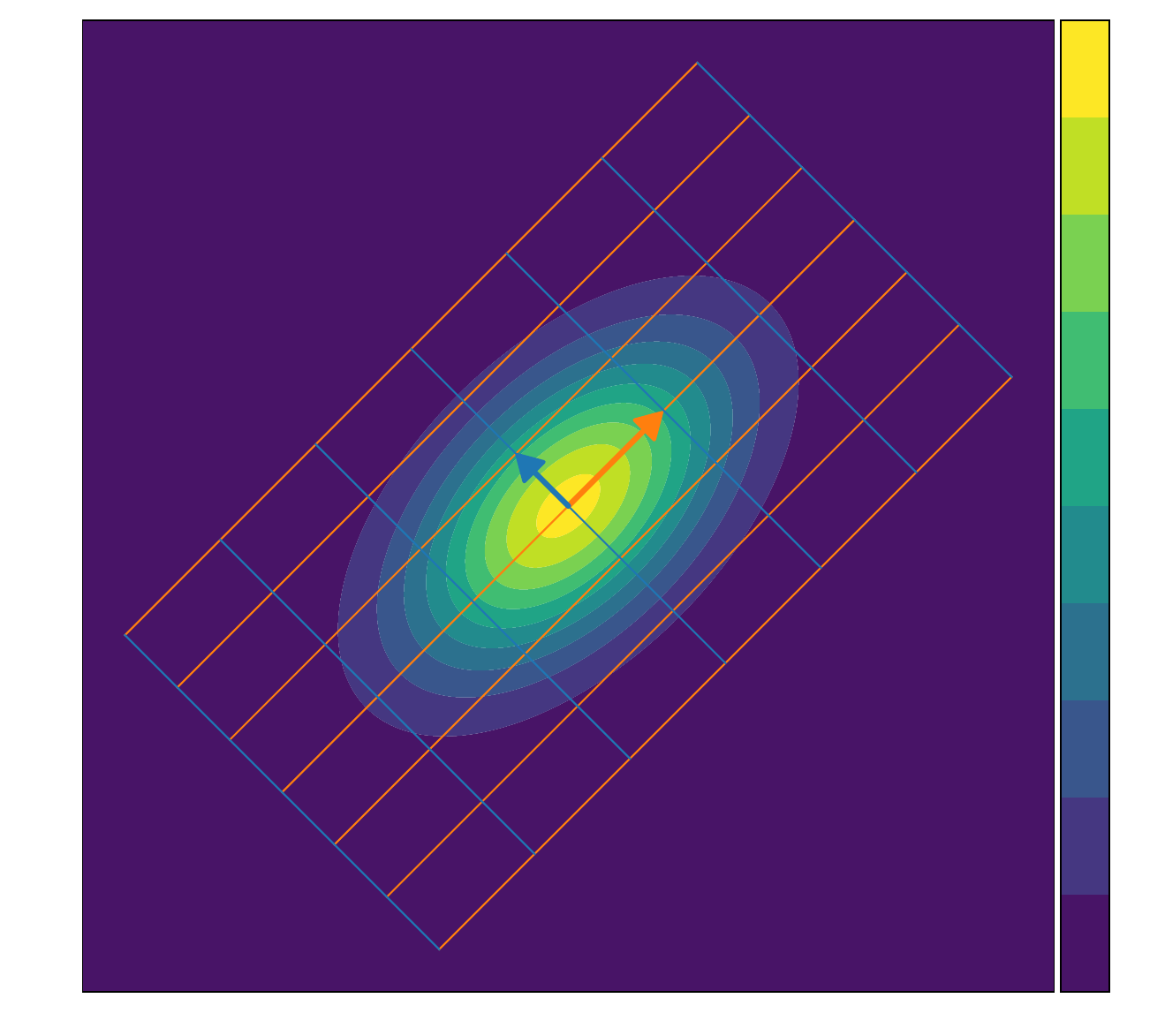}%
        \caption{NF solution of PCA additionally achieves \textit{Alignment} and \textit{Disentanglement} by orienting $\J_i$ (orange and blue arrows) as directions of \textit{aligned} and \textit{disentangled} variations in the data.}
        \label{fig:PCA solution 2}
    \end{subfigure}
    \caption{Depiction of a 2D affine NF trained on a Gaussian distribution.
    The generative pdf and the contour grid (orange/blue lines) are depicted.
    The Jacobian column vectors $\J_i$ (orange/blue arrow) point along the latent manifolds $\Man_i$, which are simply linear subspaces of $\mathbb{R}^2$.}
    \label{fig:PCA solution}
\end{figure}

With this we finally obtain the interpretation of $\{\J_i\}_{i=1:D}$ as directions of \textit{aligned} and \textit{disentangled} variations in the data:
\begin{enumerate}
    \item All $\J_i$ are sorted from largest to smallest magnitude as the eigenvectors $\U_i$ have unit length and the eigenvalues $\Lambda_{ii}$ are sorted. More formally:
    \begin{equation}
        |\J_i|^2 = \J_i^T \J_i = \Lambda_{ii}^{\frac{1}{2}} \U_i^T \U_i \Lambda_{ii}^{\frac{1}{2}} = \Lambda_{ii}
    \end{equation}
    \begin{equation}
        \Rightarrow \left|\J_i\right| \geq \left|\J_j\right| \ \forall i < j
    \end{equation}
    We refer to this property as \textbf{local alignment}.
    \item All $\J_i$ will be orthogonal to each other as the scalar product between any two eigenvectors $\U_i$ is always zero:
    \begin{equation}
        \J_i^T \J_j = \Lambda_{ii}^{\frac{1}{2}} \U_i^T \U_j \Lambda_{jj}^{\frac{1}{2}} = \boldsymbol{0}
    \end{equation}
    \begin{equation}
        \Rightarrow \J_i \perp \J_j \ \forall i \neq j
    \end{equation}
    We refer to this property as \textbf{local disentanglement}.
\end{enumerate}
See \cref{fig:PCA solution} for an illustration.
Thus PCA will find a unique solution\footnote{Except if some eigenvalues are identical} where both local alignment and disentanglement are fulfilled.

For a non-linear NF, these \textbf{local} properties cannot be naively generalized to \textbf{global} ones, as the Jacobian is not constant wrt $\z$.

\subsection{Manifold Entropic Metrics on PCA}

To represent a latent manifold of the PCA-solution we can decompose the decoder output from \cref{eq:PCA final decoder} into $\set{S}$ and $\nset{S}$ via $\z=[\z_\set{S}, \z_\nset{S}]$ as:
\begin{equation}
    \g([\z_\set{S}, \z_\nset{S}]) = \left[ \U_\set{S} \L_\set{SS}^{\frac{1}{2}} \z_\set{S} + \b_\set{S}, \U_\nset{S} \L_{\nset{S} \hspace{0.5pt} \nset{S}}^{\frac{1}{2}} \z_\nset{S}  + \b_\nset{S} \right]
\end{equation}
where $\U_\set{S}$ is a semi-orthogonal matrix denoting the submatrix containing only the columns of $\U$ with indices in $\set{S}$, $\L_\set{SS}$ denotes a diagonal matrix containing only the diagonal elements of $\L$ with indices in $\set{S}$ and $\b_\set{S}$ denotes the subvector containing only the rows of $\b$ with indices in $\set{S}$. Similarly with $\nset{S}$.

One can easily see that the latent manifold
$\Man_\set{S}(\z_\nset{S}) = \left\{ \g([\z_\set{S}, \z_\nset{S}]) : \z_\set{S} \in \mathbb{R}^{|\set{S}|} \right\}$ becomes a linear subspace of $\mathbb{R}^D$ for a fixed $\z_\nset{S}$.

We can evaluate the manifold pdf by formulating the Jacobian submatrix \cref{eq:PCA final Jacobian}:
\begin{equation}
    \J_\set{S}(\z) = \U_\set{S} \L_\set{SS}^{\frac{1}{2}}
\end{equation}
and its volume
\begin{equation}
    \left| \J_\set{S}(\z) \right| \coloneq \det\left( \J_\set{S}(\z)^T \J_\set{S}(\z) \right)^{\frac{1}{2}} = \det\left( (\U_\set{S} \L_\set{SS}^{\frac{1}{2}})^T \U_\set{S} \L_\set{SS}^{\frac{1}{2}} \right)^{\frac{1}{2}} = \prod_{i \in \set{S}} \Lambda_{ii}^{\frac{1}{2}}
\end{equation}
where we used that for a semi-orthogonal matrix it holds $\U_\set{S}^T \U_\set{S} = \mathbf{I}_{|\set{S}|}$ and the determinant of the diagonal matrix $\L_\set{SS} \in \mathbb{R}^{|\set{S}| \times |\set{S}|}$ is the product of its diagonal elements.

With this, the manifold pdf over $\set{S}$ becomes:
\begin{equation}
    q_\set{S}(\X_\set{S} = \g([\Z_\set{S}, \z_\nset{S}])) = p_\set{S}(\Z_\set{S}) \cdot \prod_{i \in \set{S}} \Lambda_{ii}^{-\frac{1}{2}}
\end{equation}
This is simply an affine transformation of a normal pdf.
Most notably the mpdf evaluated at any point $\X_\set{S} = \x$ is independent of $\z_\nset{S}$.

Using this, the manifold entropy \cref{eq:manifold entropy latent} over $\set{S}$ becomes:
\begin{equation}
    H(q_\set{S}) = \frac{|\set{S}|}{2}(\log(2\pi) + 1) + \sum_{i \in \set{S}} \frac{1}{2} \log(\Lambda_{ii})
\end{equation}
and the manifold mutual information \cref{eq:manifold mutual information latent} between $\set{S}$ and $\set{T}$ becomes:
\begin{equation}
    \mathcal{I}(q_\set{S}, q_\set{T}) = \sum_{i \in \set{S}} \frac{1}{2} \log(\Lambda_{ii}) + \sum_{j \in \set{T}} \frac{1}{2} \log(\Lambda_{jj}) - \sum_{k \in \set{ST}} \frac{1}{2} \log(\Lambda_{kk}) = 0
\end{equation}
The latter vanishes because the volume of the Jacobian over the joint index set $\set{ST}$ decomposes into the respective terms over $\set{S}$ and $\set{T}$ as $|\J_\set{ST}| = |\J_\set{S}| |\J_\set{T}|$ which is equivalent with $\J_\set{S} \perp \J_\set{T}$.

Evaluating just for a single latent dimension $\set{S}=i$, the manifold entropy becomes
\begin{equation}
    H(q_i) = \frac{1}{2}(\log(2\pi) + 1) + \frac{1}{2} \log(\Lambda_{ii})
\end{equation}
which is proportional to $\Lambda_{ii}$, thus fulfilling the \textbf{local alignment} condition:
\begin{equation}
    H(p_i) > H(p_j) \quad \forall i<j
\end{equation}

Similarly comparing two single latent dimensions $\set{S}=i$ and $\set{T}=j$, the manifold mutual information becomes
\begin{equation}
    \mathcal{I}(q_i, q_j) = 0
\end{equation}
equivalently to $\J_i \perp \J_j$, thus fulfilling the \textbf{local disentanglement} condition:
\begin{equation}
    I(p_i^*, p_j^*) = 0 \quad \forall i \neq j
\end{equation}

\subsection{Linking local geometric measures to global entropic metrics}

Reformulating the properties of alignment and disentanglement via the manifold entropic metrics allows us to link \textbf{locally} defined geometric measures, like the length $|\J_i|$ or the orthogonality $\J_i \perp \J_j$, to \textbf{globally} defined entropic measures, like the absolute information $H(q_i)$ or the shared information $\mathcal{I}(q_i, q_j)$.

Generalizing to a non-linear NF with arbitrary decoder $\g(\z)$, the Jacobian can be evaluated everywhere and the manifold entropic metrics can be readily obtained.

How one should set the partition over the index sets $\set{S} \in \mathcal{P}$ is arbitrary.
In the simplest case, one can just split the latent space into $D$ latent dimensions.
Without loss of generality we assume we have an ordered partition of index sets with $\set{S} < \set{T} \text{ s.t. } \forall i \in \set{S}, j \in \set{T}: i < j$.
Formulating the conditions for DRL in terms of our global metrics equates to
\begin{enumerate}
    \item \textbf{Alignment:} $H(q_\set{S}) \gg H(q_\set{T}) \ \forall \set{S} < \set{T}$ 
    \item \textbf{Disentanglement}: $\mathcal{I}(q_\set{S}, q_\set{T}) \approx 0 \ \forall \set{S},\set{T}$
\end{enumerate}

\subsection{Geometric perspective of the Manifold Pairwise Mutual Information} \label{sec:Geometric perspective of the Manifold Pairwise Mutual Information}

The MPMI was defined as:
\begin{equation}
\begin{split}
    \mathcal{I}_{ij}(q_i, q_j) &= \Expt{\z}{\log \left| \J_i(\z) \right| + \log \left| \J_j(\z) \right| - \log \left| \J_{\{i, j\}}(\z) \right|} = \Expt{\z}{\frac{1}{2}\log \left( \frac{|\J_{i}|^2 |\J_{j}|^2}{ \left| \J_{\{i,j\}} \right|^2 } \right)}
\end{split}
\end{equation}

We can simplify:
\begin{equation}
    \left| \J_{\{i,j\}} \right| = \det\left( \J_{\{i,j\}}^T \J_{\{i,j\}} \right)^{\frac{1}{2}} = \det\begin{pmatrix}
        |\J_{i}|^2 & \J_i^T \J_j \\ \J_j^T \J_i & |\J_{j}|^2
    \end{pmatrix}^{\frac{1}{2}} = \left( |\J_{i}|^2 |\J_{j}|^2 - \left( \J_i^T \J_j \right)^2 \right)^\frac{1}{2}
\end{equation}

Plugging this in the above equation results in
\begin{equation}
\begin{split}
    \mathcal{I}_{ij}(q_i, q_j)
    &= \Expt{\z}{\frac{1}{2}\log \left( \frac{|\J_{i}|^2 |\J_{j}|^2}{ |\J_{i}|^2 |\J_{j}|^2 - \left( \J_i^T \J_j \right)^2 } \right)} = \Expt{\z}{-\frac{1}{2}\log \left( 1 - \frac{\left( \J_i^T \J_j \right)^2}{|\J_{i}|^2 |\J_{j}|^2} \right)} \\
    &= \Expt{\z}{-\frac{1}{2}\log \left( 1 - \left( \frac{\J_i \cdot \J_j}{|\J_{i}| |\J_{j}|} \right)^2 \right)}
\end{split}
\end{equation}

Finally we use the definition of the Cosine-similarity between the two vectors $\J_{i}$ and $\J_{j}$:
\begin{equation}
    \cos(\theta_{i,j}) \coloneqq \frac{\J_i \cdot \J_j}{|\J_{i}| |\J_{j}|}
\end{equation}
and rewrite to
\begin{equation}
\begin{split}
    \mathcal{I}_{ij}(q_i, q_j)
    &= \Expt{\z}{-\frac{1}{2}\log \left( 1 - \cos(\theta_{i,j})^2 \right)}
\end{split}
\end{equation}

We can make some approximation for two special cases:
\begin{enumerate}
    \item If both vectors are almost orthogonal to each other, we can define the angle $\alpha_{i,j} \coloneqq \theta_{i,j} - \frac{\pi}{2} \ll 1$ such that we can write $\cos(\theta_{i,j}) = \cos(\alpha_{i,j} + \frac{\pi}{2}) = -\sin(\alpha_{i,j})$. This allows us to make the following approximation wrt. $\alpha_{i,j}$:
    \begin{equation}
        \mathcal{I}_{ij}(q_i, q_j) = \Expt{}{-\frac{1}{2}\log \left( 1 - \sin(\alpha_{i,j})^2 \right)} \approx \frac{1}{2}\Expt{}{\alpha_{i,j}^2}
    \end{equation}
    In this approximation the MPMI scales with the (squared) deviation from "full" orthogonality of two latent manifolds.
    \item On the other hand if both vector are almost collinear to each other, we can write $\cos(\theta_{i,j})^2 \approx 1 - \theta^2$ and make the following approximation:
    \begin{equation}
        \mathcal{I}_{ij}(q_i, q_j) \approx \Expt{}{-\frac{1}{2}\log \left( 1 - (1 - \theta_{i,j}^2) \right)} = \Expt{}{-\log \left( \theta_{i,j} \right)}
    \end{equation}
    In this approximation the MPMI scales logarithmically with the angle between two latent manifolds and goes to infinity for $ \theta_{i,j} \to 0$.
\end{enumerate}

\newpage

\section{NUMERICS} \label{sec:NUMERICS}

\newcommand{\B}{\boldsymbol{B}}

\subsection{Computation of Manifold Entropic Metrics}

Let us highlight how the manifold entropic metrics are computed in practice.
For that let us assume that we wish to evaluate the manifold entropic metrics only over a fixed index set $\set{S}$.
Remember that the manifold entropy over $\set{S}$ reads $H(q_\set{S}) = \text{const.} + \Expt{\z}{\log\left|\J_{\set{S}}\right| }$ and that over $i \in \set{S}$ reads $H(q_i) = \text{const.} + \Expt{\z}{\log\left|\J_{i}\right| }$.
The manifold total correlation can be expressed in terms of these.

The expectation over $\z \sim p(\z)=\mathcal{N}(0, \I_D)$ is evaluated as a empirical average over a batch of samples $\z^{\B} \coloneq \{\z^j\}_{j=1}^{B}$ where $B$ denotes the batchsize.
To evaluate the submatrix $\J_\set{S}$ over the batch we can follow one of two approaches, where only the first one is presented in the main paper:

\paragraph{vector-Jacobian products with reverse-mode auto-differentiation:}%

We can compute the full decoder Jacobian $\J(\z)$ of the batch via vector-Jacobian products (vjp) and then take the slice of $\set{S}$ dimension to obtain $\J_\set{S}$.
For this the batch of samples $\z^{\B}$ is pushed through the decoder producing $\x^{\B} = \g(\z^{\B})$ and the batch of vjp-s $\J_{k,\cdot}(\z^{\B}) = \boldsymbol{e}_k^T \J(\z^{\B})$ is computed once for each data dimension $k \in \{1,...,D\}$.
$\boldsymbol{e}_k$ is the standard basis vector for index $k$ and $\J_{k,\cdot}$ explicitly denotes the $k$-th row-vector of $\J$.
Concatenating all vjp-s $\{\J_{k,\cdot}(\z^{\B}) \}_{k=1:D}$ along the first dimension leads to the batch of Jacobians $\J(\z^{\B})$.
This procedure is relatively fast via reverse-mode auto-differentiation but requires to compute the full Jacobian at first by looping over $D$ dimensions, which can become computationally infeasible for very high-dimensional problems.

\paragraph{Jacobian-vector products with forward-mode auto-differentiation:}%

Alternatively we can compute only the required submatrix $\J_\set{S}(\z)$ via Jacobian-vector products (jvp).
For this we have to iterate over all latent dimensions $i \in \set{S}$ and transform $\boldsymbol{e}_i$ alongside the batch $\z^{\B}$ through the decoder to obtain the batch of jvp-s $\{\J_i(\z^{\B}) = \J(\z^{\B}) \boldsymbol{e}_i \}$.
Concatenating all jvp-s $\{\J_{i}(\z^{\B}) \}_{i\in \set{S}}$ along the second dimension leads to the batch of Jacobians $\J_\set{S}(\z^{\B})$.
This procedure is slightly slower via forward-mode auto-differentiation but only requires to compute the necessary Jacobian submatrix by looping over $|\set{S}|$ dimensions.

\obso{
Minimal example:\\
Sample $\z^{\B}$ from $p(\z)$
Initialize $\J(\z^{\B})$ as empty array\\
For $k \in \{1,...,D\}$ do:\\
\phantom{MM} $\g(\z^{\B}), \boldsymbol{e}_k^T \J(\z^{\B}) = \texttt{vjp}(\g, \z^{\B}, \boldsymbol{e}_k)$\\
\phantom{MM} $\J_{k,\cdot}(\z^{\B}) = \boldsymbol{e}_k^T \J(\z^{\B})$ \\
}
\obso{
Minimal example: \\
Sample $\z^{\B}$ from $p(\z)$
Initialize $\J_\set{S}(\z^{\B})$ as empty array\\
For $i \in \set{S}$ do:\\
\phantom{MM} $\g(\z^{\B}), \J(\z^{\B}) \boldsymbol{e}_i = \texttt{jvp}(\g, \z^{B}, \boldsymbol{e}_i)$\\
\phantom{MM} $\J_{i}(\z^{\B}) = \J(\z^{\B}) \boldsymbol{e}_i$\\
}

Minimal implementation examples of both methods in pytorch \citep{paszke2019pytorch} are provided at the end of this section via the functions \texttt{Jacobian\_reverse()} and \texttt{Jacobian\_forward()} respectively.
In our experiments we mostly opted for the first option as we used the full Jacobian anyway.

The volume (i.e. determinant) of the Jacobians are obtained by standard matrix multiplication.
This method is only feasible at test time, and would pose a bottleneck for training high-dimensional problems.
However, the literature offers numerous approximation methods for log-determinant computations \citep{han2015largescalelogdeterminantcomputationstochastic, boutsidis2016randomizedalgorithmapproximatinglog, deen2024fastaccuratelogdeterminantapproximations}, which could be adapted for such scenarios.
Additionally, recent advances in training generative models such as \cite{draxler2024freeform} demonstrate success with exact maximum likelihood training for unrestricted architectures via a tractable unbiased estimator of the derivative of the log-determinant-Jacobian. Techniques such as that could potentially make our method applicable during training with appropriate modifications.

\subsection{Analysis of Normalizing Flows}

\begin{figure}[!htbp]
    \centering
    \includegraphics[width=0.7\linewidth]{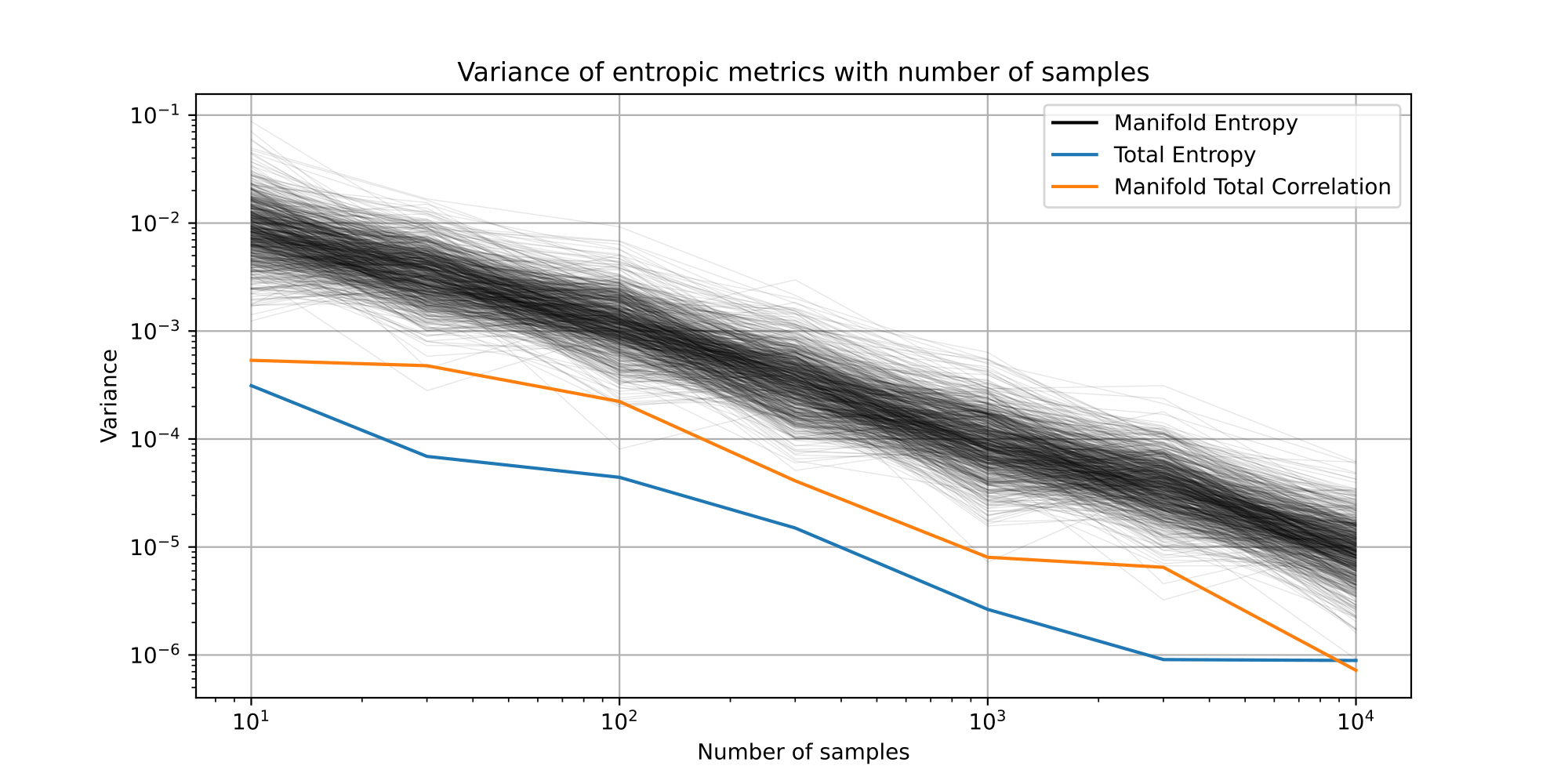}
    \caption{Variance over 10 runs of manifold entropic metrics decrease with increasing sample size.}
    \label{fig:convergence of variance}
\end{figure}

We primarily applied our metrics on NFs trained on EMNIST with an image resolution of $1 \times 28 \times 28$ ($D = 784$) and an equally sized latent space.
The final estimates converge quickly and stably with increasing sample size. \obso{and in practice it was sufficient to use $100-1000$ samples for an accurate result.}
In an experiment on a trained NF on EMNIST we observe that the variance over 10 runs is decreasing with an increasing number of samples, see \cref{fig:convergence of variance}.

As the metrics are essentially probing the trained model, we expect that one has to increase the number of samples according to the complexity of the dataset in order to obtain accurate results.

Evaluating the MPMI matrix $\mathcal{I}_{ij}(q_i, q_j)$, using a pre-computed Jacobian of 1000 images, takes 78s on a NVIDIA RTX 4090 GPU.
This rather slow inference is because the number of evaluations (most notably of $|\J_{ij}|$) scales with $\frac{D(D-1)}{2}$. Computing the Manifold Mutual Information of a partition with less subsets would be accordingly faster.

\subsection{Analysis of BigGAN}

We also conducted preliminary experiments on GANs to explore feasibility for higher-dimensional problems.
For this we analyzed a pretrained BigGAN\footnote{\url{https://github.com/huggingface/pytorch-pretrained-BigGAN}}, which was also used in \cite{voynov2020unsuperviseddiscoveryinterpretabledirections}, with an image resolution of $3 \times 256 \times 256$ and a latent dimensionality of $128$.

On a NVIDIA RTX 4090 GPU, computing the full Jacobian $\J \in \mathbb{R}^{128 \times 196608}$ for a batch of $8$ samples takes $\approx 15$ seconds.
Repeating this process $8$ times yields Jacobians for a total of $64$ samples in $\approx 2$ minutes. Following that, the computation of the Total Entropy requires $\approx 1$ second, the Manifold Entropy over all individual 128 latent dimensions takes $\approx 3$ seconds, and the MPMI over all $128 \times 128$ entries completes in $\approx 7-8$ seconds on the GPU ($\approx 1.5$ minutes on CPU). Scaling this procedure up to even larger models (data and/or latent space size) would be limited by the GPU memory and number of samples taken for evaluation.

Finally, we select the latent dimensions with the highest manifold entropy and observe that they indeed have a stronger influence on the generated image (e.g. zoom-in, left-right pan) while those with a lower manifold entropy encode less important variations. However, as BigGAN wasn’t trained to perform well at Disentanglement, we also observe that many latent variables are responsible for varying semantically similar variations in the data, i.e. they are entangled.

\subsection{Python Implementation} \label{sec:Python Implementation}
\vfill
\begin{lstlisting}[caption={Minimal implementation examples of the Jacobian computation using reverse and forward auto-differentiation in pytorch. Nomenclature: \texttt{g} is the decoder function $\g$, \texttt{batchsize} is $B$, \texttt{dim\_x} is the dimensionality of $\x$, \texttt{dim\_z} is the dimensionality of $\z$, \texttt{dims\_set} is the index set $\set{S}$, \texttt{J} is $\J$ and \texttt{J\_set} is $\J_{\set{S}}$. Since in most generative models \texttt{dim\_z} is smaller than \texttt{dim\_x}, the forward mode should be preferred..}]
import torch
from torch.autograd import grad
from torch.autograd.forward_ad import dual_level, make_dual, unpack_dual

def Jacobian_reverse(g, batchsize, dim_x, dim_z):
    z = torch.randn(batchsize, dim_z).requires_grad(True)
    x = g(z)
    J = torch.zeros(batchsize, dim_z, dim_x)
    for i in range(dim_x):
        J[:,:,i] = grad(x[:,i].sum(), z, create_graph=True)[0].detach()
    return J

def Jacobian_forward(g, batchsize, dim_x, dim_z, dims_set=None):
    if dims_set == None:
        dims_set = torch.arange(dim_z)
    z = torch.randn(batchsize, dim_z).requires_grad(True)
    J_set = torch.zeros(batchsize, len(dims_set), dim_x)
    for j_idx, j in enumerate(dims_set):
        z_grad = torch.zeros_like(z)
        z_grad[:,j] = 1
        with dual_level():
            dual_z = make_dual(z, z_grad)
            dual_x = g(dual_z)
            x, x_grad = unpack_dual(dual_x)
        J_set[:,j_idx,:] = x_grad.detach()
    return J_set
\end{lstlisting}

\newpage

\section{PROOF OF CONCEPT} \label{sec:PROOF OF CONCEPT}

\subsection{Two Moons} \label{sec:two moons}
\obso{
We sample from the two moons dataset with Gaussian noise variance $\sigma_{\text{noise}}^2 = 0.01$.
}
We split the latent space into one important dimension $c$ and one unimportant dimension $d$ such that the latent vector becomes $\z = [z_c, z_d]$.

Alignment is achieved for $H(q_c) \gg H(q_d)$:
If the manifold entropy in $c$ is much greater than that in $d$ or in other words if most of the information is absorbed by the manifold random variable $\X_c$.
Geometrically this can be seen if the latent manifold of $c$ spans both moons and that of $d$ only models the orthogonal noise.

Disentanglement is achieved for $\mathcal{I}(q_c, q_d) \approx 0$:
If the mutual information between $c$ and $d$ vanishes or in other words if both manifold random variables $\X_c$ and $\X_d$ don't share information.
Geometrically this can be seen if the latent manifolds of $c$ and $d$ intersect orthogonally everywhere.

\subsubsection{Architecture}

We implement the NFs in FrEIA \citep{freia} via 8 RQS-blocks \citep{durkan2019neural} with 4 bins each and one final orthogonal rotation matrix (adapted from \cite{ross2021tractable}) before the latent space.
\subsubsection{Training} \label{sec:two moons training}
We train a NF once for each training objective $\mathcal{L}$ as
\begin{enumerate}
    \item[(A)] Maximum Likelihood: $\mathcal{L}(\x) = \mathcal{L}_{\text{ML}}(\x)$
    
    The Maximum Likelihood loss for a sample $\x$ is:
    \begin{equation}
        \mathcal{L}_{\text{ML}}(\x) = \frac{1}{2} \left|\f(\x)\right|^2 - \log\left|\frac{\partial \f(\x)}{\partial \x}\right|
    \end{equation}
    $\left|\frac{\partial \f(\x)}{\partial \x}\right|$ is the volume of the encoder Jacobian and cheaply evaluated alongside the forward pass through the encoder.
    It is identical to the inverse of the volume of the decoder Jacobian $|\J(\f(\x))|$.
    \item[(B)] Maximum Likelihood with additional regularization to minimize the estimated "manifold total correlation" $\mathcal{I}(\x)$ \footnote{i.e. instead of sampling $\z \sim p(\z)$, we sample data samples $\x \sim p^*(\x)$ and push through the encoder $\z = \f(\x)$} with a weighting of $1$: $\mathcal{L}(\x) = \mathcal{L}_{\text{ML}}(\x) + 1 \cdot \mathcal{I}(\x)$
    
    The "manifold total correlation" loss, adapted from $L(\theta)$ eq.(14) in \citep{pmlr-v162-cunningham22a}, reads:
    \begin{equation}
        \mathcal{I}(\x) = \log\left|\J_c(\f(\x))\right| + \log\left|\J_d(\f(\x))\right| - \log\left|\J(\f(\x))\right|
    \end{equation}
    The volume terms $|\J_c|$ and $|\J_d|$ are evaluated using the full Jacobian $\J$ at $\z = \f(\x)$, via two vjp-s or jvp-s. The last term $|\J|$ cancels with that from the Maximum Likelihood loss.
    \item[(C)] Maximum Likelihood with additional regularization to minimize the reconstruction loss as in \cite{NEURIPS2021_4c07fe24} with a weighting of $5$: $\mathcal{L} = \mathcal{L}_{\text{ML}} + 5 \cdot \mathcal{L}_\text{rec}$
    
    The reconstruction loss is defined as:
    \begin{equation}
        \mathcal{L}_\text{rec}(\x) = \frac{1}{2} \left| \x - \g([z_c=f_c(\x), z_d=0]) \right|^2
    \end{equation}
    To evaluate it we only have to make one additional pass through the decoder.
\end{enumerate}

\newpage

\subsection{10-D Torus}

The transformation from azimuthal $\ph$ and radial $\r$ to cartesian $\x$ coordinates is defined $\forall i\in \{0,...,19\}$ as
\begin{equation}
    x_i = \begin{cases}
    r_{\lfloor i/2 \rfloor}\cos(\varphi_{\lfloor i/2 \rfloor}),& \text{if } i \text{ is even}\\
    r_{\lfloor i/2 \rfloor}\sin(\varphi_{\lfloor i/2 \rfloor}),& \text{if } i \text{ is odd}
    \end{cases}
\end{equation}

where $\ph, \r \in \mathbb{R}^{10}$ are reparameterized via $\z = [\z_{\set{C}}, \z_{\nset{C}}] \in \mathbb{R}^{20}$ and the (fixed) standard deviations $\sig^{(\varphi)}, \sig^{(r)} \in \mathbb{R}^{10}$
\begin{align}
    \ph(\z) &= \sig^{(\varphi)}  \odot \z_{\set{C}} \\
    \r(\z) &= \mathbf{1} +  \sig^{(r)} \odot \z_{\nset{C}}
\end{align}
where $\odot$ denotes component-wise multiplication.

\begin{figure}[!htbp]
    \centering
    \includegraphics[width=.5 \textwidth]{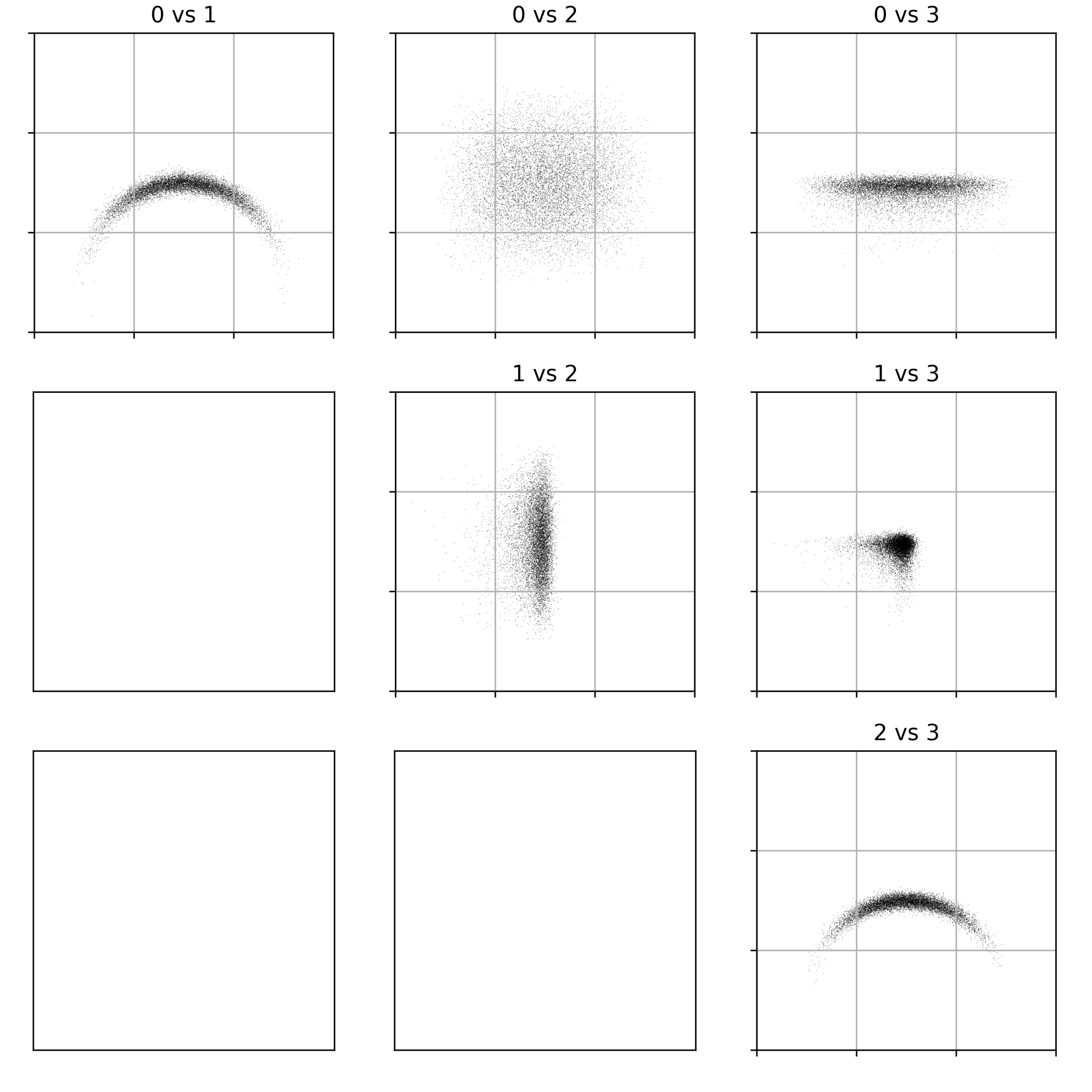}
    \caption{2D pairwise-projections of the first four data dimensions $\{x_0, x_1, x_2, x_3\}$ of the 10-D Torus dataset.}
    \label{fig:toy-dataset projections}
\end{figure}

\subsubsection{Analytical Jacobian} \label{sec:analytical jacobian}

Ultimately, data samples are produced by pushing $\z$ through an ''analytical decoder'' function $\Phi$:
\begin{equation} \label{analytical decoder}
    \x = \Phi(\ph(\z), \r(\z))
\end{equation}
where $\z$ is sampled from a standard normal $\z \sim \mathcal{N}(\z| 0, \I_{20})$.

Using this formulation, we can compute the Jacobian of $\Phi$ explicitly
\begin{equation}
    \Jphi(\z) \coloneq \frac{\partial \Phi(\z')}{\partial \z'} \Big|_{\z'=\z}
\end{equation}
Furthermore it can be split into a $\set{C}$ and $\nset{C}$ part:
\begin{equation}
    \Jphi = 
    \begin{pmatrix}
        \Jphiu{\set{C}}  , & \Jphiu{\nset{C}}
    \end{pmatrix} =
    \begin{pmatrix}
        \frac{\partial \Phi}{\partial \z_{\set{C}}}  , & \frac{\partial \Phi}{\partial \z_{\nset{C}}}
    \end{pmatrix}
\end{equation}
with
\begin{align}
    \Jphiu{\set{C}} &= \frac{\partial \Phi}{\partial \ph} \frac{\partial \ph}{\partial \z_{\set{C}}}
    + \frac{\partial \Phi}{\partial \r} \underbrace{\frac{\partial \r}{\partial \z_{\set{C}}}}_{=0} \\
    \Jphiu{\nset{C}} &= \frac{\partial \Phi}{\partial \ph} \underbrace{\frac{\partial \ph}{\partial \z_{\nset{C}}}}_{=0} + \frac{\partial \Phi}{\partial \r} \frac{\partial \r}{\partial \z_{\nset{C}}}
\end{align}

\newcommand{\itf}{{\lfloor i/2 \rfloor}}
Let us first calculate the partial derivative of $\Phi$ towards $\ph$ and $\r$:
\begin{align}
    \frac{\partial \Phi_i}{\partial \varphi_j} &= \delta_{\itf, j} \cdot
    \begin{cases}
    -r_\itf\sin(\varphi_\itf),& \text{if } i \text{ is even}\\
    r_\itf\cos(\varphi_\itf),& \text{if } i \text{ is odd}
    \end{cases} \\
    \frac{\partial \Phi_i}{\partial r_j} &= \delta_{\itf, j} \cdot
    \begin{cases}
    \cos(\varphi_\itf),& \text{if } i \text{ is even}\\
    \sin(\varphi_\itf),& \text{if } i \text{ is odd}
    \end{cases}
\end{align}
And now the partial derivative of $\ph$ and $\r$ towards $\z_{\set{C}}$ and $\z_{\nset{C}}$:
\begin{align}
    \frac{\partial \varphi_j}{\partial {z}_k} &=
    \begin{cases}
        \delta_{jk} \sigma^{(\varphi)}_k,& \text{if } k \in \set{C}\\
        0 ,& \text{if } k \in \nset{C}
    \end{cases} \\
    \frac{\partial r_j}{\partial z_k} &=
    \begin{cases}
        0,& \text{if } k \in \set{C}\\
        \delta_{j,(k-10)} \sigma^{(r)}_{(k-10)} ,& \text{if } k \in \nset{C}
    \end{cases}
\end{align}

Combining all we can write $\Jphiu{\set{C}}$ with $k \in \set{C}$ as:
\begin{equation}
    \frac{\partial \Phi_i}{\partial \z_k} = \delta_{\itf, k} \cdot \sigma_k^{(\varphi)} \cdot
    \begin{cases}
    (-r_\itf\sin(\varphi_\itf)),& \text{if } i \text{ is even}\\
    r_\itf\cos(\varphi_\itf),& \text{if } i \text{ is odd}
    \end{cases}
\end{equation}
and $\Jphiu{\nset{C}}$ with $k \in \nset{C}$ as:
\begin{equation}
    \frac{\partial \Phi_i}{\partial \z_k} = \delta_{\itf, (k-10)} \cdot \sigma^{(r)}_{(k-10)} \cdot
    \begin{cases}
    \cos(\varphi_\itf),& \text{if } i \text{ is even}\\
    \sin(\varphi_\itf),& \text{if } i \text{ is odd}
    \end{cases}
\end{equation}

\subsubsection{Ground-truth Manifold Entropic Metrics}

To obtain the ground-truth data manifold entropy for each latent dimension we must first compute $\Jphi^T \Jphi$:
\begin{equation}
    \Jphi^T \Jphi = \begin{bmatrix}
        {\Jphiu{\set{C}}}^T {\Jphiu{\set{C}}} & {\Jphiu{\set{C}}}^T {\Jphiu{\nset{C}}} \\
        {\Jphiu{\nset{C}}}^T {\Jphiu{\set{C}}} & {\Jphiu{\nset{C}}}^T {\Jphiu{\nset{C}}}
    \end{bmatrix}
\end{equation}
The upper left block-matrix ${\Jphiu{\set{C}}}^T {\Jphiu{\set{C}}} \in \mathbb{R}^{10 \times 10}$ with $j, k \in \set{C}$:
\begin{equation}
\begin{split}
     \left({\Jphi}^T {\Jphi}\right)_{jk} &= \sum_{i=0}^{19} {\Jphi_{ij} {\Jphi}_{ik}} =
     \sum_{i=0}^{9} \delta_{i j} \delta_{i k} \cdot \sigma^{(\varphi)}_j \sigma^{(\varphi)}_k \cdot (r_i^2 \sin(\varphi_i)^2 + r_i^2 \cos(\varphi_i)^2) \\
     &= \delta_{jk} \cdot {\sigma^{(\varphi)}_j}^2 \cdot \sum_{i=0}^{9} \left(1 + \sigma^{(r)}_i z^{r}_i\right)^2
\end{split}
\end{equation}
Assuming that $\sigma_i^{(r)} \ll 1 \ \forall i$, \footnote{As $\z$ is sampled from the latent prior with $\Expt{\z}{\z_i^2}=1$ it shouldn't be a problem in practice.}, which is valid in our experiments, we approximate each $\left(1 + \sigma^{(r)}_i z^{r}_i\right)^2 \approx 1$ such that
\begin{equation}
    \left({\Jphi}^T {\Jphi}\right)_{jk} \approx 10 \cdot \delta_{jk} \cdot \left({\sigma^{(\varphi)}_j}\right)^2
\end{equation}

The lower right block-matrix ${\Jphiu{\nset{C}}}^T {\Jphiu{\nset{C}}} \in \mathbb{R}^{10 \times 10}$ with $j, k \in \nset{C}$:
\begin{equation}
\begin{split}
   \left({\Jphi}^T {\Jphi}\right)_{jk} &= \sum_{i=0}^{19} {\Jphi_{ij} {\Jphi}_{ik}} =
   \sum_{i=0}^{9} \delta_{i,(j-10)} \delta_{i,(k-10)} \cdot \sigma^{(r)}_{j-10} \sigma^{(r)}_{k-10} \cdot (\cos(\varphi_i)^2 + \sin(\varphi_i)^2) \\
    &= 10 \cdot \delta_{(j-10),(k-10)} \cdot \left({\sigma^{(r)}_{(j-10)}}\right)^2
\end{split}
\end{equation}

The off-diagonal block matrix ${\Jphiu{\set{C}}}^T {\Jphiu{\nset{C}}} \in \mathbb{R}^{10 \times 10}$ and its transpose ${\Jphiu{\nset{C}}}^T{\Jphiu{\set{C}}} \in \mathbb{R}^{10 \times 10}$ with $j \in \set{C}, k \in \nset{C}$:
\begin{equation}
\begin{split}
    \left({\Jphi}^T {\Jphi}\right)_{jk} &= \sum_{i=0}^{19} {\Jphi_{ij} {\Jphi}_{ik}}\\
    &= \sum_{i=0}^{19} \delta_{i, j} \delta_{i, (k-10)} \sigma^{(\varphi)}_j \sigma^{(r)}_{k-10} \cdot \underbrace{(-r_i \sin(\varphi_i) \cos(\varphi_i) + r_i \cos(\varphi_i) \sin(\varphi_i))}_{=0}\\
    &= 0
\end{split}
\end{equation}

Therefore the $\Jphi^T \Jphi$-matrix has only diagonal entries:
\begin{equation}
    \Jphi^T \Jphi = 10 \cdot \text{diag}\left( {\sig^{(\varphi)}}^2, {\sig^{(r)}}^2 \right)
\end{equation}

With this we can finally compute the volume of each $\Jphi_i$ to
\begin{equation}
    \left| \Jphi_i \right| = \det \left( \Jphi_i^T \Jphi_i \right)^{\frac{1}{2}} = 10 \cdot
    \begin{cases}
        {\sigma_i^{(\varphi)}},& \text{if } i \in \set{C}\\
        {\sigma_i^{(r)}},& \text{if } i \in \nset{C}
    \end{cases}
\end{equation}
and the volume of $\Jphi$ is simply
\begin{equation} \label{eq:toy jac volumes decomposition}
    \left| \Jphi \right| = \prod_{i=0}^{19} \left| \Jphi_i \right|
\end{equation}

We now evaluate the metrics over each latent dimension $i \in \{0,...,19\}$.

Using \cref{eq:manifold entropy latent} we can evaluate the manifold entropy over the ground-truth manifold pdf $q^{\Phi}$ of one latent dimension $i$ to
\begin{equation}
    H(q^{\Phi}_i) = \frac{1}{2}(\log(2 \pi) + 1) + \Expt{\z}{\log\left| \Jphi_i \right|} = \frac{1}{2}(\log(2 \pi) + 1) + \log(10) +
    \begin{cases}
        \log({\sigma_i^{(\varphi)}}) ,& \text{if } i \in \set{C}\\
        \log({\sigma_i^{(r)}}) ,& \text{if } i \in \nset{C}
    \end{cases}
\end{equation}
where the expectation over $\z$ vanishes because $|\Jphi_i|$ is a constant.

We can see that the ground-truth manifold entropy is proportional to $\log(\sigma_i)$.

Using \cref{eq:manifold total correlation decomposition} and \cref{eq:toy jac volumes decomposition} the manifold total correlation can be evaluated to
\begin{equation}
    \mathcal{I} = \sum_{i=0}^{19} H(q^{\Phi}_i) - H(q^{\Phi}) = \sum_{i=0}^{19} \log\left| \Jphi_i \right| - \log\left| \Jphi \right| = 0
\end{equation}

\subsection{Pearson Cross-Correlation matrix} \label{sec:Pearson Cross-Correlation matrix}

To have a comparison of our novel MCPMI metric with a (classical) linear metric, we additionally computed and plotted the Pearson Cross-correlation matrix which is the covariance between each pair of {\em ground truth} vs {\em predicted} latent variables.

Concretely we sample random variables $\z^\text{gt}$ from a standard normal, push them through the analytical decoder eq.(\ref{analytical decoder}) to obtain $\x^\text{gt}$.
Now to obtain the {\em predicted} latent representation of the NF $\z^\text{pred}$ we push $\x^\text{gt}$ through the normalizing flow encoder eq.(\ref{eq: NF encoder}) and sort the latent variables by the magnitude of their individual Manifold Entropy. \footnote{The last resorting step serves only to visually reorganize the final evaluation matrix so that the entries are more easily comparable across columns and rows.}

The Pearson Cross-correlation is defined as
\begin{equation}
    \mathcal{R}_{ij}(z^\text{gt}_i,z^\text{pred}_j) \coloneqq \frac{\Expt{}{ \left(\z_i^\text{gt} - \Expt{}{\z_i^\text{gt}} \right) \cdot \left(\z_j^\text{pred} - \Expt{}{\z_j^\text{pred}}\right) }}{ \Expt{}{\left( \z_i^\text{gt} - \Expt{}{\z_i^\text{gt}} \right)^2} \cdot \Expt{}{\left(\z_j^\text{pred} - \Expt{}{\z_j^\text{pred}}\right)^2}}
\end{equation}
Note that the expectation goes over $\z^\text{gt}$.
Furthermore as $\z^\text{gt}$ is sampled from a standard normal distribution and $\z^\text{pred}$ should follow one as well, as the NF is trained with the same latent prior, both variables are already normalized. Thus we can simplify:
\begin{equation}
    \mathcal{R}_{ij}(z^\text{gt}_i,z^\text{pred}_j) = \Expt{}{ \z_i^\text{gt} \cdot \z_j^\text{pred} } = \Expt{}{ \z_i^\text{gt} \cdot f_j(\x^\text{gt}) } = \Expt{}{ \z_i^\text{gt} \cdot f_j(\Phi(\ph(\z^\text{gt}), \r(\z^\text{gt}))) }
\end{equation}
Finally we plot the elementwise squared Pearson correlation matrix because we are only interested in the magnitude of the correlation and not its sign.

By the above formulation we can see that this metric is defined as a distance in the latent space and is measuring to which degree the NF encoder $\f(\cdot)$ can invert the effect of the analytical decoder $\Phi(\ph(\cdot), \r(\cdot))$ for each latent dimension. 

In comparison, the MCPMI eq.(\ref{eq:MCPMI}) measures the similarity between the NF decoder and the analytical decoder and more precisely how similar their manifold pdfs are.

\subsubsection{Additional plots}

\begin{figure}[!htbp]
    \centering
    \includegraphics[width=\textwidth]{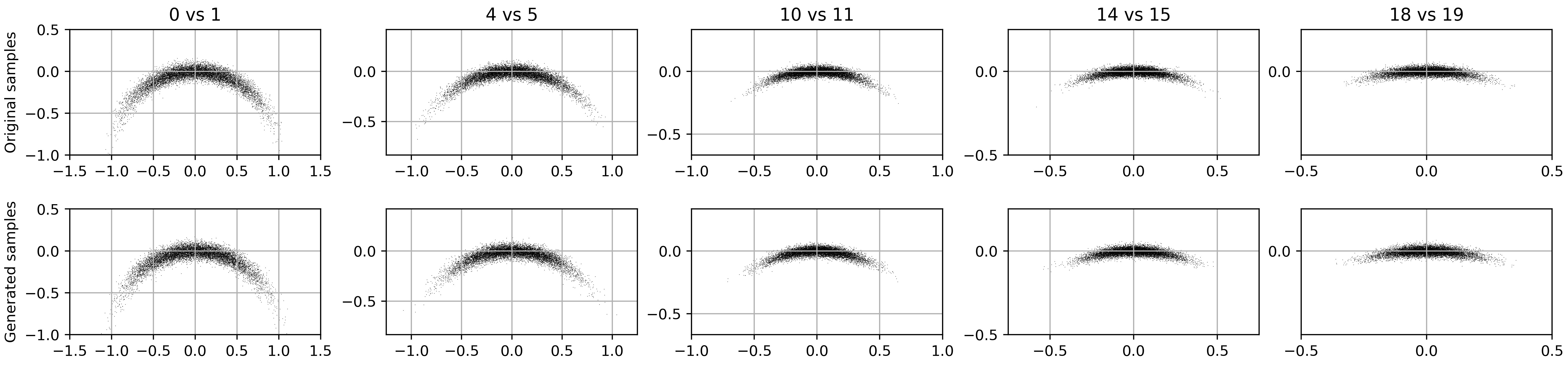}
    \caption{Original samples (top) and generated samples (bottom) visualized as 2D-projections onto two consecutive data-dimension-pairs (0,1), (4,5), (10,11), (14,15) and (18,19) (left to right).
    The length and thickness of the ''arc'' get shorter and thinner with increasing dimension $i$ as $\sigma^{(\varphi)}_i$ and $\sigma^{(r)}_i$ decrease proportionally.}
    \label{fig:toy-dataset DL samples}
\end{figure}

To inspect visually if the data density has been learned correctly, one could generate data samples and plot the projection onto two data dimensions which is infeasible as there are $\frac{20\cdot 19}{2}$ pairs of data-dimensions.
We plot the projections for 5 pairs of data dimensions in \cref{fig:toy-dataset DL samples} once for the original data samples and once for the ones generated by a model trained solely with the Maximum Likelihood objective.

\begin{figure}[!htbp]
    \centering
    \includegraphics[width=\textwidth]{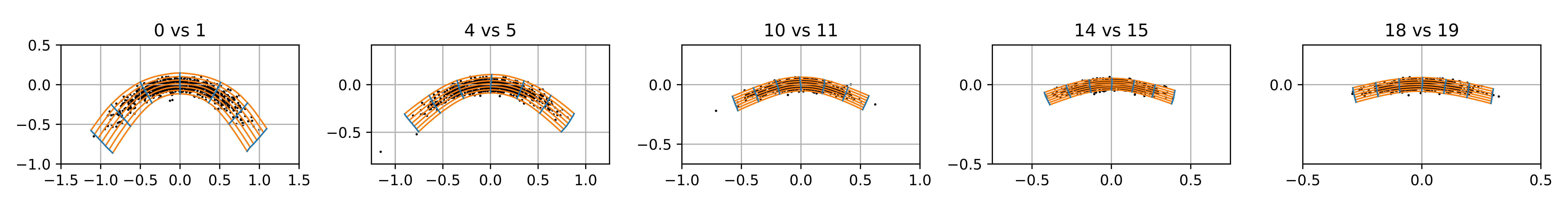}
    \caption{Contour grids of the solution with an additional disentanglement regularization}
    \label{fig:toy-dataset contour grids}
\end{figure}

To inspect visually if the decoder inverts the DGP exactly, on would have to visualize the contour grids for all $\frac{20\cdot 19}{2}$ pairs of latent-dimensions, thus a staggering $\frac{20^2\cdot 19^2}{4}$ plots in total.
To demonstrate visually that the model trained with additional Disentanglement regularization inverts the DGP, we plot the contour grids for some selected latent and data dimensions in \cref{fig:toy-dataset contour grids}.
For that we had to find the $5\cdot 2$ appropriate latent dimensions which model the respective azimuthal and radial variations in the 5 predefined data-dimensions-pairs.

\newpage

\subsubsection{Additional experimental results}
The manifold total correlation of each trained model is as follows:
\begin{itemize}
    \item[-] Vanilla Maximum-Likelihood Training: $\mathcal{I} \approx 0.90 \text{ nats/dim}$
    \item[-] Maximum-Likelihood with additional Disentanglement regularization (see \ref{sec:two moons training}): $\mathcal{I} \approx 0.014 \text{ nats/dim}$
\end{itemize}

We also verified that the approximation of the analytical Jacobian $\Jphi$ in \ref{sec:analytical jacobian} holds by computing it via auto-differentiation as well. For this we had to make sure that the DGP-function \ref{analytical decoder} is written in pytorch and is fully differentiable.
Comparing the manifold entropic metrics obtained with the Jacobian of each method showed no noticeable differences.

\newpage

\section{EXPERIMENTS} \label{sec:EXPERIMENTS}

\subsection{Normalizing Flows}

\subsubsection{Additional reconstruction loss}

The reconstruction loss for a fixed core index set $\set{C}$ is
\begin{equation}
    \mathcal{L}_\text{rec}(\x) = \frac{1}{2} \left| \x - \g([\z_\set{C}=\f_\set{C}(\x), \z_\set{D}=0]) \right|^2
\end{equation}

\begin{figure}[!htbp]
    \centering
    \includegraphics[width=\textwidth]{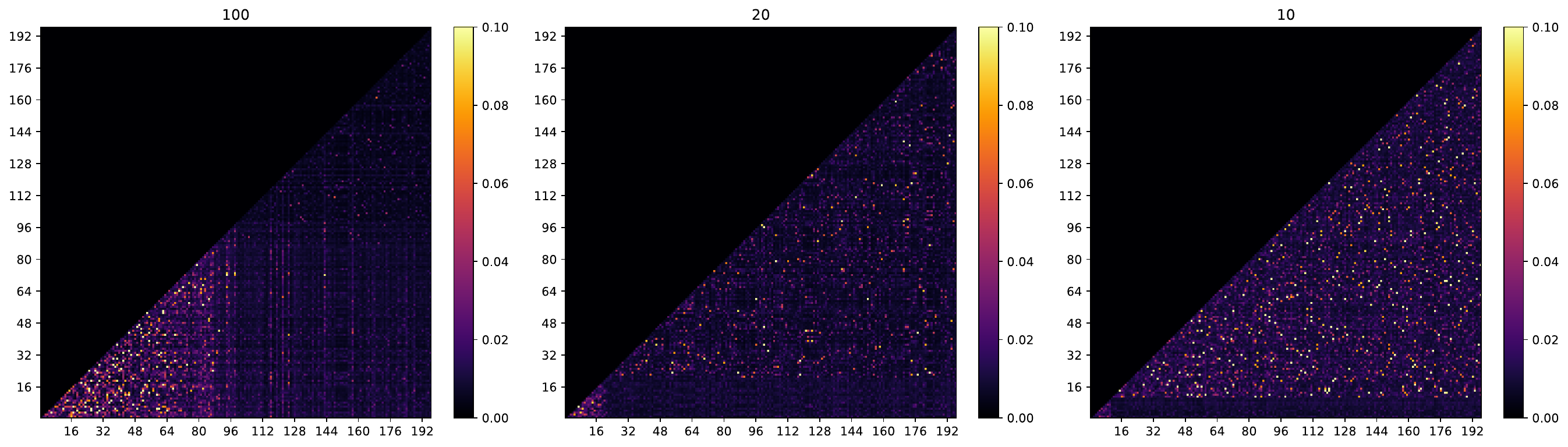}
    \caption{MPMI matrices of three trained NFs with $|\set{C}|= \in \{100, 20, 10\}$ (left to right) depicted for the first $192$ latent dimensions. One can see that the manifold mutual information $\mathcal{I}_{ij}(q_i, q_j)$ with $i\in\set{C}, j\in\nset{C}$ is less compared to the rest. This showcases that an additional reconstruction loss induces disentanglement between the two latent manifolds $\Man_\set{C}$ and $\Man_\nset{C}$.}
    \label{fig:MPMI reconstruction}
\end{figure}

We add $\mathcal{L}_\text{rec}$ to the maximum likelihood loss $\mathcal{L}_\text{ML}$ with a weighting of 1.
Training with this regularized training objective is unstable and the model always diverges, which has already been observed in \cite{cramer2023nonlinear}, thus we were only able to train for a total of 10 epochs (2350 iterations) and cherry-picked non-diverging models for the analysis.
In \cref{fig:MPMI reconstruction} we visualize the MPMI matrix of each model.

\subsubsection{Visualizing Latent Manifolds on EMNIST}
Visualizing the latent manifolds of high-dimensional data, such as images, using a contour grid is infeasible as one would have to project the samples to two dimension at a time in order to plot them.

Thus, usually this is achieved by interpolating a single latent dimension $\z_i$ over a fixed range of values (e.g. $\z_i \in [-2, 2]$), setting all other latent dimensions to their mean, (e.g. $\z_j = 0 \ \forall j \neq i$), and plotting the resulting images.
On one hand, this procedure is interpretable as one can directly see how changing one latent variable affects the resulting image, but on the other hand one needs to plot many images to get a full picture.
Furthermore this is only probing one slice of the latent space at a time, as all other latent dimensions are constant.

Instead, we follow an alternative approach to visualize the influence of one latent dimension on the generated image, using the decoder Jacobian at a sample $\z$.
This procedure could be interpreted as a more direct measurement of the interpolation between different $\z_i$-values.
In PCA one can easily visualize the eigenvectors  $\U_i = \J_i/||\J_i||_2$ (compare with \cref{eq:PCA jacobian i}) as a set of images for each $i$.
This is only possible because in PCA the $\J_i$ are constant, whereas for a general NF they are usually not.

As a simple fix, we average over a batch of Jacobian vectors to obtain the mean effect of one latent dimension:
\begin{equation}
    \overline{\J}_i \coloneq \Expt{\z}{\J_i(\z)}
\end{equation}
Formally, $\J_i(\z)$ denotes the unnormalized tangent vector of the latent manifold $\Man_i$ at a point $\x=\g(\z)$, thus $\overline{\J}_i$ can be viewed as an average unnormalized tangent vector of $\Man_i$.

This entire procedure allows us to visualize the influence of one latent dimension as a single image.
The more blurry $\overline{\J}_i$ appears, the more neighboring pixels (i.e. data dimension) are affected by one latent manifold and thus this could indicate on the strength of curvature of a latent manifold in data-space.
In practice, visualizing $\overline{\J}_i$ turned out to be a good proxy to infer the non-linear behaviour of varying a single latent dimension if the dataset is close to unimodal.
On multi-modal distribution it is unfortunately not that informative and the results appear very blurry and not that indicative, as latent manifolds can curve strongly from one mode to another.
As EMNIST contains roughly 10 modes, each associated with one digit, averaging $\overline{\J}_i$ over all modes is not that helpful.

GIN gets around this issue by learning a Gaussian Mixture Model (GMM) as the latent prior, where each mixture component is trained on one specific digit at a time. Thus we can sample from separate modes and visualize $\overline{\J}_i$ once for each.

We opted to also train a cINN \citep{ardizzone2019guided}, a NF which is conditioned on a known label.
By feeding the digit's number '0'-'9' as an additional condition to the feed-forward neural networks, the latent space is forced to be digit-independent.
This prohibits learning inter-modal variations, e.g. interpolating from a '1' to a '3' is not possible, and incentives latent representations to be shared across different conditions.
Thus the cINN allows us to sample from separate modes and visualize $\overline{\J}_i$ once for each.

\subsubsection{Comparing GIN and cINN}

To evaluate the manifold entropic metrics for GIN, we take latent samples from the learned Gaussian mixture distribution over the 10 digits.
As the entropic metrics were defined with a standard normal prior, we have to use the Jacobian $\J$ computed for the transformation from the unscaled latent samples $\z \sim \mathcal{N}(\z| 0, \I_D)$ to data samples $\x$, which is easily achieved in practice by rescaling the decoder Jacobian by the empirical standard deviations of each latent dimension and each digit.

\obso{
Thus $\z \sim \frac{1}{10} \sum_{j=0}^{9} \mathcal{N}(\hat{\mm}^{(j)}, \text{diag}({\hat{\sig}^{(j)}})^2)$ with the empirical mean $\hat{\mm}$ and the empirical variances ${\hat{\sig}^{(j)}_i}$ of each latent dimension $i$ and digit $j$.

To be comparable with the result from the cINN and as we defined the entropic metrics with a standard normal prior, we scale the Jacobian as $\J_i(\z^{(j)}) \rightarrow \J_i(\z^{(j)}) \cdot \hat{\sig}^{(j)}_i$ where $\z^{(j)}$ denotes a sample taken from the $j$-th mixture component.
}

The latent manifolds produced with GIN and those with the cINN are comparable, which can be seen by comparing \cref{fig:1-60 latent manifolds cINN} with \cref{fig:1-60 latent manifolds GIN}.
For example we can clearly infer that the 14. dimension in GIN and the 7. dimension in cINN vary the thickness of a digit.

\begin{figure}[!htbp]
    \centering
    \includegraphics[width=.6\textwidth]{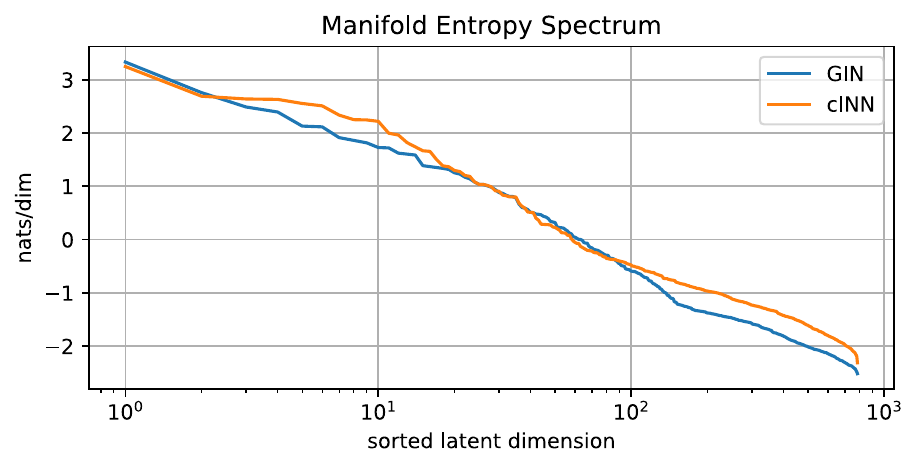}
    \caption{Manifold entropy spectrum of a trained GIN and a trained cINN shows strong Alignment in both models on a similar NF-architecture. Note that the x-axis is logarithmically scaled and the y-axis roughly scales with the logarithm of the manifold pdf variance i.e. its importance.}
    \label{fig:manifold entropy spectrum cINN GIN}
\end{figure}

In general we find that both cINN and GIN achieve comparable Alignment, which can be seen by a comparable manifold entropy spectrum \cref{fig:manifold entropy spectrum cINN GIN}.
The decrease in manifold entropy is very steep. Why these models can achieve decent Alignment without any training regularization is not yet known and left for future work.

To our surprise, we additionally found latent manifolds which are characterized by sharp edges and compression artifacts in the generated images and can be observed at the 34., 39. and 53. latent dimension in GIN and at the 18., 30. and 59. latent dimension in cINN.
We suspect that these latent manifolds, which we found consistently across many trained models, are artefacts of data preparations in the creation of the EMNIST dataset.
They were previously not found in \cite{Sorrenson2020Disentanglement} or other works.
As their influence on a generated image is very small, i.e. the magnitude in MSE is low, they are hard to notice but nonetheless revealed by our entropic metrics.

We additionally plot interpolated images of the cINN for the 20 most important latent dimensions in \cref{fig:NLL_latent_interpolations_1_20}.

\def\cwidth{0.8}

\begin{figure}[!htbp]
    \centering
    \includegraphics[width=\cwidth\textwidth]{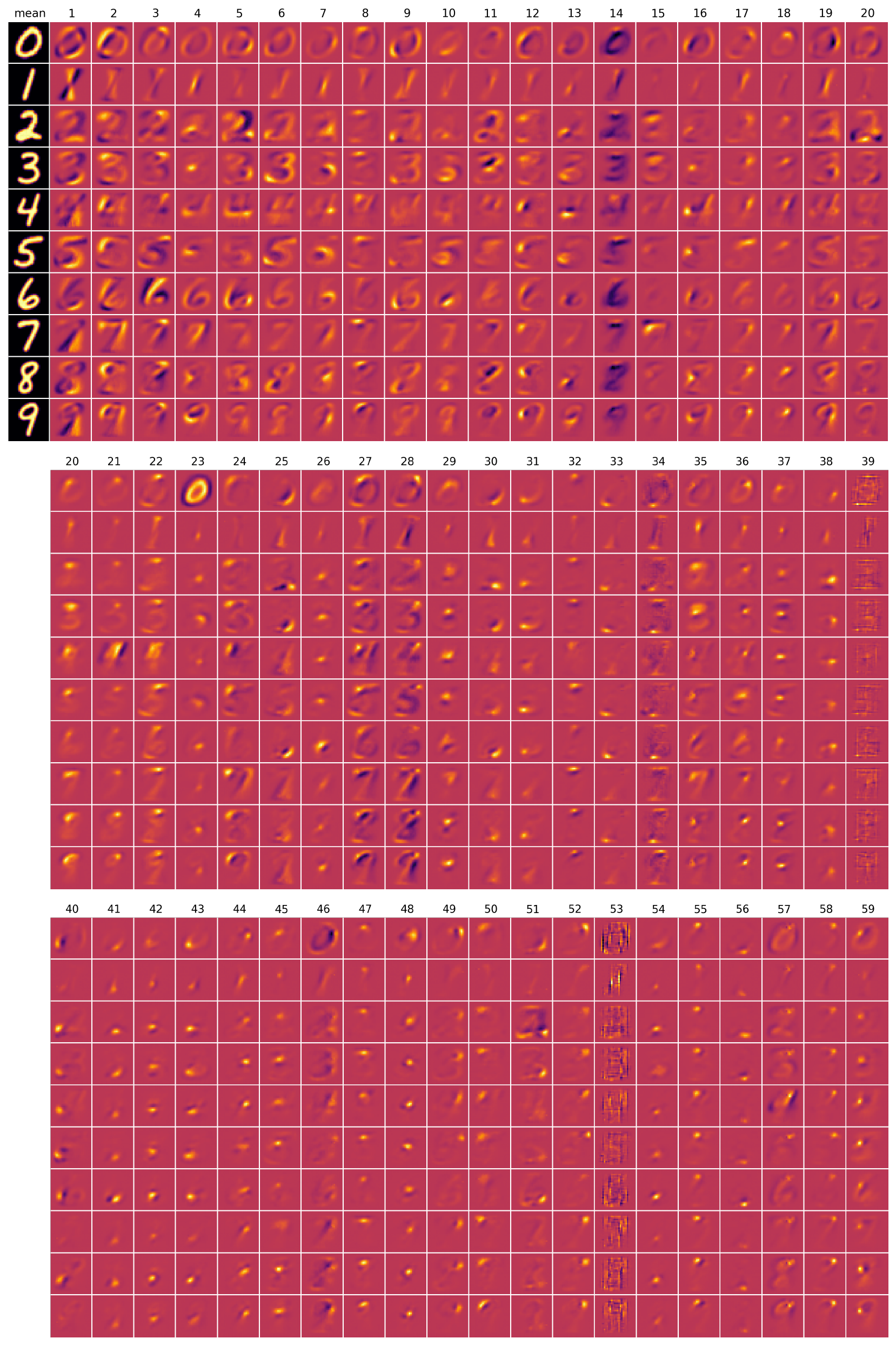}
    \caption{Visualization of the 60 most important latent manifolds (top left to bottom right) via $\overline{\J}_i$ of each mixture component of the trained GIN model. The mean of each mixture component is also plotted. Images are normalized for each latent dimension.}
    \label{fig:1-60 latent manifolds GIN}
\end{figure}

\begin{figure}[!htbp]
    \centering
    \includegraphics[width=\cwidth\textwidth]{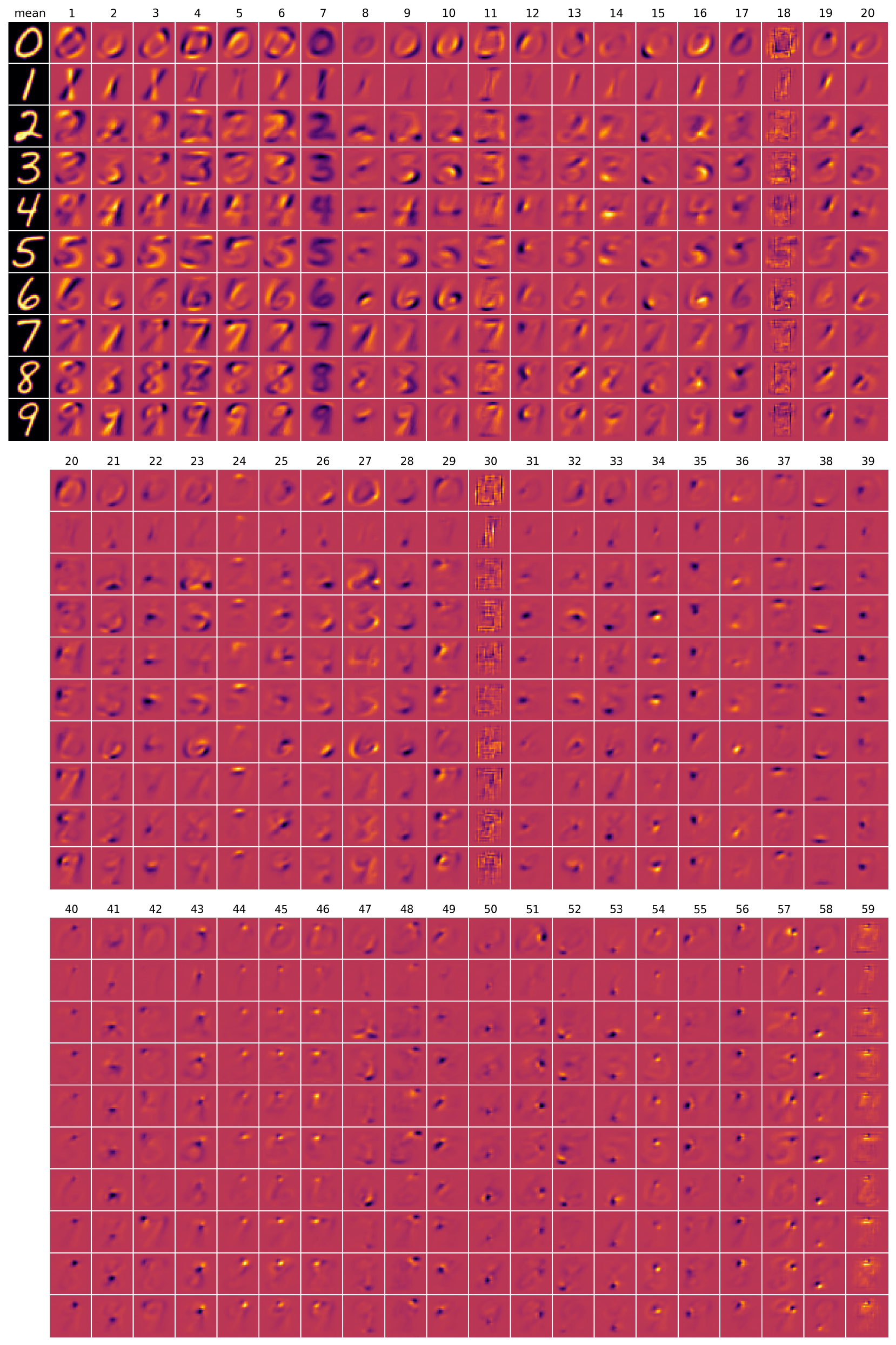}
    \caption{Visualization of the 60 most important latent manifolds (top left to bottom right) via $\overline{\J}_i$ conditioned on the digit of the trained cINN model. The mean is also plotted. Images are normalized for each latent dimension.}
    \label{fig:1-60 latent manifolds cINN}
\end{figure}

\begin{figure}[!htbp]
    \centering
    \includegraphics[width=\textwidth]{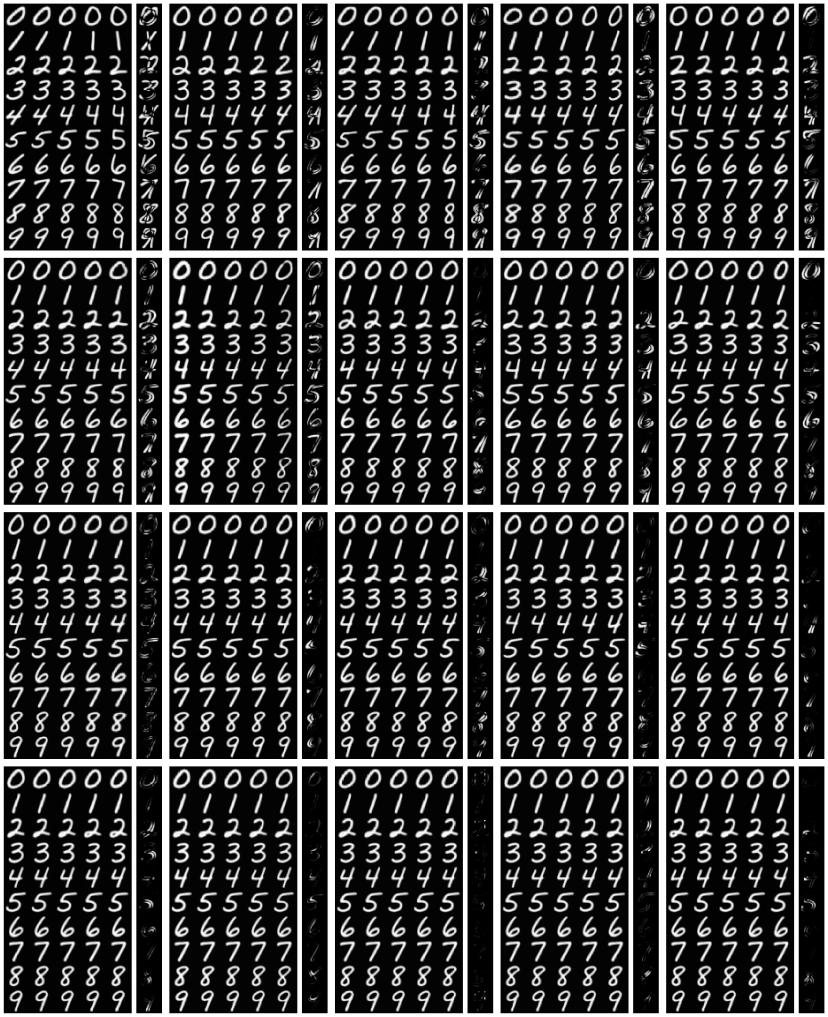}
    \caption{Image interpolations of the trained cINN model for latent dimensions 1 to 20 (top left to bottom right) conditioned on the digit 0-9. The respective dimension is interpolated in $[-2,-1,0,1,2]$ while the rest are fixed at zero. The right column shows the absolute difference from both extremes.}
    \label{fig:NLL_latent_interpolations_1_20}
\end{figure}

\subsubsection{Data preparation, training and architectures} \label{sec:emnist training}

The data is first normalized and then augmented with Gaussian noise $\sigma_{\text{noise}}=0.01$ at each iteration.

The cINN and GIN models are trained for 1000 epochs using Adam with a learning rate of 3e-4 and batchsize of 1024.

All architectures are implemented in FrEIA \citep{freia}.\\
\paragraph{cINN Architecture}\mbox{}

Checkerboard Downsampling: introduced in \cite{dinh2017densityestimationusingreal}.\\
AllInOneBlock: combines convolutional affine coupling block, 1x1 convolution, random permutation and ActNorm.\\
AllInOneBlock2: combines a fully-connected affine coupling block, random permutation and ActNorm.\\
"+10" refers to adding 10 extra dimensions to the input of the subnetworks to provide the one-hot-labels to the cINN.\\
Total number of learnable parameters: 4859768

\begin{table}[h]
    \caption{Architecture of the cINN network}
    \label{EMNIST_cINN_architecture}
    \begin{center}
    \begin{tabular}{||l|l|l|l||}
        \hline
        Type of block             & Number & Input shape & Subnetwork layer widths \\
        \hline
        Checkerboard Downsampling              & 1      & (1, 28, 28) &                                       \\
        AllInOneBlock                  & 6      & (4, 14, 14) & 2+10 $\rightarrow$ 16 $\rightarrow$ 16 $\rightarrow$ 4                       \\
        Checkerboard Downsampling              & 1      & (4, 14, 14) &                                       \\
        AllInOneBlock                  & 6      & (16, 7, 7)  & 8+10 $\rightarrow$ 64 $\rightarrow$ 64 $\rightarrow$ 16                       \\
        Flatten + PermuteRandom   & 1      & (16, 7, 7)  &                                       \\
        AllInOneBlock2                 & 8      & 784         & 392+10 $\rightarrow$ 201 $\rightarrow$ 201 $\rightarrow$ 784 \\
        \hline
    \end{tabular}
    \end{center}
\end{table}

\paragraph{GIN Architecture}\mbox{}

Comparable to the cINN architecture from above.
Refer to \cite{Sorrenson2020Disentanglement}.\\
Total number of learnable parameters: 5214400

\begin{table}[h]
    \caption{Architecture of the GIN network}
    \label{EMNIST_GIN_architecture}
    \begin{center}
    \begin{tabular}{||l|l|l|l||}
        \hline
        Type of block             & Number & Input shape & Subnetwork layer widths \\
        \hline
        Checkerboard Downsampling     & 1      & (1, 28, 28) &           \\
        GIN Convolutional Coupling    & 6      & (4, 14, 14) & 2 $\rightarrow$ 16 $\rightarrow$ 16 $\rightarrow$ 4  \\
        Checkerboard Downsampling     & 1      & (4, 14, 14) &                                       \\
        GIN Convolutional Coupling    & 6      & (16, 7, 7)  & 8 $\rightarrow$ 64 $\rightarrow$ 64 $\rightarrow$ 16  \\
        Flatten + PermuteRandom       & 1      & (16, 7, 7)  &                                       \\
        GIN Fully-Connected Coupling  & 8      & 784         & 392 $\rightarrow$ 201 $\rightarrow$ 201 $\rightarrow$ 784 \\
        \hline
    \end{tabular}
    \end{center}
\end{table}

\newpage

\subsubsection{Wavelet-Flow}

For the Wavelet-Flow we use an architecture adapted from \cite{yu2020wavelet}.

The most important distinction to a GLOW-like NF is that in each split, the finely downsampled part is trained conditionally on the (unprocessed) coarse part. This has the main benefit of being more parameter-efficient while retaining a comparable DE-performance.

In an ablation study we found that precisely this conditioning enables the desired DRL capabilities, highlighted in the main paper, whereas replacing the Haar-wavelet with a Checkerboard downsampling didn't meaningfully degrade DRL performance.
This could hint, that a hierarchical and well-structured conditional architecture could be all we need to achieve strong DRL performance.

\begin{figure}[!htbp]
    \centering
    \includegraphics[width=.5\linewidth]{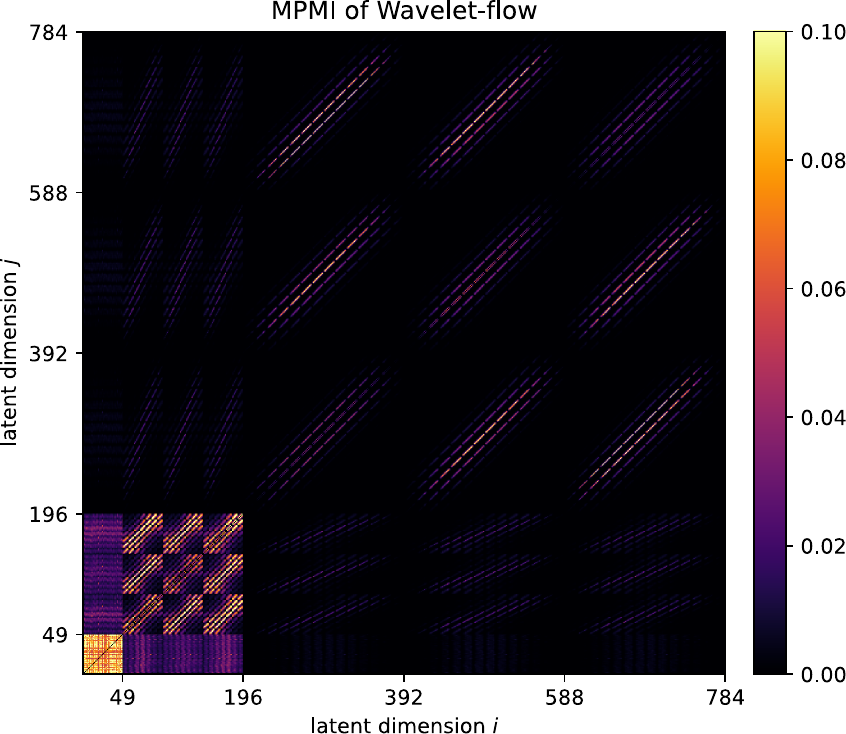}
    \caption{MPMI matrix of a trained Wavelet-flow. Only convolutional coupling blocks are used in processing the index sets $\set{M}=\{50,...,196\}$ and $\set{D}=\{197,...,784\}$ resulting in a more intricate structure of the latent space.}
    \label{fig:MPMI wavelet}
\end{figure}

In \cref{fig:MPMI conv-NF wavelet comparison} (right) we plotted the MPMI-matrix of a trained wavelet-flow which has both convolutional and fully-connected blocks processing the finer parts $\set{M}$ and $\set{D}$.
In that way, the entries in the MPMI-matrix in $\set{M}$ and $\set{D}$ appear homogeneous.
A Wavelet-flow with purely convolutional blocks in $\set{M}$ and $\set{D}$, as originally proposed in \cite{yu2020wavelet}, has comparable DRL characteristics via Alignment and Disentanglement.
Though, as the latent subspaces of $\z_\set{M}$ and $\z_\set{D}$ are only processed by convolutional layers, the resulting MPMI-matrix has a more intricate structure, plotted in \cref{fig:MPMI wavelet}.

\subsection{$\beta$-VAEs}

\subsubsection{Architecture and Training} \label{sec:beta-VAE training}
We use a VAE-architecture from Pythae-library \citep{chadebec2023pythaeunifyinggenerativeautoencoders} via the classes \texttt{Decoder\_Conv\_AE\_MNIST} and \texttt{Encoder\_Conv\_VAE\_MNIST}.

Training was performed using the AdamWScheduleFree optimizer \citep{defazio2024road} with a learning rate of 3e-4 and a batchsize of 128 for 50 epochs.

\subsubsection{Additional results} \label{sec:beta-VAE Additional results}
\begin{figure}[!htbp]
    \centering
    \includegraphics[width=.7\linewidth]{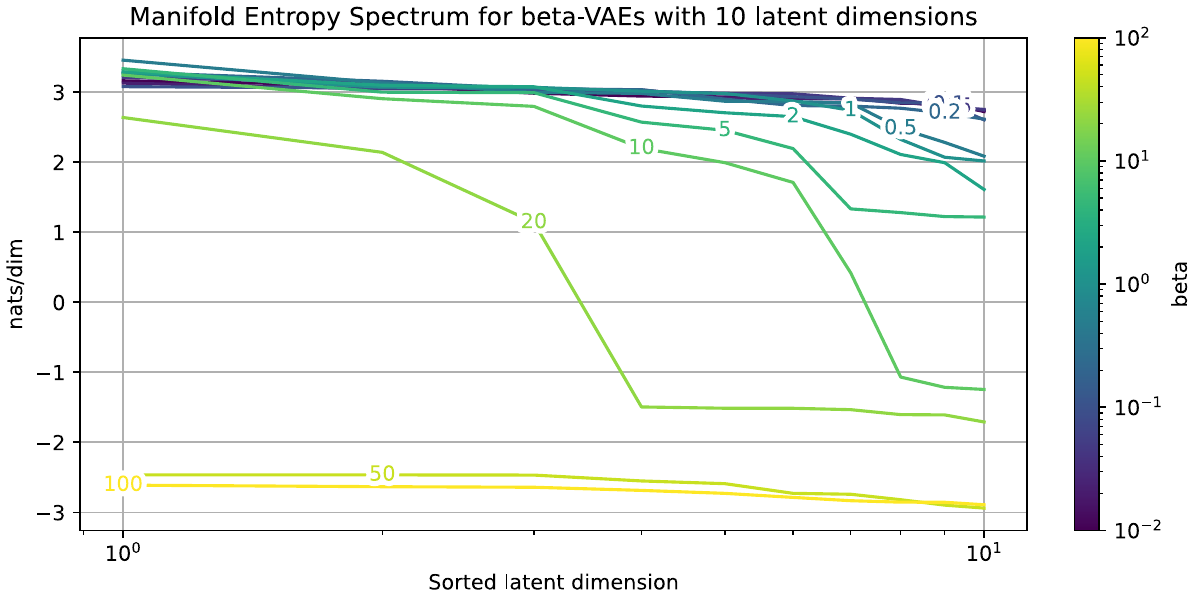}\
    \caption{Manifold entropy spectrum for different trained $\beta$-VAEs with a latent space size of 10. The latent space is being compressed more with increasing $\beta$-value and becomes uninformative for $\beta>20$.}
    \label{fig:beta-VAE manifold-entropy 100}
\end{figure}

In addition to $\beta$-VAEs with 100 latent dimensions, we also trained $\beta$-VAEs with 10 latent dimensions.
There we observe a similar behaviour to the 100-latent-dimension models, which can be seen in the manifold entropy spectrum \cref{fig:beta-VAE manifold-entropy 100}.

\begin{figure}[!htbp]
    \centering
    \includegraphics[width=.49\linewidth]{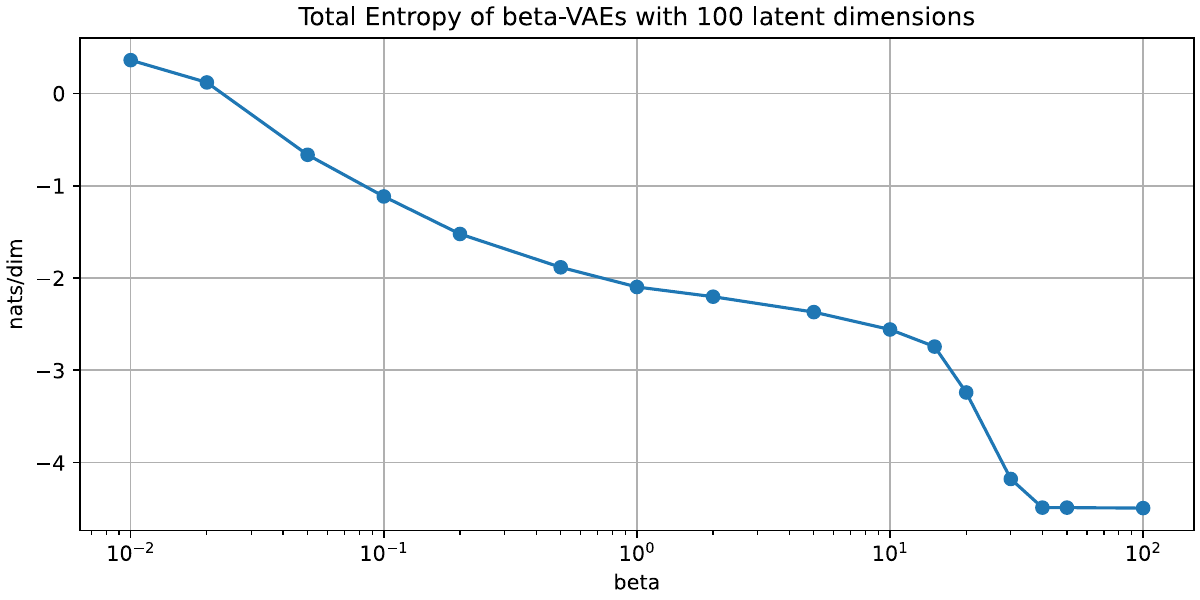}
    \includegraphics[width=.49\linewidth]{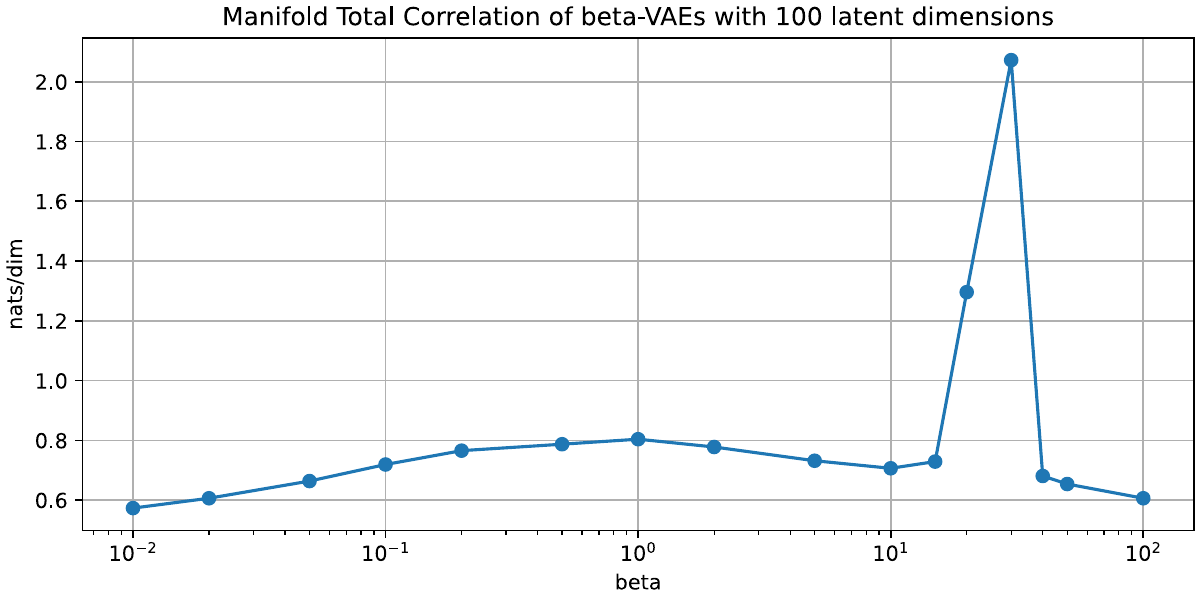}
    \caption{Total Entropy (left) and Manifold Total Correlation (right) for different trained $\beta$-VAEs with a latent space size of 100. The entropic metrics are divided by 100 to obtain a measure in nats/dim.}
    \label{fig:beta-VAE entropic metrics 100}
\end{figure}

\begin{figure}[!htbp]
    \centering
    \includegraphics[width=.49\linewidth]{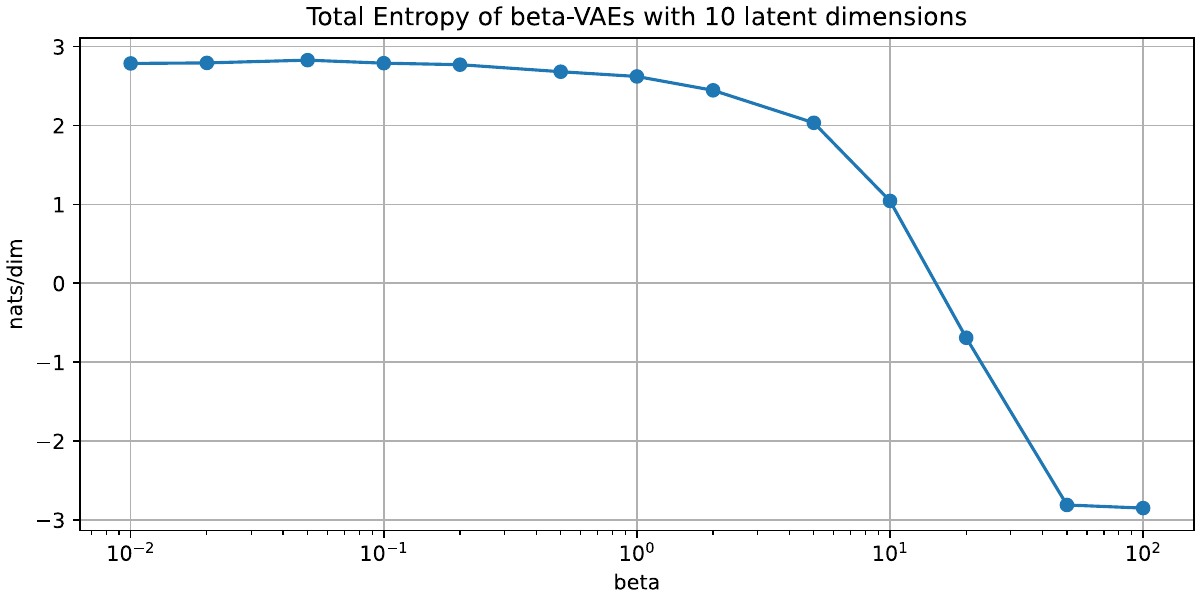}
    \includegraphics[width=.49\linewidth]{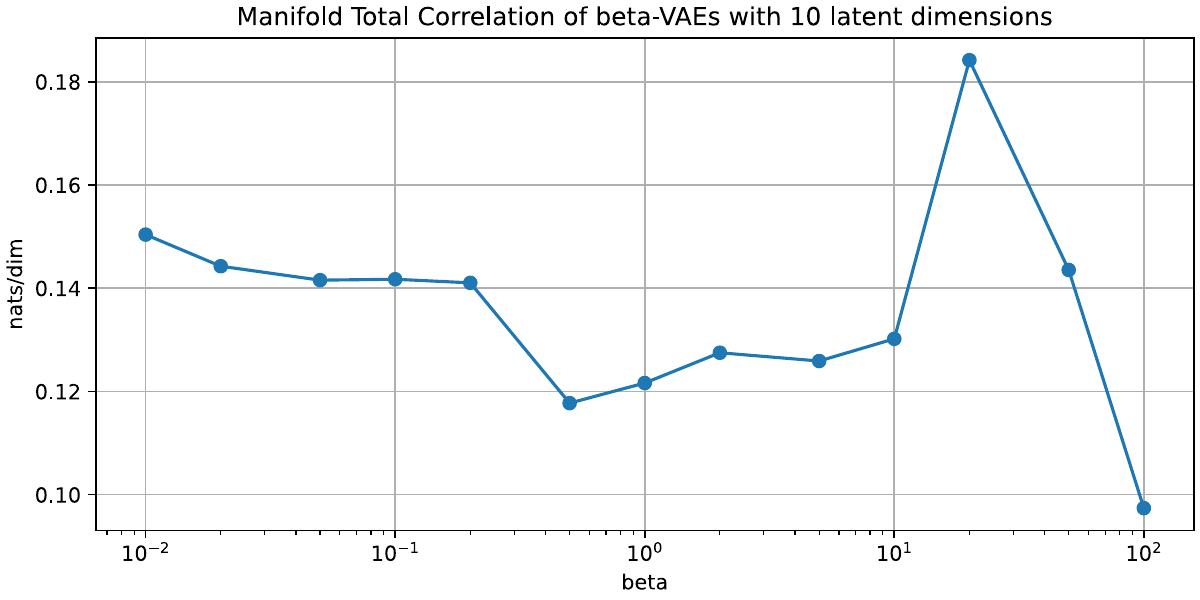}
    \caption{Total Entropy (left) and Manifold Total Correlation (right) for different trained $\beta$-VAEs with a latent space size of 10. The entropic metrics are divided by 10 to obtain a measure in nats/dim.}
    \label{fig:beta-VAE entropic metrics 10}
\end{figure}

In \cref{fig:beta-VAE entropic metrics 100},\ref{fig:beta-VAE entropic metrics 10} we additionally plot the total entropy and manifold total correlation with respect to $\beta$ for both cases.
In the 100-latent-dimension models we observe a strong increase in the manifold total correlation at $\beta \sim 20-30$, which is when the models degenerates, as can be additionally checked in the total entropy.

\newpage

\vfill

\end{document}